IMT Institute for Advanced Studies Lucca

Lucca, Italy

**Regional Active Contours based on Variational level sets and Machine Learning for Image Segmentation**

PhD programme in Computer Science and Engineering

XXVII cycle

By

**Mohammed Mohammed Abdelsamea Ahmed**

2015

**The dissertation of Mohammed Mohammed Abdelsamea Ahmed is approved.**

Program Coordinator: Prof. Rocco De Nicola, IMT Institute for Advanced Studies, Lucca, Italy

Supervisor: Assistant Prof. Giorgio Gnecco, IMT Institute for Advanced Studies, Lucca, Italy

Supervisor: Prof. Mohamed Medhat Gaber, Robert Gordon University, Aberdeen, UK

Tutor: Assistant Prof. Giorgio Gnecco, IMT Institute for Advanced Studies, Lucca, Italy

The dissertation of Mohammed Mohammed Abdelsamea Ahmed has been reviewed by:

Prof. Ivan Jordanov, University of Portsmouth, UK

Prof. Frederic Stahl, University of Reading, UK

Prof. Mykola Pechenizkiy, Eindhoven University of Technology, Netherlands

**IMT Institute for Advanced Studies, Lucca**

**2015**

Dedicated to my wife Omnia and our sons Abdelrahman and Ziad.

# Table of Contents

# List of Figures

# List of Tables

# Acknowledgments

First of all I wish to express my sincere gratitude to my advisor, Assist. Prof. Giorgio Gnecco, for his incredible guidance, serenity, and support. In particular I am highly indebted and appreciated his advice and guidance in every line that I wrote during his supervision. I would like also to express my deepest gratitude to my co-advisor, Prof. Mohamed Medhat Gaber, for his advises, guidance, great vision, and the great opportunity for hosting me at Robert Gordon University during my visiting research period.

Moreover, I would especially like to thank all of my coauthors, without whom, in all honesty, the research presented in this thesis would never have been possible. I also must thank the faculty and staff of the IMT Institute for Advanced Studies, Lucca for their professional treatment of me as more of a colleague. In particular I would like very much to thank the staff of the Facilities Office for helping me during my stay in Italy.

Last but not least, I would like to thank my wife for her patience and encouragement during my study.

The research outcomes of this thesis are reported in the following publications:

**Journal Papers**:

1. M. M. Abdelsamea and G. Gnecco. Robust Local-Global SOM-based ACM. Electronics Letters, vol. 51, pp. 142–143, 2015.

2. M. M. Abdelsamea, G. Gnecco, and M. M. Gaber. An efficient self organizing active contour model for image segmentation. Neurocomputing, vol. 149, Part B, pp. 820–835, 2015.

3. M. Minervini, M. M. Abdelsamea, and S. A. Tsaftaris, Image-based plant phenotyping with incremental learning and active contours, Ecological Informatics, vol. 23, pp. 35–48. 2014, Special Issue on Multimedia in Ecology and Environment.

4. M. M. Abdelsamea, G. Gnecco, M. M. Gaber, and E. Elyan, On the relationship between variational level set-based and SOM-based active contours, submitted to Computational Intelligence and Neuroscience, 2014.

5. M. M. Abdelsamea, G. Gnecco, M. M. Gaber, A SOM-based Chan-Vese Model for unsupervised image segmentation, submitted to Image and Vision Computing, 2014.

**Conference Papers**:

1. M. M. Abdelsamea, G. Gnecco, and M. M. Gaber. A Survey of SOM-based Active Contours for Image Segmentation. In Proceedings of the 10th Workshop on Self- Organizing Maps (WSOM 2014), Advances in Intelligent Systems and Computing, volume 295, pages 293–302. Springer, 2014.

2. M. M. Abdelsamea, G. Gnecco, and M. M. Gaber. A concurrent SOMbased Chan-Vese model for image segmentation. In Proceedings of the 10th Workshop on Self-Organizing Maps (WSOM 2014), Advances in Intelligent Systems and Computing, volume 295, pages 199–208. Springer, 2014.

# Abstract

Image segmentation is the problem of partitioning an image into different subsets, where each subset may have a different characterization in terms of color, intensity, texture, and/or other features. Segmentation is a fundamental component of image processing, and plays a significant role in computer vision, object recognition, and object tracking. Active Contour Models (*ACMs*) constitute a powerful energy-based minimization framework for image segmentation, which relies on the concept of contour evolution. Starting from an initial guess, the contour is evolved with the aim of approximating better and better the actual object boundary.

Handling complex images in an efficient, effective, and robust way is a real challenge, especially in the presence of intensity inhomogeneity, overlap between the foreground/background intensity distributions, objects characterized by many different intensities, and/or additive noise. In this thesis, to deal with these challenges, we propose a number of image segmentation models relying on variational level set methods and specific kinds of neural networks, to handle complex images in both supervised and unsupervised ways. Experimental results demonstrate the high accuracy of the segmentation results, obtained by the proposed models on various benchmark synthetic and real images compared with state-of-the-art active contour models.

# Chapter 1

# Introduction

## 1.1 Image segmentation

Image segmentation is the problem of partitioning an image $I(x)$, where $x$ is the pixel location within the image, into different subsets $\Omega_i$, where each subset may have a different characterization in terms of color, intensity, texture, and/or other features used as similarity criteria. Segmentation is a fundamental component of image processing, and plays a significant role in computer vision, object recognition, and object tracking.

Traditionally, image segmentation methods can be classified into five categories. The first category is made up of threshold-based segmentation methods [83]. These methods are pixel-based, and usually divide the image into two subsets, i.e., the foreground and the background, using a threshold on the value of some feature (e.g., gray level, color value). These methods assume that the foreground and background in the image have different ranges for the values of the features to be thresholded. Over the years, many different thresholding techniques have been developed including Minimum error thresholding, Moment-preserving thresholding, Otsu's thresholding, just to mention a few. The most popular thresholding method, Otsu's algorithm [78], improves the image segmentation performance over other threshold-based segmentation methods in the following way. The threshold used in Otsu's al-gorithm is chosen in such a way to optimize a trade-off between the maximization of the inter-class variance (i.e., between pairs of pixels belonging to the foreground and the background, respectively) and the minimization of the intra-class variance (i.e., between pairs of pixels belonging to the same region). Otsu's thresholding algorithm is good for thresholding an image whose intensity histogram is either bimodal or multimodal (i.e., it provides a satisfactory solution in the case of the segmentation of large objects with nearly uniform intensities, significantly different from the intensity of the background). However, it has not the ability to segment images with unimodal distribution (e.g., images containing small objects with different intensities), and its output is sensitive to noise. Thus, post-processing operations are usually required to obtain a final satisfactory segmentation.

The second category of methods is called boundary-based segmentation [72]. These methods detect boundaries and discontinuities in the image based on the assumption that the intensity values of the pixels linking the foreground and the background are distinct. The first/second order derivatives of the image intensity are usually used to highlight those pixels (e.g., Sobel and Prewitt edge detectors [72] as first-order methods, and the Laplace edge detector [83] as a second-order method, respectively). The difference between first and second order methods is that the latter can localize the local displacement and orientation of the boundary. By far the most accurate technique of detecting boundaries and discontinuities in an image is the Canny edge detector [25]. The Canny edge detector is less sensitive to noise than other edge detectors, as it convolves the input image with a Gaussian filter. The result is a slightly blurred version of the input image. This method is also very easy to be implemented. However, it is very sensitive to noise, and leads to segmentation results characterized by a discontinuous detection of the object boundaries.

The third category of methods [49] is called region-based segmentation. Region-based segmentation techniques divide an image into subsets based on the assumption that all neighboring pixels within one subset have a similar value of some feature, e.g., image intensity. Region growing [12] is the most popular region-based segmentation technique. In region growing, one has to identify at first a set of seeds as initial representatives of the subsets. Then, the features of each pixel are compared to the features of its neighbors. If a suitable predefined criterion are satisfied, then the pixel is classified as belonging the same subset associated with its "most similar" seed. Accordingly, region growing relies on the prior information given by the seeds and the predefined classification criterion. A second popular region-based segmentation method is region "splitting and merging". In such method, the input image is first divided into several small regions. Then, on the regions, a series of splitting and merging operations are performed and controlled by a suitable predefined criterion. As region-based segmentation is an intensity-based method, the segmentation result in general leads to nonsmooth and badly shaped boundary for the segmented object.

The fourth category of methods [81] is learning-based segmentation. There are two general strategies for developing learning-based segmentation algorithms: namely, generative learning and discriminative learning. Generative learning [21] utilizes data set of examples to build a probabilistic model, by finding the best estimates of parameters for some prespecified parametric form of a probability distribution. One problem with these methods is that the best estimates of the parameters may not provide a satisfatory model, because the parametric model itself may not be correct. Another problem is that the classification/clustering framework associated with a parametric probabilistic model may not provide an accurate description of the data due to limited number of parameters in the model even in the case in which its training is well performed. Techniques following the generative approach include K-means [67], the Expectation-Maximization algorithm [94], and Gaussian Mixture Models [36]. Discriminative learning [88, 15, 48, 113] ignores probability and attempts to construct a good decision boundary directly. Such an approach is often extremely successful, especially when no reasonable parametric probabilistic model of the data exists. It assumes that the decision boundary comes from another class of nonparametric solutions, and chooses the best element of that class according to a suitable opimality criterion. Techniques following the discriminative approach include Linear Discriminative Analysis [44], Neural Networks [69, 41, 82, 43, 39] and Support Vector Machines [22]. The main problems with these methods are their sensitivity to noise and the discontinuity of the resulting object boundaries.

The last category of methods [23, 50] are energy-based segmentation methods. This class of methods is based on an energy functional[1] and deals with the segmentation problem as an optimization problem, which tries to divide the image into regions based on the maximization/minimization of the energy functional. The most well-known energy-based segmentation techniques are called "active contours". The main idea of active contours is to choose an initial contour inside the image domain to be segmented, then make such a contour evolve by using a series of shrinking and expanding operations. Some advantages of the active contours over the aforementioned methods are that topological changes of the objects to be segmented can be handled implicitly. More importantly, complex shapes can be modeled without the need of prior knowledge about the image. Finally, rich information can be inserted into the energy functional (e.g., boundary-based and region-based information).

## 1.2 Motivation

Current active contour models with/without prior knowledge incorporated into the energy functional might be efficient/effective enough to handle complex images. However, they are not always the most efficient and effective solutions to handle images with complex intensity distributions. As a consequence, we believe that a real challenge in active contour models is to improve their efficiency and effectiveness in segmenting images characterized by complex intensity distributions. Motivated by this issue, we mainly focus on developing effective, efficient and/or robust supervised and unsupervised level set image segmentation frame-works to deal (in a global/local way) with a variety of images characterized by many intensity levels, intensity inhomogeneity, and/or presenting other computer-vision challenges (see Fig. 1.1 for some challenging images used in this thesis). Both single-channel and multi-channel images are considered in this work.

---
[1]Loosely speaking, a functional is defined as a function of a function, i.e., a function that takes a vector as its input argument and returns a scalar.

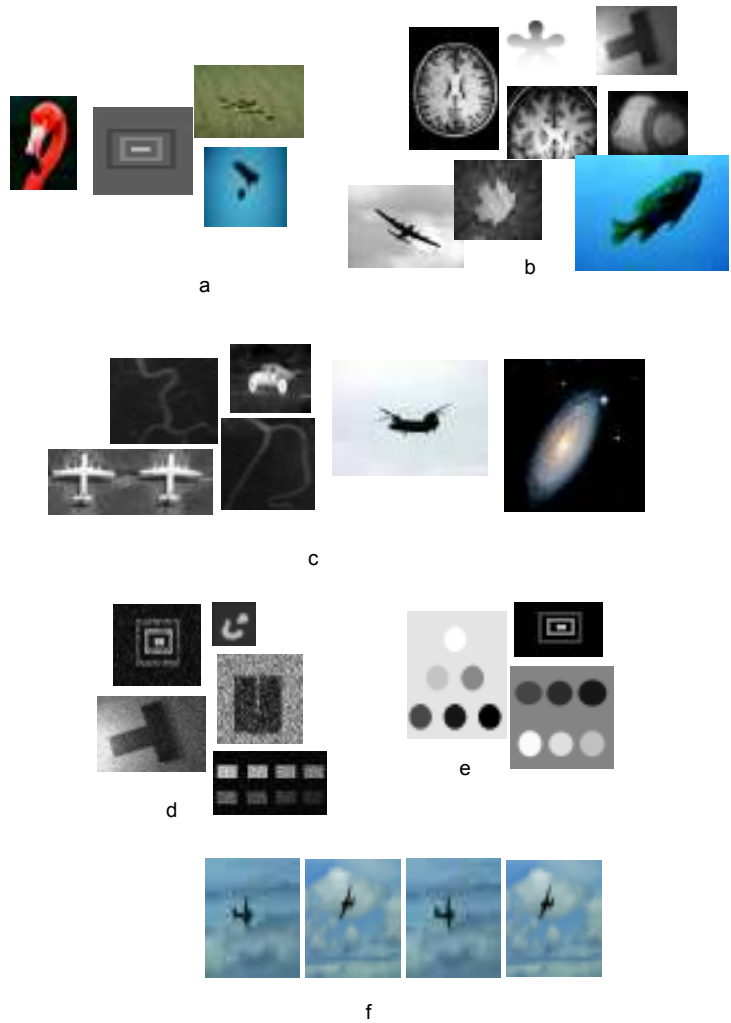

Figure 1.1: Some of the challenging single-/multi-channel images considered in this work: images with (a) foreground/background intensity overlap and inhomogeneous regions, (b) intensity inhomogeneity, (c) weak and ill-defined edges and shadows, (d) additive noise, (e) many intensity levels; sequences of images with (f) scene changes.

## 1.3 Contributions

In this thesis, we first present a survey about the state of the art of active contour models with a focus on their strengths and weaknesses. Then, we propose a number of novel active contour models, which are able to handle images presenting challenges in computer vision in an efficient, effective, and/or robust way (e.g., see Fig. 1.2 for examples of images that each model can handle). We also compare such approaches with state-of-the-art segmentation models, focusing on active contour models (*ACMs*) (e.g., a brief overview of some of the regional state-of-the-art *ACMs* and our proposed ones is reported in Table 1.1), but considering also other segmentation methods, such as thresholding ones. In the present section, we briefly describe the main contributions of the proposed *ACMs*. It is worth noting that some of the terms that are used here to describe these contributions will be further illustrated in the following chapter:

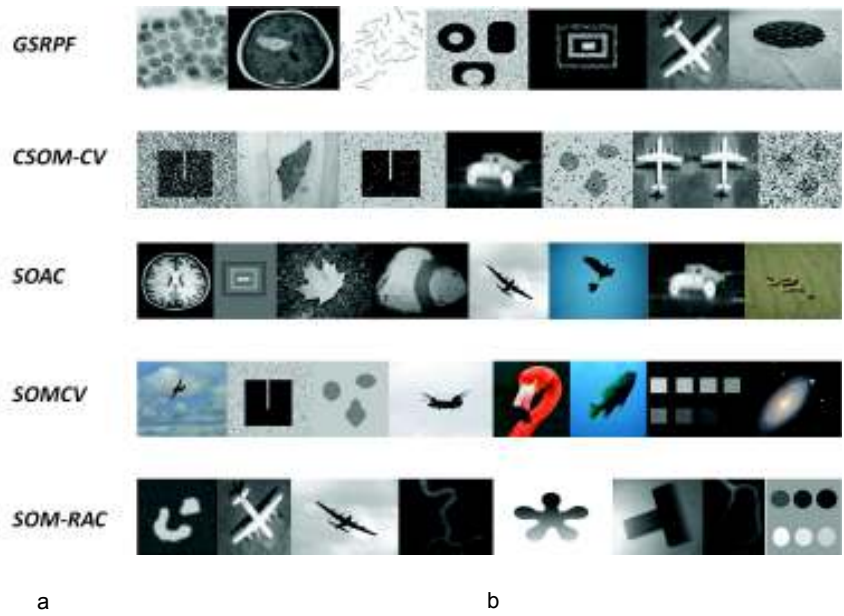

a b

Figure 1.2: Examples of challenging images (b) that can be effectively handeled by our proposed models (a).

Table 1.1: A summary of some of the regional Active Contour Models (*ACM*)s presented in the thesis.

| ACM | Ref. | Regional information | | Model overview |
|---|---|---|---|---|
| | | Local | Global | |
| *GAC* | [26] | No | No | Makes use of boundary information but behaves badly with weak edges. |
| *CV* | [28] | No | Yes | Can handle objects with blurred boundaries by making strong statistical assumptions. |
| *SBGFRLS* | [124] | No | Yes | Very efficient computationally but still makes strong statistical assumptions. |
| *GSRPF* | [11] | No | Yes | More efficient and robust compared to *SBGFRLS* but still makes strong statistical assumptions. |
| *LBF* | [61] | Yes | No | Can handle complex distributions with inhomogeneities while sensitive to initial contour. |
| *LIF* | [122] | Yes | No | Behaves likewise *LBF* and is computationally more efficient but still sensitive to the initial contour. |
| *LRCV* | [64] | Yes | No | Computationally very efficient compared to *LBF* and *LIF* but sensitive to the initial contour. |
| *GMM-AC* | [59] | No | Yes | Exploits prior knowledge but makes strong statistical assumptions. |
| *SISOM* | [104] | No | No | Localizes the salient contours using a *SOM* but topological changes cannot be handled. |
| *TASOM* | [86] | No | No | Adjusts automatically the number of *SOM* neurons but no topological changes can be handled. |
| *BSOM* | [103] | No | Yes | Exploits regional information but topological changes cannot be handled. |
| *eBSOM* | [101] | No | Yes | Produces smooth contours but topological changes cannot be handled. |
| *FTA-SOM* | [46] | No | Yes | Converges quickly but topological changes cannot be handled. |
| *CFBL-SOM* | [121] | No | Yes | Exploits prior knowledge but topological changes cannot be handled. |
| *CAM-SOM* | [84] | No | Yes | Can handle objects with concavities but topological changes cannot be handled. |
| *KDE-ACM* | [34] | Yes | No | Models arbitrary shapes relying on huge amount of supervised information. |
| *CSOM-CV* | [8] | No | Yes | Very robust to the noise relying on supervised information. |
| *SOAC* | [9] | Yes | No | Can handle complex images in a local way based on supervised information. |
| *SOMCV* | [10] | No | Yes | Reduces the intervention of the user and handles images in global way. |
| *SOM-RAC* | [7] | Yes | Yes | Is robust to noise, scene changes, and inhomogeneities but is computationally expensive. |

**Globally Signed Region Pressure Force (*GSRPF*)-based active contour model**. The *GSRPF*-based *ACM* [11] is designed to segment, using global intensity information, images possibly characterized by a non symmetric intensity distribution of the Region Of Interest (*ROI*). The model has the following strengths: 1) it can accurately modulate the sign of the "pressure" force inside and outside the contour which is used to guide the contour evolution; 2) it can handle images with many intensity levels in the foreground; 3) it is robust to additive noise; and 4) offers high efficiency and rapid convergence. The proposed *GSRPF* model is robust to contour initialization and has the ability to stop the curve evolution close even to ill-defined (weak) edges. Our model provides a parameter-free environment which allows a minimal user intervention. Experimental results on several synthetic and real images demonstrate the high accuracy of the segmentation results obtained by the proposed model in comparison to the segmentations obtained by other methods adopted from the literature.

**Concurrent Self Organizing Map-based Chan-Vese (*CSOM-CV*) model**. *CSOM-CV* [8] is a novel regional *ACM*, which relies on a Concurrent Self Organizing Map *CSOM* to approximate globally the foreground and background image intensity distributions in a supervised way, and to drive the evolution of the active contour accordingly. The model integrates such information into the framework of the Chan-Vese (*C-V*) model, which is the reason for which we coined the term *CSOM-CV* for the proposed model. The main idea of the *CSOM-CV* model is to concurrently integrate the global information extracted by a *CSOM* from a small percentage of the total number of pixels in the image. The information coming from such supervised pixels is incorporated into the level-set framework of the *C-V* model to build an effective *ACM*. The proposed model integrates the advantages of *CSOM* as a powerful classification tool and the *C-V* model as an effective tool for the optimization of a global energy functional. Experimental results show the higher effectiveness of *CSOM-CV* in segmenting both synthetic and real images, when compared with the stand-alone *C-V* and *CSOM* models.

**Self Organizing Active Contour (*SOAC*) model**. The *SOAC* model [9] can be described as a variational level set method driven by the prototypes (weights) of neurons belonging to a Self Organizing Map (*SOM*), obtained after a training session. Such prototypes are able to keep track of the dissimilarity beween the foreground and background intensity distributions. A difference with the *CSOM-CV* model is that the information in the *SOAC* model is local. The *SOAC* model can handle images characterized by many intensity levels, intensity inhomogeneity, and complex distributions, possibly with a complicated foreground and background overlap. Experimental results demonstrate the higher accuracy of the segmentation results obtained by the *SOAC* model on several synthetic and real images, when compared with the segmentations obtained by other well-known active contour models.

**SOM-based Chan-Vese (*SOMCV*) model**. *SOMCV* model [10] is similar to the *CSOM-CV* model, with the difference that now the training of the model is completely unsupervised. Also in this case, the prototypes of the trained neurons encode global intensity information. The proposed model can handle images with many intenisty levels and complex intensity distributions, and is robust to additive noise. Experimental results show the higher accuracy of the segmentation results obtained by the proposed model on several synthetic and real images, when compared with the *C-V* active contour model. A significant difference with the *CSOM-CV* model is that the intervention of the final user is significantly reduced in the *SOMCV* model, since no supervised information is used.

**SOM-based Regional Active Contour (*SOM-RAC*) model**. Finally, likewise the *SOMCV* model, also the *SOM-RAC* model [7] relies on the global information coming from selected prototypes associated with a *SOM*, which is trained off-line in an unsupervised way to model the intensity distribution of an image, and used on-line to segment an identical or similar image. In order to improve the robustness of the model, global and local information are combined in the on-line phase. Experimental results show the higher accuracy of the segmentations obtained by the *SOM-RAC* model on several synthetic and real images, when compared with a state-of-the-art local *ACM*, namely, the *Local Region-based Chan-Vese model*.

## 1.4 Thesis organization

This thesis is organized in a number of chapters. Chapter 2 illustrates the main concepts of variational level set-based *ACM*s, and reviews the development of the state-of-the-art models from a machine learning perspective. Chapter 3 reviews various kinds of *SOM*-based *ACM*s, with a focus on their strengths and weaknesses in comparison with level set-based *ACM*s. Chapter 4 presents the proposed *GSRPF* model as a global unsupervised sign pressure force *ACM*. Chapter 5 and 6 describe our proposed *CSOM-CV* and *SOAC* models as global and local supervised *ACM*, respectively. Chapter 7 presents the proposed *SOMCV* model as a global unsupervised *ACM*. Chapter 8 presents the *SOM-RAC* model as a local-global unsupervised *ACM*. Finally, Chapter 9 concludes the thesis, and presents some possible future research directions.

# Chapter 2

# Variational level set-based *ACM*s

Active contour, sometimes called "Evolving Front", is a contour *C* inside the image domain Ω which evolves and is deformed through a set of shrink/expansion operations. Such a process known as "Contour Evolution", has the purpose of fitting the contour to the boundary of an object to be segmented from an image $I(x)$, and is governed by the minimization of an energy functional.

Active Contour Models (*ACM*s) usually deal with the image segmentation problem as a functional optimization problem, as they try to divide an image into several regions by optimizing a suitable functional. Starting from an initial contour, the optimization is performed in an iterative way, evolving the current contour with the aim of approximating better and better the actual boundary (hence the denomination "active contour" models, which is actually used also for models which are not based on the explicit minimization of a functional [102]).

To build an active contour, there are mainly two methods. The first one is an explicit or Lagrangian method, which results in parametric active contours, also called Snakes. The second one is an implicit or Eulerian method, which results in geometric active contours, known as level set method.

In parametrized *ACM*s, the contour *C*, see Fig. 2.1, is represented as

$$C := \{x \in \Omega : x = (x_1(s), x_2(s)), 0 \leq s \leq 1\}, \qquad (2.1)$$

where $x_1(s)$ and $x_2(s)$ are functions of the scalar parameter *s*. A representative parametrized *ACM* is the Snakes model, proposed by Kass et al. [50] (see also [115] for successive developments).

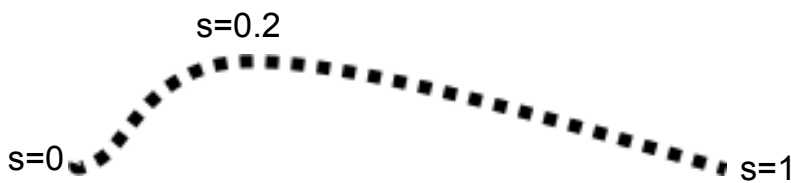

Figure 2.1: The parametric representation of a contour.

The main drawbacks of parametrized *ACM*s are the frequent occurrence of local minima in the image energy functional to be optimized (which is mainly due to the presence of a gradient energy term inside such a functional), and the fact that topological changes of the objects (e.g., merging and splitting) cannot be handled during the evolution of the contour. Instead, level set methods - which will be described in the next section - can model arbitrarily complex shapes. Moreover, another advantage with respect to parametetric methods is that they can handle also topological changes of the contours.

In this chapter, we review some representative variational level set-based *ACM*s, from a machine learning perspective, with a focus on their advantages and disadvantages in modeling the evolving contour via a level set.

## 2.1 Variational level set-based *ACM*s

The difference between parametric active contour and geometric (or variational level set-based) active contour models is that in geometric active contours, the contour is implemented via a variational level set method. Such a representation was first proposed by Osher and Sethian [77]. In such methods, the contour C, see Fig. 2.2, is implicitly represented by a function $\phi(x)$, called "level set function", where $x$ is the pixel location inside the image domain $\Omega$. The contour C is then defined as the zero level set of the function $\phi(x)$, i.e.,

$$C := \{x \in \Omega : \phi(x) = 0\}. \tag{2.2}$$

A common and simple expression for $\phi(x)$, which is used by most authors, is

$$\phi(x) = \begin{cases} +\rho, \text{ for} & x \in \text{inside}(C), \\ 0, \text{ for} & x \in C, \\ -\rho, \text{ for} & x \in \text{outside}(C), \end{cases} \tag{2.3}$$

where $\rho$ is a positive parameter (possibly dependent on $x$ and $C$, in such case it is denoted by $\rho(x, C)$).

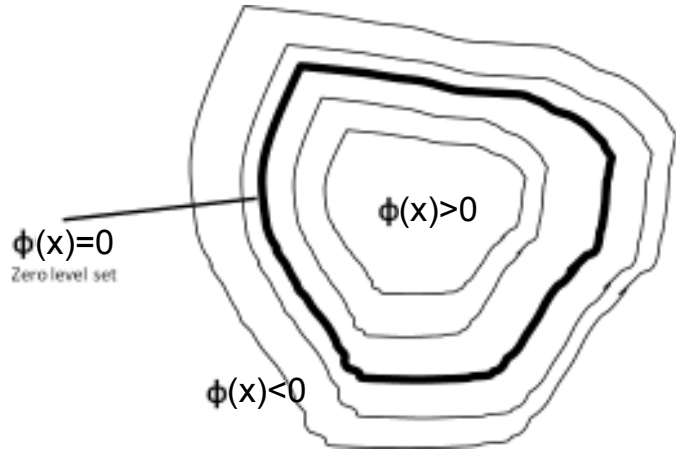

Figure 2.2: The geometric representation of a contour.

In the variational level set method, expressing the contour C in terms of the level set function $\phi$, the energy functional to be minimized can be expressed as follows:

$$E(\phi) = E_{in}(\phi(x)) + E_{out}(\phi(x)) + E_C(\phi(x)), \qquad (2.4)$$

where $E_{in}(\phi)$ and $E_{out}(\phi)$ are integral energy terms inside and outside the contour, and $E_C(\phi)$ is an integral energy term for the contour itself. More precisely, the three terms are defined as:

$$E_{in}(\phi(x)) = \int_{\phi(x)>0} e(x)dx = \int_{\Omega} H(\phi(x)) \cdot e(x)dx, \qquad (2.5)$$

$$E_{out}(\phi(x)) = \int_{\phi(x)<0} e(x)dx = \int_{\Omega} (1 - H(\phi(x))) \cdot e(x)dx, \qquad (2.6)$$

$$E_C(\phi(x)) = \int_{\Omega} \|\nabla H(\phi(x))\| dx = \int_{\Omega} \delta(\phi(x)) \cdot \|\nabla \phi(x)\| dx, \qquad (2.7)$$

where $e(x)$ is a suitable function, and $H$ and $\delta$ are, respectively, the Heaviside function and the Dirac delta distribution, i.e.,

$$H(z) = \begin{cases} 1, & \text{if } z \geq 0, \\ 0, & \text{if } z < 0, \end{cases} \qquad (2.8)$$

and

$$\delta(z) = \frac{d}{dz} H(z). \qquad (2.9)$$

Accordingly, the evolution of the level set function $\phi$ provides the evolution of the contour $C$. In the variational level set framework, the (local) minimization of the energy functional $E(\phi)$ can be obtained by evolving the level set function $\phi$ according to the following Euler-Lagrange partial differential equation[1]:

$$\frac{\partial \phi}{\partial t} = -\frac{\partial E(\phi)}{\partial \phi} \qquad (2.10)$$

---

[1] In the following, when writing partial differential equations, in general we do not write explicitly the arguments of the involved functions, which are described either in the text, or in the references from which such equations are reported.

where $\phi$ is now considered a function of both the pixel location $x$ and time $t$, and the term $\frac{\partial E(\phi)}{\partial \phi}$ denotes the functional derivative of $E$ with respect to $\phi$ (i.e., loosely speaking, the generalization of the gradient to an infinite-dimensional setting). So, Eq. (2.10) represents the application to the present optimization problem of an extension to infinite dimension of the classical gradient method for unconstrained optimization.

### 2.1.1 Unsupervised *ACM*s

According to specific kind of partial differential equation (*PDE*) (see Eq. 2.10) that models the contour evolution, variational level set methods can be divided into two categories: *Global Active Contour Models* (*GACMs*) [24, 33, 32, 118, 73], and *Local Active Contour Models* (*LACMs*) [40, 100, 31, 42].

In order to guide efficiently the evolution of the current contour, *ACM*s allow to integrate various kinds of information inside the energy functional, such as: local information (e.g., features based on spatial dependencies among pixels), global information (e.g., features that are not influenced by such spatial dependencies), shape information, prior information, and possibly also a-posteriori information learned from examples. As a consequence, depending on the kind of information used, one can further divide *ACM*s into several categories: e.g., edge-based *ACM*s [128, 54, 26, 52, 68, 111], global region-based *ACM*s [28, 91, 62, 112, 16], edge/region-based *ACM*s [29, 93, 119, 38, 107], and local region-based *ACM*s [110, 96, 106, 18, 126, 114].

In particular, edge-based ACMs make use of an edge-detector (in general, the gradient of the image intensity) to stop the evolution of the active contour on the true boundaries of the objects of interest. One of the most popular edge-based active contours is the Geodesic Active Contour (*GAC*) model [26], which is described in the following.

**Geodesic Active Contour (*GAC*) model** [26]. The level set formu-

lation of the *GAC* model can be described as follows:

$$\frac{\partial \phi}{\partial t} = g\|\nabla \phi\| \left( \text{div}\left(\frac{\nabla \phi}{\|\nabla \phi\|}\right) + \alpha \right) + \nabla g \cdot \nabla \phi, \quad (2.11)$$

where $\phi$ is the level set function, $\nabla$ is the gradient operator, $\alpha > 0$ is a balloon force term, and $g$ is the Edge Stopping Function (*ESF*) defined as follows:

$$g = \frac{1}{1 + \|\nabla G_\sigma * I\|^2}, \quad (2.12)$$

where $G_\sigma$ is a Gaussian kernel function with width $\sigma$, $*$ is the convolution operator, and $I$ is the original image intensity.

Edge-based models, such as the above-mentioned model, make use of an edge-detector, usually the gradient of the image intensity, to stop the evolution of the initial guess of the contour on the actual boundary. As a result, such kind of models can handle only images with well-defined edge information. Indeed, when images have ill-defined edges, the evolution of the contour typically does not converge to the true object boundary.

An alternative solution consists in using statistical information about a region (e.g., intensity, texture, color, etc.) to construct a stopping functional that is able to stop the contour evolution on the boundary between two different regions, as it happens in region-based models[2] [28, 91]. An example of a region-based model is illustrated as follows.

**Chan-Vese model (*C-V*)** [28]. The Chan-Vese (*C-V*) model is a well-known representative state-of-the-art global region-based *ACM* (at the time of writing, it has received more than 4000 citations, according to Scopus). After its initialization, the contour in the *C-V* model is evolved iteratively in an unsupervised fashion with the aim of minimizing a suitable energy functional, constructed in such a way that its minimum is achieved in correspondence with a close approximation of the actual boundary between two different regions. The energy functional $E_{CV}$ of the *C-V* model for a scalar-valued

---

[2]See the survey paper [58] for the recent state of the art region-based *ACM*s.

image has the expression

$$E_{CV}(C) := \mu \cdot \text{Length}(C) + v \cdot \text{Area}(\text{in}(C))$$
$$+ \lambda^+ \int_{\text{in}(C)} (I(x) - c^+(C))^2 dx$$
$$+ \lambda^- \int_{\text{out}(C)} (I(x) - c^-(C))^2 dx, \quad (2.13)$$

where $C$ is a contour, $I(x) \in \mathbb{R}$ denotes the intensity of the image indexed by the pixel location $x$ in the image domain $\Omega$, $\mu \geq 0$ is a regularization parameter which controls the smoothness of the contour, in($C$) (foreground) and out($C$) (background) represent the regions inside and outside the contour, respectively, and $v \geq 0$ is another regularization parameter, which penalizes a large area of the foreground. Finally, $c^+(C)$ and $c^-(C)$, i.e.,

$$c^+(C) := \text{mean}(I(x)|x \in \text{in}(C)), \quad (2.14)$$

and

$$c^-(C) = \text{mean}(I(x)|x \in \text{out}(C)), \quad (2.15)$$

are the mean intensities of the foreground and the background, respectively, and $\lambda^+, \lambda^- \geq 0$ are parameters which control the influence of the two image energy terms $\int_{\text{in}(C)} (I(x) - c^+(C))^2 dx$ and $\int_{\text{out}(C)} (I(x) - c^-(C))^2 dx$, respectively, inside and outside the contour. The functional is constructed in such a way that, when the regions in($C$) and out($C$) are smooth and "match" the true foreground and the true background, respectively, $E_{CV}(C)$ reaches its minimum.

Following [125], in the variational level set formulation of (2.13), the contour $C$ is expressed as the zero level set of an auxiliary function $\phi : \Omega \to \mathbb{R}$:

$$C := \{x \in \Omega : \phi(x) = 0\}. \quad (2.16)$$

Note that different functions $\phi(x)$ can be chosen to express the same contour $C$. For instance, denoting by $d(x, C)$ the minimum of the Euclidean distances of the pixel $x$ to the points on the curve $C$, $\phi(x)$ can be chosen as a signed distance function, defined as follows:

$$\phi(x) := \begin{cases} d(x, C), & x \in \text{in}(C), \\ 0, & x \in C, \\ -d(x, C), & x \in \text{out}(C), \end{cases} \quad (2.17)$$

This variational level set formulation has the advantage of being able to deal directly with the case of a foreground and a background that are not necessarily connected internally.

After replacing $C$ with $\phi$ and highlighting the dependence of $c^+(C)$ and $c^-(C)$ on $\phi$, in the variational level set formulation of the *C-V* model the (local) minimization of the cost (2.13) is performed by applying the gradient-descent technique in an infinite-dimensional setting (see Eq. 2.10 and also the reference [28]), leading to the following *PDE*, which describes the evolution of the contour:

$$\frac{\partial \phi}{\partial t} = \delta(\phi) \left[ \mu \nabla \cdot \left( \nabla \phi / \|\nabla \phi\| \right) - v - \lambda^+ \left( I - c^+(\phi) \right)^2 \right. \\ \left. + \lambda^- \left( I - c^-(\phi) \right)^2 \right], \quad (2.18)$$

where $\delta(\cdot)$ is the Dirac generalized function. The first term in $\mu$ of (2.18) keeps the level set function smooth, the second one in $v$ controls the propagation speed of the evolving contour, while the third and fourth terms in $\lambda^+$ and $\lambda^-$ can be interpreted, respectively, as internal and external forces that drive the contour toward the actual object boundary. Then, Eq. (2.18) is solved iteratively in [28] by replacing the Dirac delta by a smooth approximation, and using a finite difference scheme. Sometimes, also a re-initialization step is performed, in which the current level set function $\phi$ is replaced by its binarization (ie., a level set function of the form (2.17), representing the same current contour).

The *C-V* model can also be derived, in a Maximum Likelihood setting, by making the assumption that the foreground and the background follow Gaussian intensity distributions with the same variance [30]. Then, the model approximates globally the foreground and background intensity distributions by the two scalars $c^+(\phi)$ and $c^-(\phi)$, respectively, which are their mean intensities. Similarly, Leventon et al. proposed in [60] to use Gaussian intensity distributions with different variances inside a parametric density estimation method. Also, Tsai et al. in [97] proposed to use instead uniform intensity distributions to model the two intensity distributions. However, such models are known to perform poorly in the case of objects with inhomogeneous intensities [30].

Compared to edge-based models, region-based models usually perform better in images with blurred edges, and are less sensitive to the contour initialization.

Hybrid models that combine the advantages of both edge and regional information are able to control better the direction of evolution of the contour than the previous mentioned models. The Geodesic-Aided Chan-Vese (*GACV*) model [29] is a popular hybrid model, which includes both region and edge information in its formulation. An example of a hybrid model is the following.

**Selective Binary and Gaussian Filtering Regularized (*SBGFRLS*) Model**. The *SBGFRLS* model [124] combines the advantages of both the *C-V* and *GAC* models. It utilizes the statistical information inside and outside the contour to construct a region-based signed pressure force (*SPF*) function, which is used in place of the edge stopping function (*ESF*) (i.e., the information related to image intensity gradients) used in the *GAC* model (recall Eq. 2.12). Its level set formulation can be described as

$$\frac{\partial \phi}{\partial t} = spf(I(x)) \cdot \alpha \|\nabla \phi\|, \qquad (2.19)$$

where $\alpha$ is a balloon force parameter (controlling the rate of expansion of the level set function) and the function spf is defined as

$$spf(I(x)) = \frac{I(x) - \frac{c^+(C)+c^-(C)}{2}}{\max_{x \in \Omega} \left(\|I(x) - \frac{c^+(C)+c^-(C)}{2}\|\right)}, \qquad (2.20)$$

where $c^+(C)$ and $c^-(C)$ are defined likewise in the *C-V* model above.

Observe that compared to the *C-V* model, in Eq.( 2.18) the Dirac function term $\delta(\phi)$ has been replaced by $\|\nabla\phi\|$ which according to the authors, has an effective range on the whole image, rather than the small range of the former. Also, the bracket in Eq.( 2.18) is replaced by the *spf* function defined in Eq. 2.20. To regularize the curve the authors in [124] (following the practice of others, e.g., [128, 124, 87]), rather than relying on the computationally costly $\mu\nabla\cdot(\nabla\phi/\|\nabla\phi\|)$ term, convolve the level set curve with a Gaussian kernel $g_\sigma$, i.e.,

$$\phi \leftarrow g_\sigma * \phi, \qquad (2.21)$$

where the width $\sigma$ of the Gaussian $K_\sigma$ has a role similar to the one of $\mu$ in Eq.( 2.18) of the *C-V* model. If the value of $\sigma$ is small, then the level set function is sensitive to the noise and it does not allow the level set function to flow into the narrow regions of the object.

Overall this model is faster, computationally more efficient, and performs better than the conventional *C-V* model as pointed out [124]. However, it still has similar drawbacks as the *C-V* model, such as its inefficiency in handling images with several intensity levels, its sensitivity to the contour initialization, and its inability to handle images with intensity inhomogeneity (i.e., the effect of slow variations in object illumination possibly occurring during the image acquisition process).

In order to deal with images with intensity inhomogeneity, several authors have introduced in the *SPF* function terms that relate to local and global intensity information [109, 110, 96, 106]. However, these models are still sensitive to contour initialization and additive noise. Furthermore, when the contour is close to the object boundary, the influence of the global intensity force may distract the contour from the real object boundary, leading to object leaking [64], i.e., the presence of a final blurred contour.

In general, global models cannot segment successfully objects that are constituted by more than one intensity class. On the other hand, sometimes this is possible by using local models, which rely on local information as their main component in the associated variational level set framework. However, such models are still sensitive to the contour initialization and may lead to ob-ject leaking. Some examples of such local region-based *ACMs* are illustrated in the following.

**Local Binary Fitting (*LBF*) model** [61]. The evolution of the contour in the *LBF* model is described by the following *PDE*:

$$\frac{\partial \phi}{\partial t} = -\delta_\epsilon(\phi)(\lambda_1 e_1 - \lambda_2 e_2) + v\delta_\epsilon(\phi) div\left(\frac{\nabla \phi}{\|\nabla \phi\|}\right)$$
$$+ \mu\left(\nabla^2 \phi - div\left(\frac{\nabla \phi}{\|\nabla \phi\|}\right)\right) \quad (2.22)$$

where $v$ and $\mu$ are non-negative constants, $\epsilon > 0$, and the functions $e_1$ and $e_2$ are defined as follows:

$$e_1(x) = \int_\Omega g_\sigma(x-y) \|I(y) - f_1(x)\|^2 dy, \quad (2.23)$$

$$e_2(x) = \int_\Omega g_\sigma(x-y) \|I(y) - f_2(x)\|^2 dy, \quad (2.24)$$

where $f_1$ and $f_2$ are, respectively, internal and external gray-level fitting functions at point $x$ and $g_\sigma(x)$ is the kernel function of width $\sigma$. Also, $\delta_\epsilon(\phi)$, is a regularized Dirac function, defined as follows:

$$\delta_\epsilon(x) = \frac{1}{\pi} \frac{\epsilon}{\epsilon^2 + x^2}, \quad (2.25)$$

Finally, div is the divergence operator, whereas the functions $f_1$ and $f_2$ are defined as follows:

$$f_1(x) = \frac{g_\sigma(x)\left[H(\phi(x))I(x)\right]}{g_\sigma(x)H(\phi(x))}, \quad (2.26)$$

$$f_2(x) = \frac{g_\sigma(x)\left[(1 - H(\phi(x)))I(x)\right]}{g_\sigma(x)(1 - H(\phi(x)))}. \quad (2.27)$$

In general, the *LBF* model can produce good segmentations of objects with intensity inhomogeneities. Furthermore, it has a better performance than the well-known Piecewise Smooth (*PS*) model ([105] , [98]) for what concerns segmentation accuracy and computational efficiency. However, the *LBF* model only takes into account the local gray-level information. Thus in this model, it is easy to be trapped into a local minimum of the energy functional, and the model is also sensitive to the initial location of the active contour. Finally, over-segmentation problems may occur.

**Local Image Fitting (*LIF*) energy model** [122]. K. Zhang et al. proposed in [122] the *LIF* energy model to extract local image information in their proposed energy functional. The evolution of the contour in their model can be described by the following *PDE*:

$$\frac{\partial \phi}{\partial t} = \left(I - I^{LFI}\right)(m_1 - m_2)\,\delta_\epsilon\left(\phi\right) \tag{2.28}$$

where the local fitted image *LFI* is defined as follows:

$$I^{LFI} = m_1 H_\epsilon\left(\phi\right) + m_2\left(1 - H_\epsilon\left(\phi\right)\right) \tag{2.29}$$

where $m_1$ and $m_2$ are the average local intensities inside and outside the contour, respectively.

The main idea of this model is to use the local image information to construct a functional, which takes into account the difference between the fitted image and the original one to segment an image with intensity inhomogeneities.

The complexity analysis and experimental results showed that the *LIF* model is more efficient than the *LBF* model, while yielding similar results.

However, the obtained models are still sensitive to contour initialization and high levels of additive noise. A model that has been shown high accuracy when handling images with intensity inhomogeneity compared to the two above-mentioned models is the following one.

**Local Region-based Chan-Vese (*LRCV*)** [64]. The *LRCV* model is a natural extension of the already-mentioned Chan-Vese (*CV*) model. Such an extension is obtained by integrating local intensity information into the objective functional. This is the main feature of the *LRCV* model, which provides to it the capability of handling images with intensity inhomogeneity, which is missing instead in the *C-V* model.

The objective functional $E_{LRCV}$ of the *LRCV* model has the expression

$$E_{LRCV}(C) := \lambda^+ \int_{in(C)} (I(x) - c^+(x,C))^2 dx$$
$$+ \lambda^- \int_{out(C)} (I(x) - c^-(x,C))^2 dx, \qquad (2.30)$$

where $c^+(x,C)$ and $c^-(x,C)$ are functions which represent the local weighted mean intensities of the image around the pixel $x$, assuming that it belongs, respectively, to the foreground/background:

$$c^+(x,C) := \frac{\int_{in(C)} g_\sigma(x-y) I(y) \, dy}{\int_{in(C)} g_\sigma(x-y) dy}, \qquad (2.31)$$

$$c^-(x,C) := \frac{\int_{out(C)} g_\sigma(x-y) I(y) \, dy}{\int_{out(C)} g_\sigma(x-y) dy}, \qquad (2.32)$$

where $g_\sigma$ is a Gaussian kernel function with $\int_{\mathbb{R}^2} g_\sigma(x) dx = 1$ and width $\sigma > 0$.

Following the same procedures used in the *C-V* model results in the following *PDE*, which describes the evolution of the contour:

$$\frac{\partial \phi}{\partial t} = \delta(\phi) \left[ -\lambda^+ \left(I - c^+(x,\phi)\right)^2 + \lambda^- \left(I - c^-(x,\phi)\right)^2 \right], \qquad (2.33)$$

Eq. (2.33) can be solved iteratively by replacing the Dirac delta by a smooth approximation, and using a finite difference scheme. Also a regularization step can be performed (as in [64]), in which the current level set function $\phi$ is replaced by its convolution by a Gaussian kernel with suitable width $\sigma' > 0$.

A drawback of the *LRCV* model is that it relies only on the local information coming from the current location of the contour,

so it is sensitive to the contour initialization.

### 2.1.2 Supervised *ACM*s

From a machine learning perspective, *ACM*s for image segmentation use both supervised and unsupervised information. Both kinds of *ACM*s rely on parametric and/or nonparametric density estimation methods to approximate the intensity distributions of the subsets to be segmented (e.g., foreground/background). The main idea of such models is to make statistical assumptions on the image intensity distribution and to solve the segmentation problem by a Maximum Likelihood (*ML*) or Maximum A-Posteriori probability (*MAP*) approach. For instance, for scalar-valued images, in both parametric/nonparametric region-based *ACM*s, the objective energy functional has usually an integral form (see, e.g., [59]), whose integrands are expressed in terms of functions $e_i(x)$ having the form

$$e_i(x) := -\log(p_i(I(x))), \forall i \in \mathcal{I}. \tag{2.34}$$

Here, $p_i(I(x)) := p(I(x)|x \in \Omega_i)$ is the conditional probability density of the image intensity $I(x)$, conditioned on $x \in \Omega_i$, so the log-likelihood term $\log(p_i(I(x)))$ quantifies how much an image pixel is likely to be an element of the subset $\Omega_i$. In the case of supervised *ACM*s, the models $p_i(I(x))$ are estimated from a training set, one for each subset $\Omega_i$. Similarly, for a vector-valued image $\mathbf{I}(x)$ with $D$ components, the terms $e_i(x)$ have the form

$$e_i(x) := -\log(p_i(\mathbf{I}(x))), \forall i \in \mathcal{I}, \tag{2.35}$$

where $p_i(\mathbf{I}(x)) := p(\mathbf{I}(x)|x \in \Omega_i)$.

Now, we briefly discuss some supervised *ACM*s, which take advantage of the availability of labeled training data. As an example, Lee et al. proposed in [59] a supervised *ACM*, which is formulated in a parametric form. In the following, we refer to such a model as a Gaussian Mixture Model (*GMM*)-based *ACM*, since it exploits supervised training examples to estimate the parameters of multivariate Gaussian mixture densities. In such a model, the level set evolution *PDE* is given, e.g., in the case of multi-spectral images $\mathbf{I}(x)$, by

$$\frac{\partial \phi}{\partial t} = \delta(\phi)\left[\beta\kappa(\phi) + \log p_{\text{in}}(\mathbf{I}) - \log p_{\text{out}}(\mathbf{I})\right], \qquad (2.36)$$

where $\beta \geq 0$ is a regularization parameter, and $\kappa(\phi)$ is the average curvature of the level set function $\phi$. The two terms $p_{\text{in}}(\mathbf{I}(x))$ and $p_{\text{out}}(\mathbf{I}(x))$ in (2.36) are then expressed as

$$p_{\text{in}}(\mathbf{I}(x)), p_{\text{out}}(\mathbf{I}(x)) := \sum_{k=1}^{K} \alpha_k \mathcal{N}(\mu_k, \Sigma_k, \mathbf{I}(x)), \qquad (2.37)$$

where $K$ is the number of computational units, $\mathcal{N}(\mu_k, \Sigma_k, \cdot)$, $k = 1, \ldots, K$ are Gaussian functions with centers $\mu_k$ and covariance matrices $\Sigma_k$, and the $\alpha_k$'s are the coefficients of the linear combination. All the parameters ($\alpha_k$, $\mu_k$, $\Sigma_k$) are then estimated from the training examples. Besides *GMM*-based *ACMs*, also Nonparametric Kernel Density Estimation (*KDE*)-based models with Gaussian computational units have been proposed in [34, 35] with the same aim. In the case of scalar images, they have the form

$$p_{\text{in}}(I(x)) := \frac{1}{\|L^+\|} \sum_{i=1}^{\|L^+\|} \mathcal{K}\left(\frac{I(x) - I(x_i^+)}{\sigma}\right), \qquad (2.38)$$

$$p_{\text{out}}(I(x)) := \frac{1}{\|L^-\|} \sum_{i=1}^{\|L^-\|} \mathcal{K}\left(\frac{I(x) - I(x_i^-)}{\sigma_{KDE}}\right), \qquad (2.39)$$

where the pixels $x_i^+$ and $x_i^-$ belong, respectively, to given sets $L^+$ and $L^-$ of training pixels inside the true foreground/background, $\sigma_{KDE} > 0$ is the width of the Gaussian kernel used in the *KDE*-based model, and

$$\mathcal{K}(u) := \frac{1}{\sqrt{2\pi}} \exp\left(-\frac{u^2}{2}\right). \qquad (2.40)$$

Of course, such models can be extended to the case of vector-valued images (in particular, replacing $\sigma_{KDE}^2$ by a covariance matrix).

### 2.1.3 Other level set-based *ACM*s

**Supervised Boundary-based** *GAC* **(*sBGAC*)** [79]. The *sBGAC* model is a supervised level-set based *ACM*, which was proposed by Paragios et al. in [79] with the aim of providing a boundary-based framework that is derived by the *GAC* for texture image segmentation. Its main contribution is the connection between the minimization of a *GAC* objective function with a contour propagation method for supervised texture segmentation. However, *sBGAC* is still limited to boundary-based information, which results in a high sensitivity to the noise and to the initial contour.

**Geodesic Active Region Model (*GARM*)** [80]. *GARM* was proposed in [80] with the aim of reducing the sensitivity of *sBGAC* to the noise and to the contour initialization, by integrating the region-based information along with the boundary information. *GARM* is a supervised texture segmentation *ACM* implemented by variational level set.

    The inclusion of supervised examples in *ACM*s can improve significantly their performance by constructing a Knowledge Base (*KB*), to be used as a guide in the evolution of the contour. However, state-of-the-art supervised *ACM*s often make strong statistical assumptions on the image intensity distribution of each subset to be modeled. So, the evolution of the contour is driven by probability models constructed based on given reference distributions. Therefore, the applicability of such models is limited by how accurate the probability models are.

# Chapter 3

# SOM-based *ACM*s

Self Organizing Maps (*SOM*s) have attracted the attention of many computer vision scientists, particularly when dealing with image segmentation as a contour extraction problem. The idea of utilizing the prototypes (weights) of a *SOM* to model an evolving contour has produced a new class of Active Contour Models (*ACM*s), known as *SOM*-based *ACM*s. Such models have been proposed in general with the aim of exploiting the specific ability of *SOM*s to learn the edge-map information via their topology preservation property, and overcoming some drawbacks of other *ACM*s, such as trapping into local minima of the image energy functional to be minimized in such models. In this chapter (in a similar way to the previous chapter), the main principles of *SOM*s and their application in modeling active contours are highlighted. Then, we review existing *SOM*-based *ACM*s with a focus on their advantages and disadvantages in modeling the evolving contour via different kinds of *SOM*s.

## 3.1  Self Organizing Maps (*SOM*s)

The *SOM* [55, 56], which was proposed by Kohonen, is an unsupervised neural network whose neurons update concurrently their weights in a self-organizing manner, in such a way that, during the learning process, its neurons evolve adaptively into specific detectors of different input patterns. A basic *SOM*, see Fig. 3.1, is composed of an input layer, an output layer, and an intermediate connection layer. The input layer contains a unit for each component of the input vector. The output layer consists of neurons that are typically located either on a 1-*D* or a 2-*D* grid, and are fully connected with the units in the input layer. The intermediate connection layer is composed of weights (also called prototypes) connecting the units in the input layer and the neurons in the output layer (in practice, one has one weight vector associated with each output neuron, where the dimension of the weight vector is equal to the dimension of the input). The learning algorithm of the *SOM* can be summarized by the following steps:

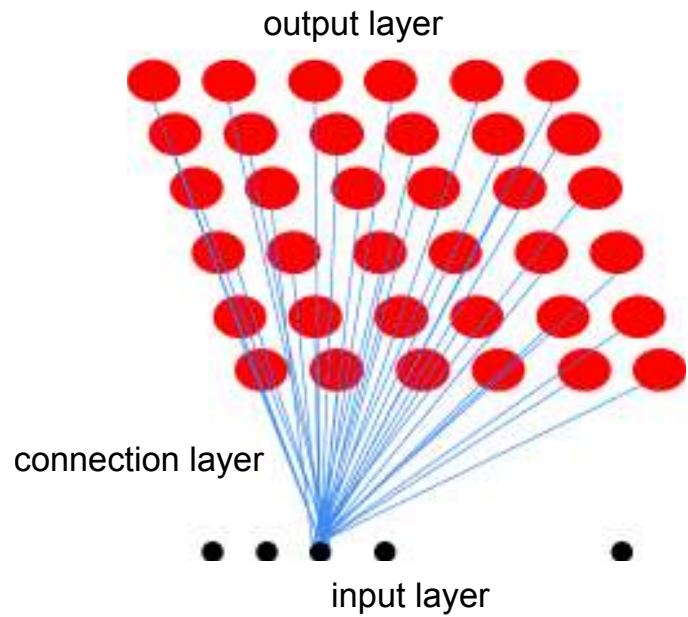

Figure 3.1: The SOM architecture.

1. initialize randomly the weights of the neurons in the output layer, and select a suitable learning rate and a suitable neighborhood size around a "winner" neuron;

2. for each training input vector, determine the winner neuron

using a suitable rule;

3. update the weights on the selected neighborhood of the winner neuron;

4. repeat Steps 2-3 above by selecting another training input vector, until learning is accomplished (i.e., a suitable stopping criterion is satisfied).

Given a collection of training samples and a number of classes, *SOM*s can be also used as a concurrent system for pattern classification, hence for image segmentation. In this context, a specific model is represented by the Concurrent Self Organizing Maps (*CSOM*s) [74]. The classification process of a *CSOM* starts by training a series of *SOM*s (one for each class) in a parallel way, using for each *SOM* a subset of samples coming from its associated class.

During the training process, the neurons of each *SOM* are topologically arranged in the corresponding map on the basis of their prototypes (weights) and of the ones of the neurons within a certain geometric distance from them, and are moved toward the current input using the classical self-organization learning rule of a *SOM*, which is expressed by

$$w_n(t+1) := w_n(t) + \eta(t)h_{bn}(t)[x(t) - w_n(t)], \tag{3.1}$$

where $t = 0, 1, 2, 3, \ldots$ is a time index, $w_n(t)$ is the prototype of the neuron $n$ at time $t$, $x(t)$ is the input vector at time $t$, $\eta(t)$ is a learning rate, and $h_{bn}(t)$ is the neighborhood kernel at time $t$ of the neuron $n$ around a specific neuron $b$, called best-matching unit (*BMU*). More precisely, in each *SOM* and at the time $t$, an input vector $x(t) \in \mathbb{R}^D$ is presented to feed the network, then the neurons in the map compete one with the other to be the winner neuron $b$, which is the chosen as the one whose weight $w_b(t)$ is the closest to the input vector $x(t)$ in terms of a similarity measure, which is usually the Euclidean distance $\|\cdot\|_2$. In this case, $\|x(t) - w_b(t)\|_2 := \min_n \|x(t) - w_n(t)\|_2$, where $n$ varies in the set of neurons of the map.

Once the learning of all the *SOM*s has been accomplished, the class label of a previously-unseen input test pattern is deter-mined by the criterion of the minimum quantization error. More precisely, the *BMU* neuron associated with the input test pattern is determined for each *SOM*, and the winning *SOM* is the one for which the prototype of the associated *BMU* neuron has the smallest distance from the input test pattern, which is consequently assigned to the class associated with that *SOM*. *SOM*s have been used extensively for image segmentation, but often not in combination with *ACM*s [99, 89, 76, 75, 95, 17, 127, 92, 85, 53]. In the following subsection, we review, in brief, some of the existing *SOM*-based segmentation models.

## 3.2 *SOM*-based Segmentation Models

In [65], a *SOM*-based clustering technique has been used as a thresholding technique for image segmentation. First, the intensity histogram of the image was used to feed a *SOM* in order to partition the histogram into several regions. The algorithm was applied to text recognition, after choosing the number of regions in a suitable way.

Huang et al. in [45] proposed to use a *SOM* in two stages, for color image segmentation. The first stage aims to identify a large initial set of color classes, while the second ones aims to identify a final batch of segmented clusters.

In [47], Y. Jiang et al. used a *SOM* to segment a multi-spectral image (composed of five-dimension vectors), by clustering the pixels based on their color and spatial features. Then, those clustered blocks were merged into a specific number of regions, and some morphological operations were applied. In general *SOM*s, have been extensively used in the field of segmentation and all the developed *SOM*-based segmentation models (as stated in [37, 117, 14, 90, 51, 92]) yielded improved segmentations in comparison to traditional non *SOM*- based techniques.

Moreover, also other kinds of neural networks have been used with the aim of approximating the edge map: e.g., multilayer perceptrons [71], whose approximation capability has been extensively investigated (see, e.g., [19, 57]).

We conclude mentioning that, when a *SOM* is used as a supervised/unsupervised image segmentation technique, the segmented objects so-obtained have usually disconnected boundaries, which are often sensitive to the noise. However, in order to improve the robustness of edge-based *ACM*s to the blur and to ill-defined edge information, *SOM*s have been also used in combination with *ACM*s, with the explicit aim of modelling the active contour and controlling its evolution, adopting a learning scheme similar to Kohonen's learning algorithm [55], resulting in *SOM*-based *ACM*s [104, 86] (which belong, in this case, to the class of edge-based *ACM*s). The evolution of the active contour in a *SOM*-based *ACM* is guided by the feature space constructed by the *SOM* when learning the weights associated with the neurons of the map. A review of *SOM*-based *ACM*s is provided in the following subsections.

### 3.2.1 An example of a *SOM*-based *ACM*

The basic idea of existing *SOM*-based *ACM*s is to model and implement the active contour using a *SOM* neural map, relying in the training phase on the edge map of the image (i.e., the set of points obtained by an edge-detection algorithm) to update the weights of the neurons of the *SOM* network, and consequently to control the evolution of the active contour. The points of the edge map act as inputs to the network, which is trained in an unsupervised fashion (in the sense that no supervised samples belonging to the foreground/background, respectively, are provided). As a result, during training the weights associated with the neurons in the output map move toward points belonging to the nearest salient contour.

In the following, we illustrate the general ideas of using a *SOM* in modeling the active contour, by describing a classical example of a *SOM*-based *ACM*, which was proposed in [104] by Venkatesh and Rishikesh.

**Spatial Isomorphism Self Organizing Map (*SISOM*)-based *ACM*** [104]. The *SISOM*-based *ACM* is the first *SOM*-based *ACM*

which appeared in the literature. It was proposed with the aim of localizing the salient contours in an image using a *SOM* to model the evolving contour. The *SOM* is composed of a fixed number of neurons (and consequently a fixed number of "knots" or control points for the evolving curve) and has a fixed structure. The model requires a rough approximation of the true boundary as an initial contour. Its *SOM* network is constructed and trained in an unsupervised fashion, based on the initial contour and the edge map information. The contour evolution is controlled by the edge information extracted from the image by an edge detector. As Fig. 3.2 illustrates, the main steps of the *SISOM*-based *ACM* can be summarized as follows:

1. construct the edge map of the image to be segmented;

2. initialize the contour to enclose the object of interest in the image;

3. obtain the $x_1$- and $x_2$- coordinates of the edge points to be presented as inputs to the network;

4. construct a *SOM* with a number of neurons equal to the number of the edge points of the initial contour and two weights associated with each neuron; the points on the initial contour are used to initialize the weights of the *SOM*;

5. repeat the following steps for a fixed number of iterations:

    (a) select randomly an edge point and feed its coordinates to the network;

    (b) determine the best-matching neuron;

    (c) update the weights of the neurons in the network by the classical unsupervised learning scheme of the *SOM* [55], which is composed of a competitive phase and a cooperative one;

    (d) compute a neighborhood parameter for the contour according to the updated weights and a threshold.

Fig. 3.2 illustrates the evolution procedure of the *SISOM*-based *ACM*. On the left-side of the figure, the neurons of the map are represented by gray circles, while the black circle represents the winner neuron associated with the current input to the map (in this case, the red circle on the right-hand side of the figure, which is connected by the blue segments to all the neurons of the map). On the right-hand side, instead, the positions of the white circles represent the initial prototypes of the neurons, whereas the positions of the black circles represent their final values, at the end of learning. The evolution of the contour is controlled by the learning algorithm above, which guides the evolution of the protoypes of the neurons of the *SOM* (hence, of the active contour) using the points of the edge map as inputs to the *SOM* learning algorithm. As a result, the final contour is represented by a series of prototypes of neurons located near the actual boundary of the object to be segmented.

We conclude by mentoning that, in order to produce good segmentations, the *SISOM*-based *ACM* requires the initial contour (which is used to initialize the prototypes of the neurons) to be very close to the true boundary of the object to be extracted, and the points of the initial contour have to be assigned to the neurons of the *SOM* in a suitable order: if such assumptions are satisfied, the contour extraction process performed by the model is robust to the noise. Moreover, differently from other *ACM*s, the model does not require a particular energy functional to be optimized.

### 3.2.2 Other *SOM*-based *ACM*s

In this subsection, we describe other *SOM*-based *ACM*s, and highlight their advantages and disadvantages.

**Time Adaptive Self Organizing Map (*TASOM*)-based *ACM*** [86]. The *TASOM*-based *ACM* was proposed by Shah-Hosseini and Safabakhsh as a development of the *SISOM*-based *ACM*, with the aim of inserting neurons incrementally into the *SOM* map or deleting them incrementally, thus determining automatically the required number of control points of the extracted contour. More-over, each neuron is provided with its specific dynamic learning rate and neighbourhood function. As a consequence, the *TASOM*-based *ACM* can overcome one of the main limitations of the *SISOM*-based *ACM*, i.e., its sensitivity to the contour initialization, in the sense that the initial guess of the contour in the *TASOM*-based *ACM* can be far from the actual object boundary. Likewise the *SISOM*-based *ACM*, topological changes of the objects (e.g., splitting and merging) cannot be handled, since both models rely completely on the edge information (instead than on regional information) to drive the contour evolution.

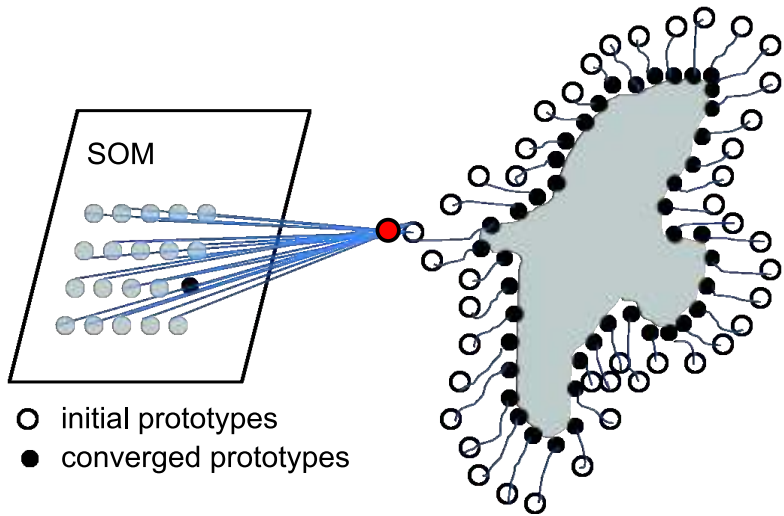

Figure 3.2: The architecture of the *SISOM*-based *ACM* proposed in [104].

**Batch Self Organizing Map (*BSOM*)-based *ACM*** [102, 103]. This model is a modification of the *TASOM*-based *ACM*, and was proposed by Venkatesh et al. with the aim of dealing better with the leaking problem (i.e., the presence of a final blurred contour), which often occurs when handling images with ill-defined edges. Such a problem is due to the explicit usage by the *TASOM*-based *ACM* of only edge information to model and control the evolution of the contour. In the *BSOM*-based *ACM*, instead, the image intensity variation inside a local region is used along with the edge information to control the movement of the contour. In this way, the robustness of the model is increased in handling images with blurred edges. At the same time, the *BSOM*-based *ACM* is less sensitive to the initial guess of the contour, when compared to parametrized *ACM*s like Snakes, and to the *SOM*-based *ACM*s described above. However, likewise all such models, the *BSOM*-based *ACM* has not the ability to handle topological changes of the objects to be segmented. An extension of the *BSOM*-based *ACM* was proposed in [101, 20] and applied therein to the segmentation of pupil images. Such a modified version of the basic *BSOM*-based *ACM* increases the smoothness of the extracted contour, and prevents the extracted contour from being extended over the true boundaries of the object.

**Fast Time Adaptive Self Organizing Map (*FTA-SOM*)-based *ACM*** [46]. This is another modification of the *TASOM*-based *ACM*, and was proposed by Izadi and Safabakhsh with the aim of decreasing its computational complexity. The *FTA-SOM*-based *ACM* is based on the observation that choosing the learning rate parameters of the prototypes of the neurons of the *SOM* in such a way that they are equal to a large fixed value when they are far from the boundary, and to a small value when they are near the boundary, can lead to a significant increase of the convergence speed of the active contour. Accordingly, in each iteration, the *FTA-SOM*-based *ACM* finds the minimum distance of each neuron from the boundary, then its sets the associated learning rate as a fraction of that distance.

**Coarse to Fine Boundary Location Self Organizing Map (*CFBL-SOM*)-based *ACM*** [121]. The above *SOM*-based *ACM*s work in an unsupervised fashion, as the user is required only to provide an initial contour to be evolved automatically. In [121], Zeng et al. proposed the *CFBL-SOM*-based *ACM* as the first supervised *SOM*-based *ACM*, i.e., a model in which the user is allowed to provide supervised points (supervised "seeds") from the desired boundaries. Starting from this coarse information, the neurons of the *SOM* are then employed to evolve the active contour to the desired boundaries in a "coarse-to-fine" approach. The *CFBL*-

*SOM*-based *ACM* follows such a strategy when controlling the evolution of the contour. So, an advantage of the *CFBL-SOM*-based *ACM* over the *SOM*-based *ACM*s described above is that it allows to integrate prior knowledge on the desired boundaries of the objects to be segmented, which comes from the interaction of the user with the *SOM*-based *ACM*s segmentation framework. When compared with such *SOM*-based *ACM*, this property provides the *CFBL-SOM*-based *ACM* with the ability of handling objects with more complex shapes, inhomogeneous intensity distributions, and weak boundaries.

**Conscience, Archiving and Mean-movement mechanisms Self Organizing Map (*CAM-SOM*)-based *ACM* [84].** The *CAM-SOM*-based *ACM* was proposed by Sadeghi et al. as an extension of the *BSOM*-base *ACM*, by introducing three mechanisms called Conscience, Archiving and Mean-Movement. The main achievement of the *CAM-SOM*-based *ACM* is to allow more complex boundaries (such as concave boundaries) to be captured, and to provide a reduction of the computational cost. By the Conscience mechanism, the neurons are not allowed to "win" too much frequently, which makes the capture of complex boundaries possible. The Archiving mechanism allows a significant reduction in the computational cost. By such mechanism, neurons whose prototypes are close to the boundary of the object to be segmented and whose values have not changed significantly in the last iterations are archived and eliminated from subsequent computations. Finally, in order to ensure a continuous movement of the active contour towards concave regions, the Mean-Movement mechanism is used in each epoch to force the winner neuron to move towards the mean of a set of feature points, instead of a single feature point. Together, the Conscience and Mean-Movement mechanisms prevent the contour from stopping the contour evolution at the entrance of object concavities.

**Extracting Multiple Objects**. The main limitation of various *SOM*-based *ACM*s is their inability to detect multiple contours and to recognize multiple objects. As mentioned above, a similar problem arises in parametric *ACM*s such as Snakes. To deal with the multi-ple contour extraction problem, Venkatesh et al. proposed in [103] to use a splitting criterion. However, if the initial contour is outside the objects, contours inside an object still cannot be extracted. Sadeghi et al. proposed in [84] a splitting criterion (to be checked at each epoch) such that the main contour can be divided into several sub-contours whenever the criterion is satisfied. The process is repeated until each of the sub-contours encloses one single object. However, the merging process is still not handled implicitly by the model, which reduces its scope, especially when handling images containing multiple objects in the presence of noise or ill-defined edges. Moreover, Ma et al. proposed in [66] to use a *SOM* to classify the edge elements in the image. This model relies first on detecting the boundaries of the objects. Then, for each edge pixel, a feature vector is extracted and normalized. Finally, a *SOM* is used as a clustering tool to detect the object boundaries when the feature vectors are supplied as inputs to the map. As a result, multiple contours can be recognized. However, the model shares the same limitations of other models that use a *SOM* as a clustering tool for image segmentation [99, 116], resulting in disconnected boundaries and sensitivity to the noise.

# Chapter 4

# Globally Signed Pressure Force Model

## 4.1 Introduction

One of the most popular and widely used global active contour models is the region-based *ACM*, which often relies on the assumption of homogeneous intensity in the regions of interest. As a result, most often than not, when images violate this assumption the performance of this method is limited. Thus, handling images that contain foreground objects characterized by multiple intensity classes present a challenge. In this chapter, we present a novel active contour model based on a new Signed Pressure Force (*SPF*) function which we term *Globally Signed Region Pressure Force* (*GSRPF*). It is designed to take into account, in a global way, of the skewness of the intensity distribution of the region of interest (*ROI*). It can accurately modulate the signs of the pressure force inside and outside the contour, and handle images with multiple intensity classes in the foreground. Moreover, it is robust to additive noise, and offers high efficiency and rapid convergence. The proposed *GSRPF* model is robust to contour initialization and has the ability to stop the curve evolution close even to ill-defined (weak) edges. *GSRPF* provides a nearly parameter-free segmentation environment, requiring minimal user intervention. Experimental re-sults on several synthetic and real images have demonstrated the higher accuracy of the segmentation results obtained by *GSRPF*, in comparison to the segmentations obtained by other methods adopted from the literature.

The majority of global intensity-based active contour models assume that the regions of interest are composed by subregions that are nearly homogeneous in intensity. Consequently, when these assumptions are violated, the performance of these models is far from the desired one. In this chapter, we propose the *GSRPF* as a new intensity-driven region-based *ACM* that can efficiently segment the foreground (i.e., the object(s)) when it is characterized by a non symmetric distribution. This non symmetry could arise either from intensity variations or from the fact that the object could be composed into two or more intensity classes. To provide a computationally efficient solution and reduce the possibility of trapping into local minima, the GSRPF model is based on an *SPF*-like formulation.

## 4.2 The *GSRPF* Model

It is obvious that relying only on the global mean (inside and outside the contour) as in the *C-V* model is not sufficient to model intensity distributions when the images to be segmented have foregrounds characterized by more complex intensity distributions. To overcome this problem, we introduce the global median in addition to a global mean inside the energy term to be minimized. Given a contour *C*, $x$ the pixel location in the image $I(x)$, the energy term is defined as

$$E_{GSRPF}(C, c^+, m^+, c^-) := \int_{in(C)} \lambda^+ e^+(x) dx$$
$$+ \int_{out(C)} 2\lambda^- e^-(x) dx, \quad (4.1)$$
$$e^+(x) := |I(x) - c^+|^2 + |I(x) - m^+|^2, \quad (4.2)$$
$$e^-(x) := |I(x) - c^-|^2, \quad (4.3)$$

where the positive constants $\lambda^+$ and $\lambda^-$ define the weight of each term (inside and outside the contour), $c^+$ and $m^+$ are scalars approximating the mean and median intensity respectively, for the image $I$ inside the contour, and $c^-$ is a scalar approximating the mean outside the contour. Following the standard variational level set formulations [28], we replace the contour curve $C$ with the level set function $\phi$ [125], thus obtaining

$$E_{GSRPF}(\phi, c^+, m^+, c^-) = \int_{\phi>0} \lambda^+ e^+(x) dx$$
$$+ \int_{\phi<0} 2\lambda^- e^-(x) dx. \quad (4.4)$$

In a similar way to other intensity-driven active contour models, the statistical descriptors $c^+$, $m^+$, and $c^-$ are defined

$$\begin{cases} c^+(\phi) = \text{mean}(I(x)|\phi(x) \geq 0), \\ m^+(\phi) = \text{median}(I(x)|\phi(x) \geq 0), \\ c^-(\phi) = \text{mean}(I(x)|\phi(x) < 0). \end{cases} \quad (4.5)$$

Using the level set function $\phi$ to represent the contour $C$ in the domain $\Omega$, the energy $E_{GSRPF}$ can be written as a functional as follows:

$$E_{GSRPF}\left(\phi, c^+, m^+, c^-\right) = \int_\Omega \lambda^+ e^+(x) H(\phi(x)) dx$$
$$+ \int_\Omega 2\lambda^- e^-(x)(1 - H(\phi(x))) dx, \qquad (4.6)$$

where $H$ is the Heaviside function.

By keeping $c^+$, $m^+$, and $c^-$ fixed, we minimize the energy functional $E_{GSRPF}\left(\phi, c^+, m^+, c^-\right)$ with respect to $\phi$ to obtain the gradient descent flow as

$$\frac{\partial \phi}{\partial t} = \delta\left(\phi\right)\left[-\lambda^+ e^+(x) + 2\lambda^- e^-(x)\right], \qquad (4.7)$$

where $\delta$ is the generalized Dirac delta function.

By considering the higher order statistics $m^+$, the proposed model can overcome the limitation of the *C-V* model about the symmetry of the intensity distribution, which is not accurate in most of the real-life images. In the binary gray level images, our model as an energy minimization model behaves exactly the same as the *C-V* model, where $m^+ = c^+$. However, in order to improve the robustness to the contour initialization when handling gray-level images, in the next subsection we include in the model an *SPF* function.

### 4.2.1 The *GSRPF* sign pressure function formulation

Although we could rely on Eq. 4.7 to update our level set, obtaining an "SPF" like formulation would reduce the possibility of trapping into local minima by well modulating the interior and exterior forces.

In this section, we propose one such formulation, which we term Globally Signed Region Pressure Force (*GSRPF*) model. It is proposed in such a way that it can modulate the signs of the pressure force inside and outside the object of interest using the statistical quantities defined in Eq. 4.5 to be combined with the minimization of an energy functional similar to the one in Eq. 4.6.

First, we assume $\lambda^+ = \lambda^- = 1$, then we define the *SPF* function as follows:

$$spf(I(x)) = spf_1 \cdot spf_2(I(x)), \tag{4.8}$$

where

$$\begin{cases} spf_1 := sign(2c^+ + 2m^+ - 4c^-), \\ spf_2(I(x)) := sign(I(x) - \frac{c^{+2}+m^{+2}-2c^{-2}}{2c^++2m^+-4c^-}), \end{cases} \tag{4.9}$$

where $c^+, m^+$, and $c^-$ are defined in Eq. 4.5.

Rather than a constant force, we use a force that is a quadratic function of $I(x)$ to control the propagation of the evolving curve, i.e., we define

$$\alpha(I(x)) := \left(I(x) - \frac{c^{+2} + m^{+2} - 2c^{-2}}{2c^+ + 2m^+ - 4c^-}\right)^2. \tag{4.10}$$

The motivation behind the proposed propagation function $\alpha(I(x))$ is to dynamically increase the interior and exterior forces acting on the curve when it is far from the boundaries (thus reducing such possibility of entrapment in local minimal) and decrease the forces when the curve is close to the boundaries (thus allowing the curve to stop very close to the actual boundaries).

The (per-pixel) multiplication of the proposed $\alpha(I(x))$ and $spf(I(x))$ results in a new region-based signed pressure force function, which we term Globally Signed Region Pressure Force (*GSRPF*) function:

$$gsrpf(I(x)) := \alpha(I(x)) \cdot spf(I(x)). \tag{4.11}$$

The proposed *GSRPF* function has the capacity to modulate the sign of the pressure forces and implicitly control the propagation of the evolving curve so that the contour shrinks when it is outside the object of interest and expands when it is inside the object.

Following the *spf* formulation described in Chapter 2, the final level set formulation of our model is described by the follow-

ing PDE:

$$\frac{\partial \phi}{\partial t} = gsrpf(I(x)) \cdot |\nabla \phi|. \quad (4.12)$$

In order to achieve computational efficiency, we use a Gaussian kernel to regularize the level set function $\phi$ to keep the interface regular. The parameter $\sigma$ of the Gaussian kernel is the only tunable parameter of the model.

As we will demonstrate in the Section 4.4 the proposed model:

- is capable of identifying objects of complex intensity distribution (by taking into account the skewness of the distribution in the model);

- is robust to additive noise (e.g., a higher order statistics is considered in our model to accommodate non symmetric and noisy distributions);

- is not sensitive to the contour initialization (since only global information is considered for the curve evolution);

- is computationally efficient (since it does not require a re-initialization of the level set function, and regularizes the contour efficiently); and

- requires a few iterations to converge.

## 4.3 Implementation

To illustrate the ease of implementation of our model, the main steps of the algorithm can be summarized as follows.

1. Initialize the level set function $\phi$ to be binary i.e., set

$$\phi(x, t = 0) := \begin{cases} -\rho & x \in \Omega_0 \setminus \Omega'_0, \\ 0 & x \in \Omega'_0, \\ \rho & x \in \Omega \setminus \Omega_0, \end{cases} \quad (4.13)$$

where $\rho > 0$ is a constant, $\Omega_0$ is a subset in the image domain $\Omega$ and $\Omega'_0$ is the boundary of $\Omega_0$.

2. Calculate the GSRPF function according to Eq. 4.11.

3. Evolve the level set according to Eq. 4.12.

4. Regularize the level set using a Gaussian kernel function.

5. If the curve evolution has converged, stop and return the result. Otherwise return to Step 2.

## 4.4 Experimental study

In this section we demonstrate the superiority of the proposed method, compared to implementations of some of the methods reviewed in Chapter 2, when applied to challenging synthetic and real images. We implemented the proposed algorithm in Matlab R2009b on a PC (2.5-GHz Intel(R) Core(TM) 2 Duo, 2.00 GB RAM). For a fair comparison, we used reference Matlab implementations of the *C-V* and *SBGFRLS*.

To demonstrate the effectiveness of our approach in handling images where the background has multiple intensity classes, we created a synthetic image for this purpose (shown in Fig. 4.1), without additive noise and with noise. We compare the performance of the proposed model with the *C-V* and *SBGFRLS* models, and vary the parameter $\sigma$. As Fig. 4.1(a) illustrates, by increasing the value of $\sigma$, the proposed *GSRPF* is not sensitive to the noise and finds all the regions of the object for a large rang of $\sigma$. On the other hand, the *SBGFRLS* model (see Fig. 4.1(d)) is not able to evolve the contour properly through the noisy regions, even when altering the values of $\alpha$ and $\sigma$. Similarly, as Fig. 4.1(e) shows, the $C-V$ model is unable to segment the image, even when considering $\mu$ values.

To demonstrate the accuracy of the proposed method quantitatively, we adopt the precision and recall metrics, and compare the segmentation results with the ground truth. Fig. 4.2 shows the effect of $\sigma$ on the accuracy of the segmentation result using the synthetic image with noise shown in Fig. 4.1(a) as the ground truth. Based on the results of this experiment, the value $\sigma = 1.4$ is recommended to handle noisy images with multiple classes in the foreground.

Table 4.1 shows the robustness of our model when different levels of noise are added to the synthetic image of Fig. 4.1. The high precision of the obtained segmentations at most noise levels confirms the ability of the proposed *GSRPF* model to find all the regions of the object, irrespective of noise strength.

Table 4.1: The robustness of the *GSRPF* model ($\sigma = 1.4$) to the noise level: Precision and Recall metrics for different Gaussian noise levels, measured by the standard deviation (*SD*).

| *SD* | 10 | 20 | 30 | 40 | 50 |
|---|---|---|---|---|---|
| Precision(%) | 100 | 100 | 100 | 99 | 89 |
| Recall (%) | 10 | 99 | 89 | 80 | 71 |

Fig. 4.3(b) illustrates the ability of the *GSRPF* model to find accurately the boundaries of objects with various convexities, shapes, and noisy background. Even though *SBGFRLS* can also identify the objects, it is unable to segment the hole inside one of the objects, as shown in Fig. 4.3(c). The *C-V* model is unable to segment the same image (as shown in Fig. 4.3(d)) because it is trapped into a local minimum.

To demonstrate the speed and adaptability of the proposed method, in Fig. 4.4 we show the curve evolution for a few iterations. It is readily evident that our model converges fast to an accurate delineation of the foreground object.

Fig. 4.5 shows the robustness of the proposed *GSRPF* model and also the sensitivity of the *SBGFRLS* and *C-V* models to different contour initializations. The interior and exterior forces are able to guide efficiently the evolution of the contour, nonwithstanding the location of the initial contour. Indeed, the initial position of the contour does not affect the final segmentation, as Fig. 4.5(b), (f), (j), and (n) show, and the presence of the shadow of the plane does not lead to over-segmentation. On the other hand, the *SBGFRLS* model is unable to accurately segment the object when the contour is initialized outside the object, as shown in Fig. 4.5(g), (k), and (o). On the other hand, the *C-V* model is more robust to the initialization compared to *SBGFRLS*, with the exception of Fig. 4.5(p).

Fig. 4.6 demonstrates the ability of our method in handling images arising in natural and life sciences. In Fig. 4.6(a), all the models accurately delineate the boundaries of a brain malignancy. Fig. 4.6(b) shows the ability of our model to extract accurately an Arabidopsis rosette from a complicated background (e.g. soil, pot, tray); however, the other two models are not able to extract all the plant parts, as seen in Fig. 4.6(c) and 4.6(d). Similarly, Fig. 4.6(c) and (d) show the ability of *GSRPF* to segment multiple objects in the scene, such as cells and chromosomes. On the other hand, the segmentation results of the *SBGFRLS* and *C-V* models are not satisfactory. This is mainly to be attributed to the fact that both models impose certain conditions on the foreground intensity distribution, and as such they cannot minimize the overlap between the object and background intensity distributions.

To demonstrate the computational efficiency of the proposed method when compared to other global methods, Table 4.2 shows the CPU time in seconds and the number of iterations to convergence for all the images considered here. Overall, the proposed method is able to segment the images in roughly half the number of iterations when compared to *SBGFRLS*, another *spf*-like model.

Table 4.2: The CPU time and the number of iterations required by the proposed *GSRPF* model, and by the *SBFRLS* and *C-V* models, to segment the foreground in some of the images considered here.

| Figure | GSRPF | | SBGFRLS | | C-V | |
|---|---|---|---|---|---|---|
| | CPU Time(s) | Iterations | CPU Time(s) | Iterations | CPU Time(s) | Iterations |
| Fig. 3(a) | 0.06 | 11 | 0.12 | 17 | - | - |
| Fig. 5(a) | 0.56 | 21 | 0.82 | 45 | 4.89 | 339 |
| Fig. 6(a) | .03 | 10 | .05 | 13 | 1.5 | 75 |
| Fig. 6(b) | 4.02 | 46 | 7.58 | 84 | 82.89 | 806 |
| Fig. 6(c) | 1.93 | 41 | 2.79 | 67 | 16.36 | 406 |
| Fig. 6(d) | 0.92 | 29 | - | - | - | - |

## 4.5 Summary

In this chapter, we have proposed a novel energy-based active contour model based on a new Globally Signed Region Pressure Force (*GSRPF*) function. The *GSRPF* model considers the global information extracted from an image and accommodates also foreground intensity distributions that are not necessarily symmetric. It automatically and efficiently modulates the signs of the pressure forces inside and outside the contour. Compared with other methods, the proposed model is less sensitive to noise, contour initialization, and can handle images with complex intensity distributions in the foreground and/or background. Our model is a Gaussian regularizing level set model that relies only on a single parameter. It is designed to have a quadratic behavior, and converges in a few iterations without penalizing the segmentation accuracy. Results on synthetic and real images from a variety of scenarios demonstrate the superiority of our model in segmentation accuracy when compared with well regarded global level set methods. As a global signed pressure force model, *GSRPF* relies on strong statistical assumptions. As a consequence, a global level set-based model, termed Concurrent Self Organizing Map-based Chan-Vese (*CSOM-CV*) model is proposed and presented in the following chapter, with the aim of attenuating such kinds of assumptions.

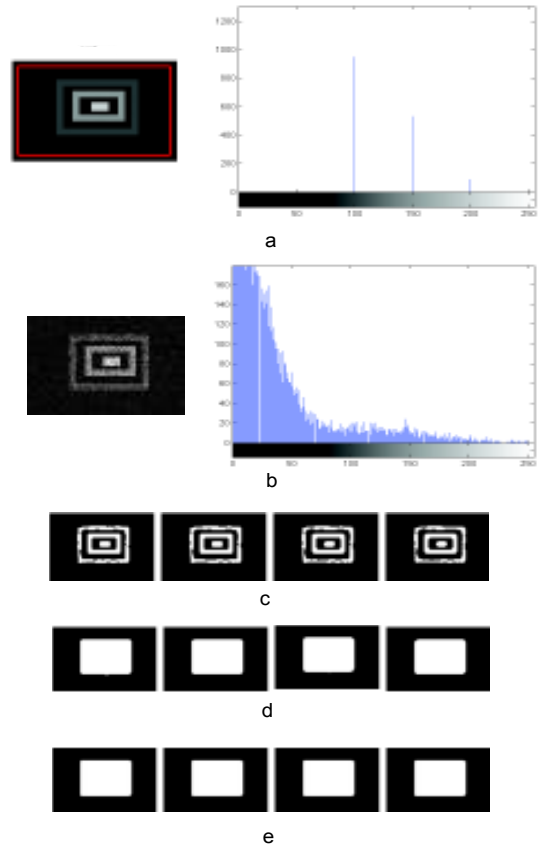

Figure 4.1: A synthetic image with multiple classes in the foreground, and the performance of the proposed, *SBGFRLS*, and *C-V*, models for some choices of their parameters. (a) the original 123 x 80 image with three different intensities 100, 150 and 200, and its histogram; (b) the same image with Gaussian noise added of standard deviation (SD) 30, and its histogram. Overlaid is also the initial contour (in red) used in all the subsequent tests. From left to right: the segmentation results of our model (c) with different $\sigma$ values (1.4, 1.6, 1.8, and 2); (d) of *SBGFRLS* with different $\sigma$ and $\alpha$ values ((2,10), (2,50), (2.5, 10), and (2.5,50), respectively); and (e) of the *C-V* model with different $\mu$ values (1.4, 1.6, 1.8, and 2).

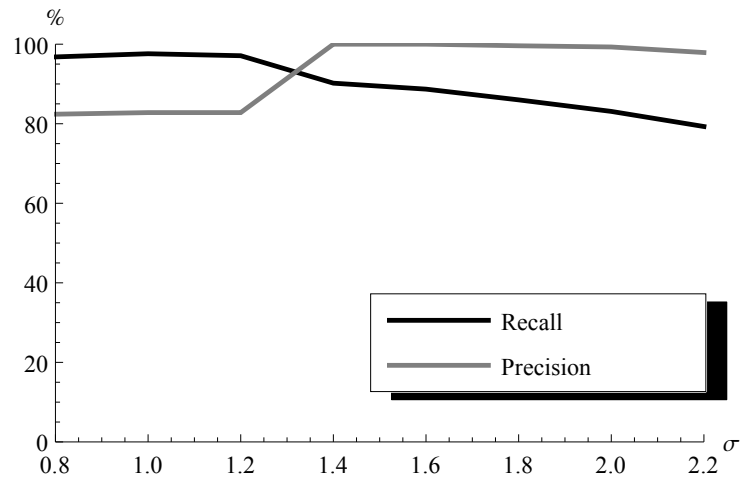

Figure 4.2: The sensitivity of our model to the parameter $\sigma$ in terms of Recall and Precision, in segmenting the image in Fig. 4.1 with Gaussian noise with standard deviation , SD = 30.

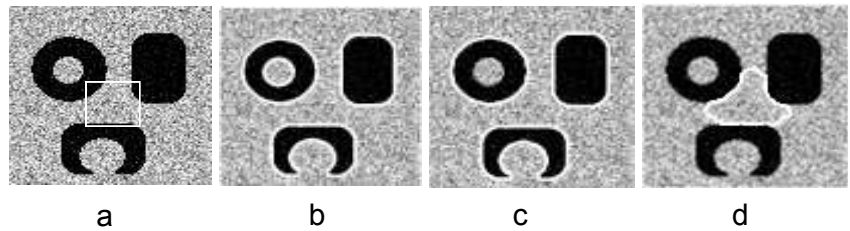

Figure 4.3: The segmentation results on a 101 x 99 synthetic image containing different objects of variable convexity and shape, and noisy background. From left to right: the original image (with the initial contour), the segmentation obtained by the proposed model ($\sigma$ = 1.4), and the ones obtained by the *SBGFRLS* and *C-V* models.

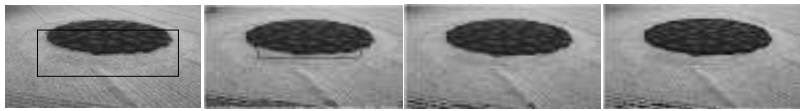

Figure 4.4: The rapid evolution of the proposed model ($\sigma = 3.5$) on a 481 x 321 real image (downloaded from [1]). From left to right: initial contour, contour after 6 and 9 iterations, and final contour (15 iterations).

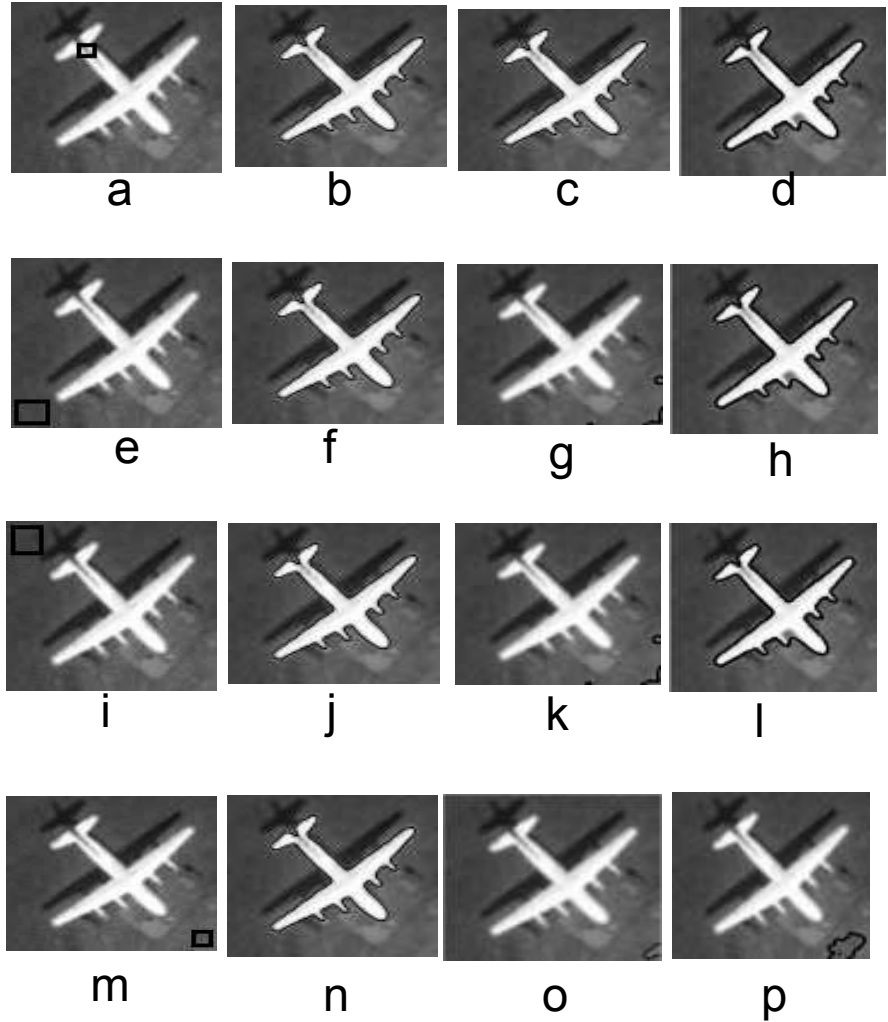

Figure 4.5: Robustness to the contour initialization when segmenting a 135 x 125 plane image obtained from [2]. Arranged in columns there are the original image with different contour initializations, and then, from left to right, the segmentation results of the proposed *GSRPF* model ($\sigma = 1.4$), and of the *SBGFRLS* (with $\sigma = 1$ and $\alpha = 25$), and *C-V* (with $\mu = 0.2$) models, respectively, when using the same initial contour.

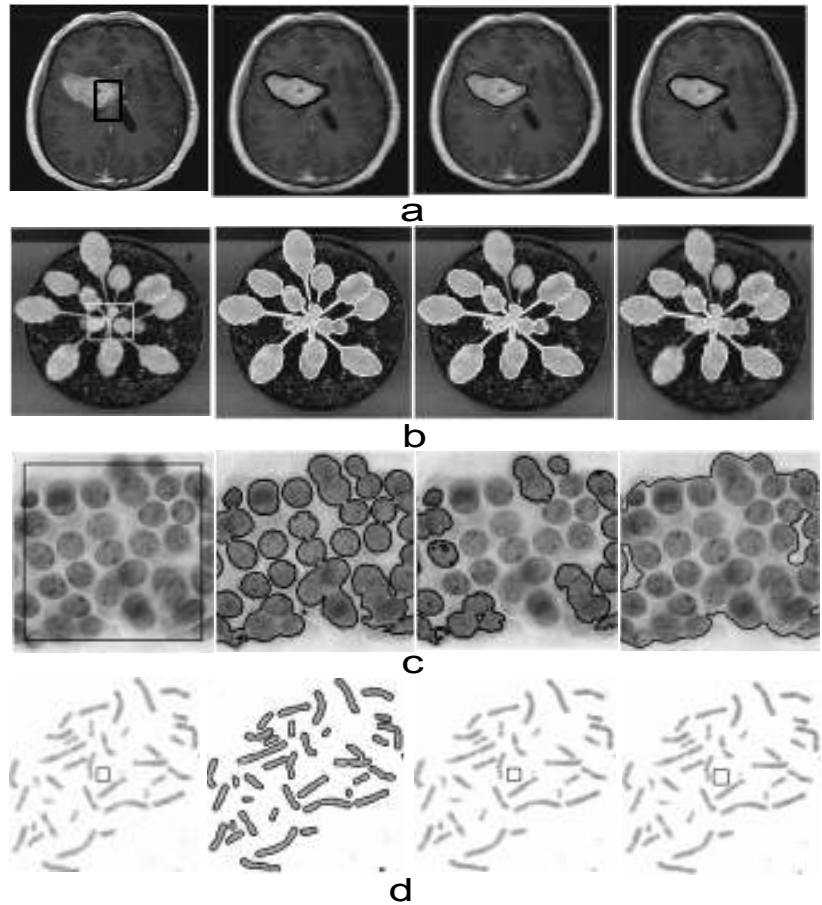

Figure 4.6: Segmentation results when different real images encountered in natural and life sciences are used. Arranged in rows there are: (a) a 109 x 119 brain MRI image, from [3]; (b) a 436 x 422 Arabidopsis optical image with complex background; (c) a 256 x 256 cellulose microscopy image, from [4]; and (d) a 256 x 256 chromosome microscopy image, from [4]. Arranged in columns there are the original image (with the initial contour), and then, from left to right the results of the proposed *GSRPF* model, and of the *SBGFRLS* and *C-V* models respectively, when using the same initial contour. (Parameters are as in Fig. 4.5, except in (a) for *GSRPF* ($\sigma = 1$)).

# Chapter 5

# Concurrent *SOM*-based Chan-Vese Model

## 5.1 Introduction

Concurrent Self Organizing Maps (*CSOM*s) deal with the pattern classification problem in a parallel processing way, aiming to minimize a suitable objective function. Similarly, Active Contour Models (*ACM*s) (e.g., the Chan-Vese (*C-V*) model) deal with the image segmentation problem as an optimization problem by minimizing a suitable energy functional. The effectiveness of *ACM*s is a real challenge in many computer vision applications. In this chapter, we propose a novel regional *ACM*, which relies on a *CSOM* to approximate the foreground and background image intensity distributions in a supervised way, and to drive the evolution of the active contour accordingly. We term our model Concurrent Self Organizing Map-based Chan-Vese (*CSOM-CV*) model [8]. The main idea of the *CSOM-CV* model is to concurrently integrate the global information extracted by a *CSOM* from a few supervised pixels into the level-set framework of the *C-V* model to build an effective *ACM*. The proposed model integrates the advantages of *CSOM* as a powerful classification tool and *C-V* as an effective tool for the optimization of a global energy functional. Experimental results show the effectiveness of *CSOM-CV* in segmenting synthetic and real images, when compared with the stand-alone *C-V* and *CSOM* models.

Most of the existing global regional *ACM*s rely explicitly on a particular probability model (e.g., Gaussian, Laplacian, etc.), which results in restricting their scope in handling images in a global fashion, and affects negatively their performance when processing noisy images. On the other hand, *SOM*-based models have the advantage of being able to predict the underlying image intensity distribution relying on their "topology preserving property", which is typical of *SOM*s. However, the application of such models in segmentation usually results in disconnected boundaries. Moreover, they are often quite sensitive to the noise. Motivated by the issues above, we propose the *CSOM-CV* model to combine *SOM*s and global *ACM*s in order to deal with the image segmentation problem reducing the disadvantages of both approaches, while preserving the aforementioned advantages.

## 5.2 The *CSOM-CV* model

In this section, we describe our Concurrent Self Organizing Map based Chan-Vese Model (*CSOM-CV*). Such model is composed of an off-line session and an on-line one, which are described, respectively, in Subsections 5.2.1 and 5.2.2.

### 5.2.1 Training session

The *CSOM-CV* model we propose makes use of two *SOM*s, one associated with the foreground, the other to the background. We make a distinction between the two *SOM*s by using, respectively, the superscripts + and − for the associated weights. We assume that two sets of training samples belonging to the true foreground $\Omega^+$ and the true background $\Omega^-$ of a training image $I^{(tr)}$ are available. They are defined as: $L^+ := \{x_1^+, \ldots, x_{|L^+|}^+ \in \Omega^+\}$ and $L^- := \{x_1^-, \ldots, x_{|L^-|}^- \in \Omega^-\}$, where $|L^+|$ and $|L^-|$ are their cardinalities.

In the following, we describe first the learning procedure of the *SOM* trained with the set of foreground training pixels $L^+$. In the training session, after choosing a suitable topology of the *SOM* associated with the foreground, the intensity $I^{(tr)}(x_t^+)$ of a randomly-extracted pixel $x_t^+ \in L^+$ of the foreground of the training image is applied as input to the neural map at time $t = 0, 1, \ldots, t_{\max}^{(tr)} - 1$, where $t_{\max}^{(tr)}$ is the number of iterations in the training of the neural map. Then, the neurons are self-organized in order to preserve - at the end of training - the topological structure of the image intensity distribution of the foreground. Each neuron $n$ of the *SOM* is connected to the input by a weight vector $w_n^+$ of the same dimension as the input (which - in the case of gray-level images considered in this work - has dimension 1). After their random initialization, the weights $w_n^+$ of the neurons are updated by the self-organization learning rule (3.1), which we re-write in the form specific for the case considered here:

$$w_n^+(t+1) := w_n^+(t) + \eta(t) h_{bn}(t)[I^{(tr)}(x_t^+) - w_n^+(t)], \qquad (5.1)$$

In this case, the *BMU* neuron $b$ is the one whose weight vector is the closest to the input $I^{(tr)}(x_t)$ at time $t$. Both the learning rate $\eta(t)$ and the neighborhood kernel $h_{bn}(t)$ are designed to be time-decreasing in order to stabilize the weights $w_n^+(t)$ for $t$ sufficiently large. In this way - due to the well-known properties [55] of the self-organization learning rule (5.1) - when the training session is completed, one can accurately model and often approximate the input intensity distribution of the foreground, by associating the intensity of each input to the weight of the corresponding *BMU* neuron. In particular, in the following we make the choice

$$\eta(t) := \eta_0 \exp\left(-\frac{t}{\tau_\eta}\right), \qquad (5.2)$$

where $\eta_0 > 0$ is the initial learning rate and $\tau_\eta > 0$ is a time constant, whereas $h_{bn}(t)$ is selected as a Gaussian function centered on the

*BMU* neuron, i.e., it has the form

$$h_{bn}(t) := \exp\left(-\frac{\|r_b - r_n\|_2^2}{2r^2(t)}\right), \tag{5.3}$$

where $r_b, r_n \in \mathbb{R}^2$ are the location vectors in the output neural map of neurons $b$ and $n$, respectively, and $r(t) > 0$ is a time-decreasing neighborhood radius (this choice of the function $h_{bn}(t)$ guarantees that, for fixed $t$, when $\|r_b - r_n\|_2$ increases, $h_{bn}(t)$ decreases to zero gradually to smooth out the effect of the *BMU* neuron on the weights of the neurons far from the *BMU* neuron itself, and when $t$ increases, the influence of the *BMU* neuron becomes more and more localized). In particular, in the following we choose

$$r(t) := r_0 \exp\left(-\frac{t}{\tau_r}\right), \tag{5.4}$$

where $r_0 > 0$ is the initial neighborhood radius of the map, and $\tau_r > 0$ is another time constant.

The learning procedure of the other *SOM* differs only in the random choice of the training pixel (which is now denoted by $x_t^-$, and belongs to the set $L^-$), and in the weights of the network, which are denoted by $w_n^-$.

### 5.2.2 Testing session

Once the training of the two *SOM*s has been accomplished, the two trained networks are applied on-line in the testing session, during the evolution of the contour $C$, to approximate and describe globally the foreground and background intensity distributions of a similar test image $I(x)$. Indeed, during the contour evolution, the two mean intensities $\text{mean}(I(x)|x \in \text{in}(C))$ and $\text{mean}(I(x)|x \in \text{out}(C))$ in the current approximations of the foreground and background are presented as inputs to the two trained networks. We now define the quantities

$$w_b^+(C) := \text{argmin}_n \, |w_n - \text{mean}(I(x)|x \in \text{in}(C))| \,, \tag{5.5}$$
$$w_b^-(C) := \text{argmin}_n \, |w_n - \text{mean}(I(x)|x \in \text{out}(C))| \,, \tag{5.6}$$

where $w_b^+(C)$ is the prototype of the *BMU* neuron to the mean intensity inside the current contour, while $w_b^-(C)$ is the prototype of the *BMU* neuron to the mean intensity outside it. Then, we define the functional of the *CSOM − CV* model as

$$E_{CSOM-CV}(C) := \lambda^+ \int_{\text{in}(C)} e^+(x,C)dx$$
$$+\lambda^- \int_{\text{out}(C)} e^-(x,C)dx, \quad (5.7)$$

$$e^+(x,C) := \left(I(x) - w_b^+(C)\right)^2, \quad (5.8)$$

$$e^-(x,C) := \left(I(x) - w_b^-(C)\right)^2. \quad (5.9)$$

where the parameters $\lambda^+, \lambda^- \geq 0$ are, respectively, the weights of the two image energy terms $\int_{\text{in}(C)} e^+(x,C)dx$ and $\int_{\text{out}(C)} e^-(x,C)dx$, inside and outside the contour.

Now, as in [28], we replace the contour curve $C$ with the level set function $\phi$, obtaining

$$E_{CSOM-CV}(\phi) = \lambda^+ \int_{\phi>0} e^+(x,\phi)dx + \lambda^- \int_{\phi<0} e^-(x,\phi)dx, \quad (5.10)$$

where we have also made explicit the dependence of $e^+$ and $e^-$ on $\phi$. In terms of the Heaviside step function $H(\cdot)$, the *CSOM-CV* energy functional can be also written as follows:

$$E_{CSOM-CV}(\phi) = \lambda^+ \int_\Omega e^+(x,\phi)H(\phi(x))dx$$
$$+\lambda^- \int_\Omega e^-(x,\phi)(1-H(\phi(x)))dx. \quad (5.11)$$

Finally, proceeding likewise in [28], by an application of the gradient-descent technique in an infinite-dimensional setting, the evolution of the contour is described by the *PDE*

$$\frac{\partial \phi}{\partial t} = \delta(\phi)\left[-\lambda^+ e^+ + \lambda^- e^-\right],  \tag{5.12}$$

which shows how the learned neurons of the two *SOM*s are used to determine the internal and external forces acting on the contour. Moreover, in a similar way to [122], we perform - at each iteration of a finite-difference approximation of (5.12) - the regularization of the current level set function by replacing it with its convolution with a Gaussian filter of suitable width. Finally, the contour evolution is performed for $t_{\max}^{(evol)}$ iterations (unless convergence is obtained before). Another difference with the *C-V* model is the absence of the regularization terms in $\mu$ and $\nu$. This can be justified as follows. As pointed out in [122, 123], the convolution of the current level set function with a Gaussian filter can be used as an efficient and robust approach to regularize it. In such an approach, the width of the Gaussian filter is used to control the regularization strength, as the parameters $\mu$ and $\nu$ do in the *C-V* model. So, in a similar way to our previous models, we have not included in our formulation the regularization parameters $\mu$ and $\nu$.

## 5.3 Implementation

The procedural steps of the training and testing sessions for the *CSOM-CV* model are summarized in Algorithm 5.3.

## Algorithm 1 *CSOM − CV* segmentation framework

1: **procedure**
   - Input:
     - Training and test scalar-valued images, and supervised pixels of the training image belonging, respectively, to the sets $L^+$ and $L^-$.
     - Topology of the two neural maps (with 2-dimensional prototypes), and their respective numbers $FN$ and $BN$ of neurons in the output layer.
     - Number of iterations $t_{\max}^{(tr)}$ for training the two neural maps.
     - Maximum number of iterations $t_{\max}^{(evol)}$ for the contour evolution.
     - $\eta_0 > 0$: starting learning rate.
     - $r_0 > 0$: starting radius of the maps.
     - $\tau_\eta, \tau_r > 0$: time constants in the learning rate and contour smoothing parameter.
     - $\lambda^+, \lambda^- \geq 0$: weights of the energy terms, respectively, inside and outside the contour.
     - $\sigma$: Gaussian smoothing parameter.
     - $\rho > 0$: constant in the binary approximation of the level set function.
   - Output:
     - Segmentation result.

   *TRAINING SESSION:*
2:   Initialize randomly the prototypes of the neurons of the two maps.
3:   **repeat**
4:     Choose randomly a pixel $x_t^+ \in L^+$ and determine the *BMU* neuron of the *SOM* associated with the foreground to the input intensity $I^{(tr)}(x_t^+)$.
5:     Update the prototypes $w_n^+$ of the *SOM* associated with the foreground using (5.1), (5.2), (5.3), and (5.4).
6:   **until** learning of the prototypes is accomplished (i.e., the number of iterations $t_{\max}^{(tr)}$ is reached).
7:   Proceed similarly for the training of the *SOM* associated with the background, with $x_t^+ \in L^+$ and $w_n^+$ replaced, respectively, by $x_t^- \in L^-$ and $w_n^-$.

   *TESTING SESSION:*
8:   Choose a subset $\Omega_0$ (e.g., a rectangle) in the image domain $\Omega$ with boundary $\Omega_0'$, and initialize the level set function as:

$$\phi(x) := \begin{cases} \rho, & x \in \Omega_0 \setminus \Omega_0', \\ 0, & x \in \Omega_0', \\ -\rho, & x \in \Omega \setminus (\Omega_0 \cup \Omega_0'). \end{cases} \quad (5.13)$$

9:   **repeat**
10:    Calculate the functions $w_b^+$ and $w_b^-$ from (5.5) and (5.6).
11:    Evolve the level set function $\phi$ according to a finite difference approximation of (5.12).
12:    At each iteration of the finite-difference scheme, re-initialize the current level set function to be binary by performing the update

$$\phi \leftarrow \rho \left( H(\phi) - H(-\phi) \right), \quad (5.14)$$

   then regularize by convolution the obtained level set function:

$$\phi \leftarrow g_{\sigma'} \otimes \phi, \quad (5.15)$$

   where $g_{\sigma'}$ is a Gaussian kernel with $\int_{\mathbb{R}^2} g_{\sigma'}(x)dx = 1$ and width $\sigma'$.
13:  **until** the curve evolution converges (i.e., the curve does not change anymore) or the maximum number of iterations $t_{\max}^{(evol)}$ is reached.
14: **end procedure**

## 5.4 Experimental study

In this section, we demonstrate the effectiveness of the *CSOM-CV* model, when compared to the stand-alone *CSOM* and *C-V* models, in handling synthetic and real images. For a fair comparison, the *CSOM-CV*, *C-V* and the *CSOM* models used in this experiment are all implemented in Matlab R2012a on a PC with the following configuration: 1.8 GHz Intel(R) Core(TM) i3-3217U, and 4.00 GB RAM. In each experiment, the *CSOM-CV* parameters are fixed as follows: $\eta_0 = 0.9$, $\sigma = 1.5$, and the weight parameters (i.e., $\lambda^+$, $\lambda^-$) are fixed to 1. Also, $r_0 = \max(M, N)/2$, where $M$ and $N$ are the numbers of rows and columns of the neural map, $t_{\max}^{(tr)} = 10000$, $t_{\max}^{(evol)} = 1000$, $\tau_\eta = t^{(tr)\max}$, $\tau_r = t_{\max}^{(tr)}/\ln(r_0)$, $\rho = 1$. The *SOM*s are composed of $3 \times 3$ neurons in most experiments (i.e., $M = N = 3$). In the *C-V* model, $\lambda^+$, $\lambda^-$ are also fixed to 1, $\mu$ is chosen such that the final contour is smooth enough, and $\nu = 0$ (as made in [27, p. 268]). All the gray-level images considered in this section are 8-bit images, so the range of the values assumed by the intensity is 0-255.

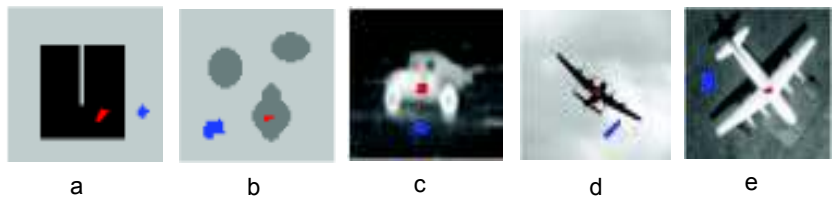

      a          b          c          d          e

Figure 5.1: The training images used in this chapter together with the supervised foreground pixels (red) and the supervised background pixels (blue) used in training sessions of the *CSOM-CV* and the *CSOM* models. By its definition, no supervised pixel is used by the *C-V* model.

To demonstrate the robustness of *CSOM-CV* to the noise, in the experiment described in Fig. 5.2 we have used the noise-free images of Fig. 5.1(a) and (b) in the training sessions of *CSOM-CV* and *CSOM*, then the trained *SOM*s have been applied on-line by the two models to their noisy versions as test images. As shown in Fig. 5.2, for this case *CSOM-CV* is more robust and less sensitive to the noise than *C-V* (which does not make use of supervised training examples) and *CSOM*, since the regions of the foreground are detected more accurately by *CSOM-CV*.

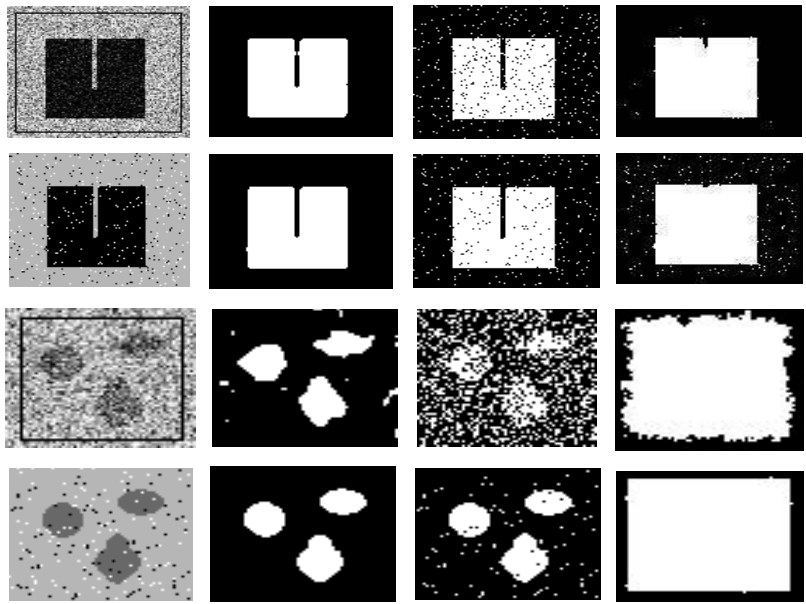

Figure 5.2: The robustness of the *CSOM-CV* model to two different kinds of noise: the first column shows, from top to down, two noisy versions of the image shown in Fig. 5.1(a), and two noisy versions of the image shown in Fig. 5.1(b), respectively, with the addition of Gaussian noise with standard deviation $SD = 50$ (first and third row) and salt and pepper noise (second and fourth row). The initial contours used by the *CSOM-CV* and *C-V* models are also shown (first and third row); finally, the second, third, and fourth columns show, respectively, the corresponding binary segmentation obtained by the *CSOM-CV*, *CSOM*, and *C-V* models.

Fig. 5.3 illustrates the effectiveness of *CSOM-CV* in handling other images. The segmentation results of the *CSOM-CV* model shown in the first row demonstrate its ability to segment objects with blurred edges and background, while on the same im-ages the *CSOM* and *C-V* models incur, respectively, in over- and under- segmentation problems. Similarly, as shown, respectively, in the second and third rows, *CSOM-CV* outperforms *CSOM* and *C-V* also in handling images characterized by nonhomogeneous background intensity distribution, and in the presence of a shadow.

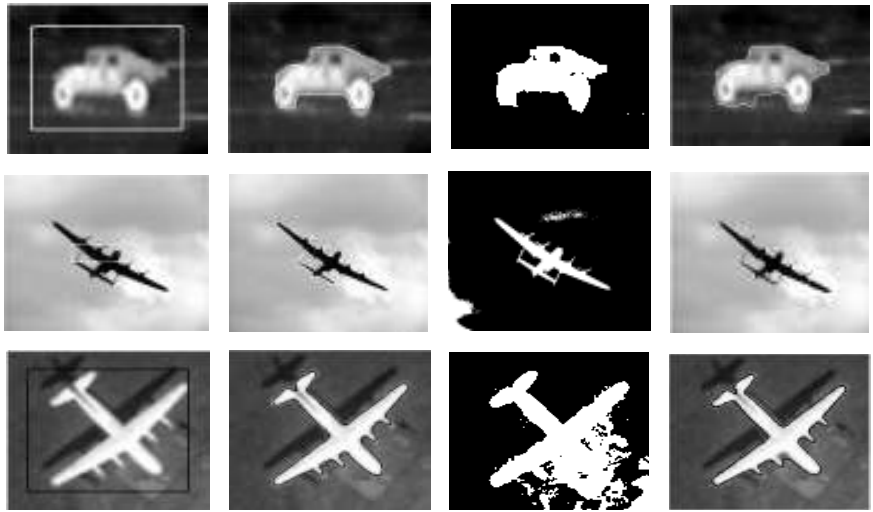

Figure 5.3: The segmentation results obtained on real and synthetic gray-level images. The first row shows the original images with the initial contours, while the second, third, and fourth rows show, respectively, the corresponding segmentation results obtained by the *CSOM-CV*, *CSOM*, and *C-V* models.

To demonstrate the computational efficiency of the *CSOM-CV* model when compared to the *CSOM* and *C-V* models, Table 5.1 shows, for each of the three methods, the *CPU* time (in seconds) required to segment the images shown in Fig. 5.2 and 5.3. For the *CSOM-CV* and *C-V* models, the number of iterations performed before convergence of the active contour is also reported in the table. As illustrated by Table 5.1, we can observe that the *CSOM-CV* model has demonstrated to be much faster than the *CSOM* and *C-V* models in all the listed cases, thus confirming the efficiency of the *CSOM-CV* model. Moreover, as illustrated in Table 5.2, we have also used the Precision, Recall, and *F*-measure metrics (where the "positive" pixels are the foreground pixels) to evaluate quantitatively the segmentation results of all the models, confiming the effectiveness of the *CSOM-CV* model when compared to the *CSOM* and *C-V* models.

Table 5.1: The contour evolution time and number of iterations required by the *CSOM-CV* and *C-V* models to segment the foreground for some of the images shown in this chapter. The *CPU* time of *CSOM* is also included.

| Image in | Image size | *CSOM-CV* model | | *CSOM* model | *C-V* model | |
|---|---|---|---|---|---|---|
| | | *CPU* Time (s) | # Iter. | *CPU* Time (s) | *CPU* Time (s) | # Iter. |
| Fig. 5.2 row 1 | 114 × 101 | 0.73 | 20 | 16.7 | 3.2 | 158 |
| Fig. 5.2 row 2 | 114 × 101 | 0.62 | 18 | 14.7 | 3.64 | 219 |
| Fig. 5.2 row 3 | 64 × 61 | 0.078 | 10 | 4.9 | 0.04 | 4 |
| Fig. 5.2 row 4 | 64 × 61 | 0.07 | 10 | 5 | 0.98 | 30 |
| Fig. 5.3 row 1 | 118 × 93 | 0.04 | 10 | 6.12 | 2.12 | 137 |
| Fig. 5.3 row 2 | 300 × 225 | 0.62 | 37 | 42.03 | 6.68 | 205 |
| Fig. 5.3 row 3 | 135 × 125 | 0.15 | 17 | 10.1 | 4.18 | 266 |

Table 5.2: The Precision, Recall, and *F*-measure metrics for the *CSOM-CV*, *CSOM*, and *C-V* models.

| Image in | *CSOM-CV* model | | | *CSOM* model | | | *C-V* model | | |
|---|---|---|---|---|---|---|---|---|---|
| | P(%) | R(%) | F-m.(%) | P(%) | R(%) | F-m.(%) | P(%) | R(%) | F-m.(%) |
| Fig. 5.2 row 1 | 99.7 | 99.8 | 99.8 | 93.5 | 94.7 | 94 | 97 | 88.3 | 92.5 |
| Fig. 5.2 row 2 | 99.8 | 99.9 | 99.8 | 94.7 | 97.5 | 96.1 | 94.2 | 87.2 | 90.5 |
| Fig. 5.2 row 3 | 48.2 | 93.9 | 63.7 | 16.4 | 72.6 | 26.8 | 12.7 | 96.4 | 22.5 |
| Fig. 5.2 row 4 | 56.8 | 97.4 | 71.1 | 48.7 | 96.4 | 64.7 | 12.2 | 100 | 21.7 |
| Fig. 5.3 row 1 | 100 | 94.3 | 97.1 | 99.6 | 92.1 | 95.7 | 92.8 | 82.9 | 87.6 |
| Fig. 5.3 row 2 | 63.5 | 89.5 | 74.3 | 39.2 | 95.4 | 55.6 | 73 | 60 | 65.9 |
| Fig. 5.3 row 3 | 95.7 | 99.8 | 97.7 | 46.5 | 100 | 63.5 | 94.9 | 61.4 | 74.6 |

## 5.5 Summary

In this chapter, we have proposed a novel *SOM*-based *ACM* model, the Concurrent Self Organizing Map-based Chan-Vese (*CSOM-CV*) model, which relies mainly on a set of prototypes coming from two trained *SOM*s to guide the evolution of the active contour. The *CSOM-CV* model is a supervised and global region-based *ACM*. It has been demonstrated to be efficient and robust to two different kinds of noise. As compared to the *C-V* model, our proposed solution consists instead in modeling globally in a supervised way the intensity distributions of the foreground/background (relying on a few supervised pixels) without using parametric models, but relying on a set of prototypes resulting from the training of a *CSOM*. So, the main reasons for which, as shown experimentally in Section 5.4, the proposed model affects positively the *C-V* model in terms of speed-up in the testing phase and robustness to the noise are that - differently from the proposed model - the *C-V* model refers to Gaussian intensity distributions of the foreground/background, and does not include supervised examples. Moreover, as compared to *CSOM* and in general to *SOM*-like models used in image segmentation, our solution consists in modeling the active contour using a variational level set method and relying at the same time on a few prototypes coming from the learned *CSOM*. In this way, the *CSOM-CV* model is able to produce a final segmentation result characterized by a smooth contour while most *SOM*-like models usually produce segmentations characterized by disconnected boundaries. The enhanced version of *CSOM-CV*, termed *Self Organizing Active Contour* (*SOAC*) model, has been proposed with the aim of enlarging its scope and applicability.

# Chapter 6

# Self Organizing *AC* Model

## 6.1 Introduction

Active contour models can significantly improve their performance by constructing a knowledge base, and use it to supervise the movement of the contour during its evolution. However, the state-of-the-art supervised *ACM*s make usually statistical assumptions on the image intensity distribution of each subset to be modeled. This results in that the evolution of the contour is driven by the probability model constructed based on the given reference distributions. As a consequence, the scope of those models is limited by how accurate the probability model is. The situation is even worsened by the limited availability of training samples. In this chapter, we present a supervised *ACM*, termed *Self Organizing Active Contour* (*SOAC*) model [9], which is an extension and improvement of our previous *CSOM-CV* model presented in Chapter 5.

Compared to the *CSOM-CV* model, the *SOAC* model makes the following important improvement: its regional descriptors $w_b^+(x, C)$ and $w_b^-(x, C)$ depend on the pixel location $x$, while the the *CSOM-CV* model uses regional descriptors of the form $w_b^+(C)$ and $w_b^-(C)$, which are constant functions. So, the *CSOM-CV* model is a global *ACM* (i.e., the spatial dependences of the pixels are not taken into account in such a model, since it just considers only the average intensities inside and outside the contour), whereas the *SOAC* model makes also use of local information, which provides it the ability of handling more complex images. For this reason, the *CSOM-CV* model is not able to deal successfully with some of the images presented in this chapter, although it still showed better performance than *CSOM* and *C-V* for some images (as detailed in the previous chapter). However, differently from the *SOAC* model, it is not able to deal properly with images presenting challenges such as intensity inhomogeneity and foreground/background intensity overlap.

## 6.2 The *SOAC* Model

Our proposed solution to address efficiently the limitations of current *ACM*s, mentioned in Chapter 2, and unsupervised *SOM*-based *ACM*s, discussed in Chapter 6, is to deal implicitly - through *SOMS*s - with the decision boundary between the subsets, instead of relying on a particular probability model for a pixel to be an element of each subset. In this way, images that contain intensity inhomogeneity, an overlap between the foreground/background intensity distributions, and/or objects characterized by many different intensities, can be handled in an efficient way, as demonstrated by the obtained experimental comparisons of our model with other *ACM*s, which are reported in Section 6.4 after the presentation of our model.

In this section, we describe in details our proposed *Self Organizing Active Contour* (*SOAC*) model. We first consider the case of scalar-valued images in Subsection 6.2.1. Then, in Subsection 6.2.2, we detail the changes needed to deal with the case of vector-valued images. Finally, in Subsection 6.3, algorithmic details are provided for the two cases.

### 6.2.1 The *SOAC* model for scalar-valued images

The *SOAC* segmentation framework is composed of two sessions: a supervised training session, and a testing session. The training process is perfomed in a similar way as the training session of the *CSOM-CV* model, as described in Subsection 5.2.1 of chapter 5).

Once the training sessions of both *SOM*s have been completed, the trained networks are applied online, during the evolution of the contour C, to a test image of intensity $I(x)$, with the aim of approximating and describing locally, respectively, the foreground and background intensity distributions. Indeed, during the contour evolution [1], the two local weighted average intensities

$$\frac{\int_{in(C)} g_\sigma(x-y)I(y)\,dy}{\int_{in(C)} g_\sigma(x-y)dy} \qquad (6.1)$$

and

$$\frac{\int_{out(C)} g_\sigma(x-y)I(y)\,dy}{\int_{out(C)} g_\sigma(x-y)dy} \qquad (6.2)$$

are presented as inputs to the two trained networks, respectively, for each pixel $x \in in(C)$ and $x \in out(C)$. Here, $g_\sigma$ is a Gaussian kernel function with $\int_{\mathbb{R}^2} g_\sigma(x)dx = 1$ and width $\sigma$, which determines the effective size of the neighborhood of $x$ on which the integrals in (6.1) and (6.2) are performed. The prototypes of the *BMU* neurons associated to the inputs (6.1) and (6.2) are, respectively,

$$w_b^+(x,C) := \operatorname{argmin}_{n \in \{1,\ldots,FN\}} \left| w_n^+ - \frac{\int_{in(C)} g_\sigma(x-y)I(y)\,dy}{\int_{in(C)} g_\sigma(x-y)dy} \right|, \qquad (6.3)$$

$$w_b^-(x,C) := \operatorname{argmin}_{n \in \{1,\ldots,BN\}} \left| w_n^- - \frac{\int_{out(C)} g_\sigma(x-y)I(y)\,dy}{\int_{out(C)} g_\sigma(x-y)dy} \right|. \qquad (6.4)$$

Such prototypes are extracted as local regional intensity descrip-tors of the foreground and background, respectively, and included in the energy functional to be minimized in our proposed *SOAC* model, which has the following expression:

$$E_{SOAC}(C) := \lambda^+ \int_{in(C)} e^+(x,C)dx$$
$$+\lambda^- \int_{out(C)} e^-(x,C)dx, \qquad (6.5)$$

$$e^+(x,C) := \left(I(x) - w_b^+(x,C)\right)^2, \qquad (6.6)$$

$$e^-(x,C) := \left(I(x) - w_b^-(x,C)\right)^2, \qquad (6.7)$$

where the parameters $\lambda^+, \lambda^- \geq 0$ are the weights of the two image energy terms $\int_{in(C)} e^+(x,C)dx$ and $\int_{out(C)} e^-(x,C)dx$, respectively, inside and outside the current contour.

---

[1]In practice, this process of the testing session can be done off-line because there is no need here for the user to initialize a contour to determine the domains of the interior and exterior forces as they are automatically determined by the previously trained *SOM*s.

The terms $e^+(x,C)$ and $e^-(x,C)$ in (6.6) and (6.7) are able to model more complex intensity distributions than the terms $(I(x) - c^+(C))^2$ and $(I(x) - c^-(C))^2$ used in the energy formulation of the *C-V* model: in particular, they are able to model skewed and multimodal intensity distributions.

Now, we replace the contour curve $C$ with the level set function $\phi$, obtaining

$$E_{SOAC}(\phi) = \lambda^+ \int_{\phi>0} e^+(x,C)dx + \lambda^- \int_{\phi<0} e^-(x,C)dx. \qquad (6.8)$$

where we have also made explicit the dependence of $e^+$ and $e^-$ on $\phi$. In terms of the Heaviside step function $H(\cdot)$, the energy functional of the *SOAC* model can be also written as follows:

$$E_{SOAC}(\phi) = \lambda^+ \int_\Omega e^+(x,C)H(\phi(x))dx$$
$$+\lambda^- \int_\Omega e^-(x,C)(1 - H(\phi(x)))dx. \qquad (6.9)$$

Finally, the evolution of the contour can be described by the *PDE*

$$\frac{\partial \phi}{\partial t} = \delta(\phi)[-\lambda^+ e^+ + \lambda^- e^-],  \qquad (6.10)$$

which shows how the learned neurons for each subset are used to determine the internal and external forces acting on the contour.

### 6.2.2 The *SOAC* model for vector-valued images

The *SOAC* model can be extended to the case of vector-valued images. Such an extension is particularly useful for the segmentation of multi-spectral images (see Section 6.4 for some related experiments). In the vectorial case, the image $\mathbf{I}(x)$ is made up of $D$ channels $I_i(x)$ ($i = 1, \ldots, D$), and also the *SOM* weights are vectors of dimension $D$. The only significant change with respect to the scalar case described in Subsection 6.2.1 is that, in the determination of the *BMU* neurons, the absolute values in formulas (6.3) and (6.4) are replaced by Euclidean norms in $\mathbb{R}^D$.

## 6.3 Implementation

The procedural steps of the training and testing sessions for the *SOAC* model are summarized in Algorithm 2 for the scalar case (only a slight modification is needed in the vectorial case, as discussed in Subsection 6.2.2).

## Algorithm 2 *SOAC* segmentation framework for scalar-valued images

1: **procedure**
   - Input:
     - Training and test scalar-valued images, and supervised pixels of the training image belonging, respectively, to the sets $L^+$ and $L^-$.
     - Topology of the two neural maps (with 1-dimensional prototypes), and their respective numbers *FN* and *BN* of neurons in the output layer.
     - Number of iterations $t_{\max}^{(tr)}$ for training the two neural maps.
     - Maximum number of iterations $t_{\max}^{(evol)}$ for the contour evolution.
     - $\eta_0 > 0$: starting learning rate.
     - $r_0 > 0$: starting radius of the maps.
     - $\tau_\eta, \tau_r > 0$: time constants in the learning rate and contour smoothing parameter.
     - $\lambda^+, \lambda^- \geq 0$: weights of the energy terms, respectively, inside and outside the contour.
     - $\sigma, \sigma' > 0$: Gaussian intensity and contour smoothing parameters.
     - $\rho > 0$: constant in the binary approximation of the level set function.
   - Output:
     - Segmentation result.

   *TRAINING SESSION:*
2:  Initialize randomly the prototypes of the neurons of the two maps.
3:  **repeat**
4:    Choose randomly a pixel $x_t^+ \in L^+$ and determine the *BMU* neuron of the *SOM* associated with the foreground to the input intensity $I^{(tr)}(x_t^+)$.
5:    Update the prototypes $w_n^+$ of the *SOM* associated with the foreground using (5.1), (5.2), (5.3), and (5.4).
6:  **until** learning of the prototypes is accomplished (i.e., the number of iterations $t_{\max}^{(tr)}$ is reached).
7:  Proceed similarly for the training of the *SOM* associated with the background, with $x_t^+ \in L^+$ and $w_n^+$ replaced, respectively, by $x_t^- \in L^-$ and $w_n^-$.

   *TESTING SESSION:*
8:  Choose a subset $\Omega_0$ (e.g., a rectangle) in the image domain $\Omega$ with boundary $\Omega_0'$, and initialize the level set function as:

$$\phi(x) := \begin{cases} \rho, & x \in \Omega_0 \setminus \Omega_0', \\ 0, & x \in \Omega_0', \\ -\rho, & x \in \Omega \setminus (\Omega_0 \cup \Omega_0'). \end{cases} \quad (6.11)$$

9:  **repeat**
10:   Calculate the functions $w_b^+$ and $w_b^-$ from (6.3) and (6.4).
11:   Evolve the level set function $\phi$ according to a finite difference approximation of (6.10).
12:   At each iteration of the finite-difference scheme, re-initialize the current level set function to be binary by performing the update

$$\phi \leftarrow \rho\left(H(\phi) - H(-\phi)\right), \quad (6.12)$$

   then regularize by convolution the obtained level set function:

$$\phi \leftarrow g_{\sigma'} \otimes \phi, \quad (6.13)$$

   where $g_{\sigma'}$ is a Gaussian kernel with $\int_{\mathbb{R}^2} g_{\sigma'}(x)dx = 1$ and width $\sigma'$.
13: **until** the curve evolution converges (i.e., the curve does not change anymore) or the maximum number of iterations $t_{\max}^{(evol)}$ is reached.
14: **end procedure**

## 6.4 Experimental study

In this section, we demonstrate the effectiveness and robustness of our proposed method, compared to implementations of some of the state-of-the-art *ACM*s described in chapter 2, in handling synthetic and real images which present well-known challenges in computer vision. More precisely, we study the behavior of the *SOAC* model - when compared to other state-of-the-art image segmentation models - in handling scalar-valued and vector-valued images containing objects with non-homogeneous intensity (e.g., in the presence of various different gray values and simple/complex object shapes, intensity inhomogeneity in both the foreground and background, different kinds of noise, ill-defined edges, foreground/background intensity overlap, etc.). For a fair comparison, all models have been implemented in Matlab R2012a on a PC with the following configuration: 2.5 GHz Intel(R) Core(TM) 2 Duo, and 2.00 GB RAM. Moreover, all the *ACM* reference models used in this experiment are Gaussian regularizing level set models.

In each experiment, the $r_0$, $\sigma$ and $\sigma'$ parameters are expressed in pixels. In the experiments performed, the parameters were tuned by trial and error on a validation image (the one in Fig. 6.5(row 1)), and their values reported in the chapter were the ones that yielded the best results on it. Then, the parameters were fixed to such values for all the other images considered in the chapter, with the exception of the images in rows 3 and 4 of Fig. 6.5 and the images in Fig. 6.8, which required an additional manual tuning of the locality parameter $\sigma$, because of the presence of intensity inhomogeneity in such images (see also [61] and [64] as other examples of local *ACM*s for a which a similar locality parameter was tuned manually for each experiment). When parameters slightly different from the ones reported in the chapter were used, only a small difference in performance was observed. In more details, in the experiments, the *SOAC* parameters were fixed as follows: $\eta_0 = .1$, $\sigma = .1$ (apart for the images in rows 3 and 4 of Fig. 6.5 and the images in Fig. 6.8, for which we chose $\sigma = 21$), $\sigma' = 1.5$, and the weight parameters (i.e., $\lambda^+$, $\lambda^-$ for the scalar-valued case, and $\lambda_i^+$, $\lambda_i^-$ in the vector-valued case) were fixed to 1. Also, $r_0 = .5$,

$t^{(tr)}_{\max} = 10000$, $t^{(evol)}_{\max} = 1000$, $\tau_\eta := t^{(tr)}_{\max}$, $\tau_r := t^{(tr)}_{\max}/\ln(r_0)$, $\rho = 1$. The two *SOM*s had the same 1-D structure, and were composed of 3 output neurons. In the *C-V* model, $\lambda^+$, $\lambda^-$ for the scalar-valued case and $\lambda_i^+$, $\lambda_i^-$ in the vector-valued case were also fixed to 1, $\mu$ was chosen such that the final contour was smooth enough and $\nu = 0$ (as done in [27, p. 268]). Moreover, the number of computational units *K* of the *GMM*- based model (see Chapter 2) were chosen to be 2 (also larger numbers of computational units were considered, but the best results were obtained for $K = 2$). The $\sigma'$ parameter was fixed for both the *GMM*-based and *KDE*-based models to be equal to the one of the *SOAC* model. The other parameters of the models used in the comparison with the *SOAC* model were chosen following the recommendations of the papers in which the models were, respectively, proposed: for instance, the parameter $\sigma_{KDE}$ in the *KDE*-based model was fixed as the average nearest neighbor distance, as recommended in [35]. For the case of gray-level images, the range of the values assumed by the intensity is 0-255, as all the considered gray-level images are 8-bit images. Unless stated otherwise, the training image used in the training session coincided with the test image. Otherwise, it was an image similar to the test image (obtained, e.g., by adding Gaussian noise). In all the testing sessions, the initial contour was chosen as rectangular.

The images considered in the chapter were chosen for their complexity, and because several known *ACM*s often showed undesirable results when tested on them. In Table 6.1, we report the reference from which each image was taken (apart from the first artificial image), its size, and the associated number of pixels in the foreground and background that have been used in the training phase. In more details, the image in Fig. 6.5(row 1) was taken from [109] as an example of a real infrared image with very weak boundaries, whereas the one in Fig. 6.5(row 2) comes from [120], and is an example of an image with some shadows. The other images in Fig. 6.5(rows 3, 4, and 5) were downloaded from the Computer Vision Database [5], and represent real images corrupted by intensity inhomogeneity, whereas the one in Fig. 6.8(row 1) was taken from [120] as an example of a real *MRI* image with intensity inhomogeneity. The image in Fig. 6.8(row 2) comes from [108], and is a real *CT* brain image with intensity inhomogeneity. The one in Fig. 6.10(row 1) was taken from [123] as a real image with a very noisy background. The other images in Fig. 6.10(rows 2 and 3) were downloaded from the Berkeley image segmentation data set [6], and represent real images containing a significant overlap in the foreground/background intensity distributions.

In order to demonstrate the effectiveness of our model in handling images containing an overlap in the foreground/background intensity distributions and/or containing objects in the foreground characterized by several intensities, we created first a synthetic image with such characteristics, which is shown in Fig. 6.1(a), whereas Fig. 6.1(b) shows its histogram. In the remaining of Fig. 6.1, we illustrate the segmentation performance obtained by the *SOAC* model when compared to unsupervised *ACM*s such as *LRCV* and *C-V* in handling such an image. Also, the performance of *SOAC* in handling the noisy version of the same image was tested and compared with some supervised *ACM*s trained on the same data. As a motivation for the use of supervised data, Fig. 6.1(f) illustrates the good capability of *SOAC* in handling the image shown in Fig. 6.1(a), whereas both the *LRCV* (Fig. 6.1(h)) and the *C-V* (Fig. 6.1(i)) models, which are unsupervised, failed to segment the same image. Moreover, due to the foreground/background intensity overlap, both the *KDE*-based (Fig. 6.1(m)) and the *GMM*-based (Fig. 6.1(n)) models failed to separate the three objects of the foreground when segmenting the same synthetic image with the addition of Gaussian noise (Fig. 6.1(j)). On the other hand, as shown in Fig. 6.1(l), in such a case *SOAC* outperformed both the supervised *KDE*-based and *GMM*-based models. In order to have a fair comparison with another *SOM*-based model, Fig. 6.1(k) shows for the same image the output mask of the *CSOM* classifier, which illustrates its high sensitivity to noise. This confirms that *SOAC* was not biased by the performance of the *CSOM* classifier. Additionally, *SOAC* was tested on the same synthetic image shown in Fig. 6.1(a) with the addition of salt and pepper noise (see Fig. 6.2), and was able to correctly find the desired object also in such a situation.

To demonstrate the accuracy of *SOAC* quantitatively and

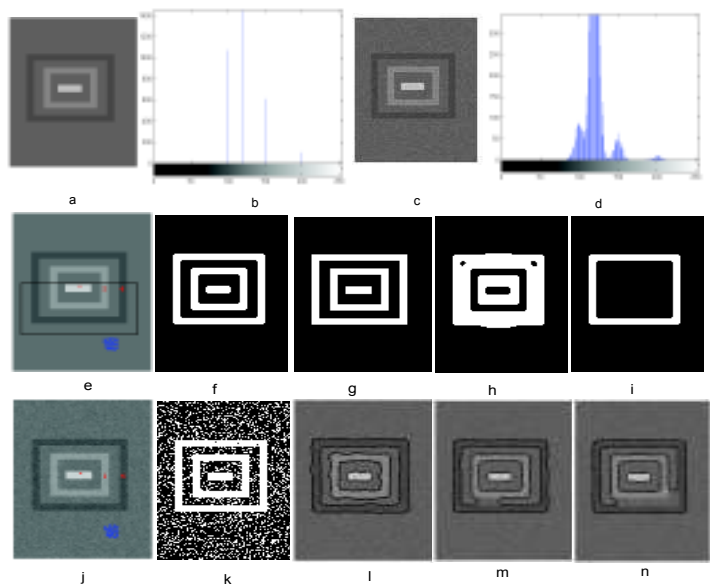

Figure 6.1: A synthetic image containing objects characterized by many different intensities and an overlap in the foreground/background intensity distributions, and a comparison among its segmentations obtained by the *SOAC* model and the *LRCV*, *C-V*, *KDE*-based, *GMM*-based, and *CSOM* models: (a) the original 90 × 122 image with the three different intensities 100, 150 and 200 in its foreground, and 120 in its background, and (b) its histogram; (c) the same image with the addition of Gaussian noise with standard deviation (*SD*) equal to 5, and (d) its histogram; (e) the original image in (a) with the addition of a rectangular initial contour (in black), and training examples (in red for the foreground, in blue for the background); (g) its ground truth, and its segmentation results obtained - starting from the initial contour in (e) - by (f) the *SOAC* model, (h) the *LRCV* model, and (i) the *C-V* model; (j) the noisy version of the same image, already shown in (c), with the addition of the initial contour and the training examples; its segmentation by (k) the *CSOM* model, (l) the *SOAC* model, (m) the *KDE*-based model, and (n) the *GMM*-based model.

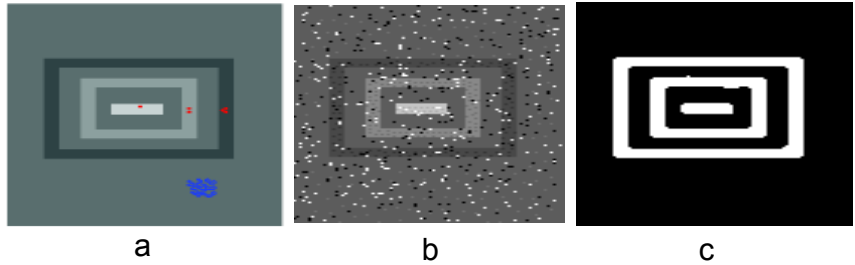

a          b          c

Figure 6.2: (a) the same synthetic image considered in Fig. 6.1, with the supervised training examples; (b) its noisy version, obtained by the addition of salt and pepper noise; (c) the segmentation result by *SOAC* model.

show the robustness of our model to increasing levels of noise, we adopt the Precision (*P*), Recall (*R*), and *F*-measure (*F*-m) metrics.

      As it can be observed obviously, *SOAC* and *CSOM* share a similar mechanism in discovering, in the training session, the underlying intensity distribution of a given image. However, *SOAC* differs from *CSOM* for the additional presence of its variational level set framework. For these reasons, in the following we compare the segmentation behaviors of *SOAC* and *CSOM*. As expected, such results show a smaller noise sensitivity of *SOAC*, when additive noise appears in the test image. More precisely, Fig. 6.4 shows the values assumed by the Precision, Recall, and *F*-measure metrics for the *SOAC* and *CSOM* models applied to the synthetic image shown in Fig. 6.3(a), corrupted by several levels of additive noise. Fig. 6.3(b) shows the corresponding ground truth, whereas Fig. 6.3(c) and Fig. 6.3(d) show, respectively, the training examples, and the initial contour used by the *SOAC* model. The performance of *SOAC* is compared with the *CSOM* classifier, relying on the same training data (Fig. 6.3(c)). As Fig. 6.4 illustrates, in the presence of additive noise *SOAC* achieved higher performance than *CSOM*, confirming that the *SOAC* model is less sensitive to additive noise than *CSOM*.

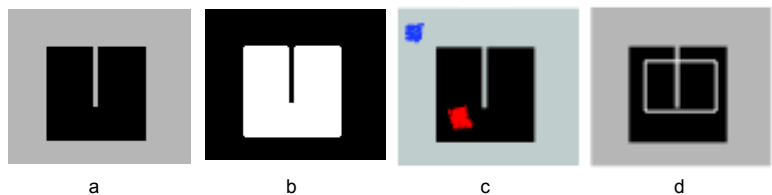

Figure 6.3: (a) A synthetic image; (b) its ground truth; (c) training examples (in red for the foreground, in blue for the background); (d) the initial contour used by the *SOAC* model (in white).

Table 6.1: For each image considered in the chapter: the reference from which it was taken (apart from the first artificial one), its size, and the associated number of foreground/background pixels used in the training phase.

| Fig. | Ref. | Image size | # Training Pixels | |
|---|---|---|---|---|
| | | | Foreground | Background |
| 6.1 (row 3) | - | 90 × 122 | 134 | 165 |
| 6.5 (row 1) | [109] | 118 × 93 | 160 | 172 |
| 6.5 (row 2) | [120] | 319 × 127 | 171 | 236 |
| 6.5 (row 3) | [5] | 300 × 225 | 300 | 557 |
| 6.5 (row 4) | [5] | 300 × 225 | 957 | 2881 |
| 6.5 (row 5) | [5] | 300 × 203 | 1349 | 1284 |
| 6.8 (row 1) | [120] | 152 × 128 | 995 | 151 |
| 6.8 (row 2) | [108] | 174 × 238 | 615 | 718 |
| 6.10 (row 1) | [123] | 481 × 321 | 713 | 1464 |
| 6.10 (row 2) | [6] | 481 × 321 | 731 | 1778 |
| 6.10 (row 3) | [6] | 481 × 321 | 897 | 2452 |

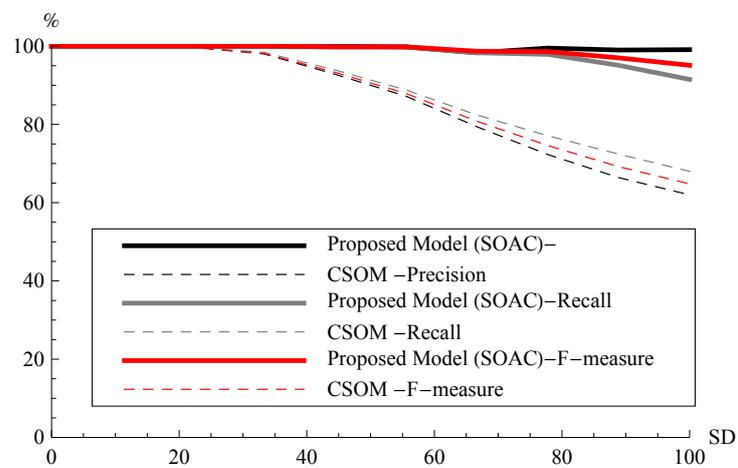

Figure 6.4: The sensitivity of *SOAC* to different levels of noise added to the image shown in Fig. 6.3(a) (Gaussian noise with *SD* = 10, 20, 30, 40, 50, 60, 70, 80, 90, and 100, respectively), in terms of Recall, Precision, and *F*-measure. For a comparison, also the case of *CSOM* is considered. The number of training pixels for both the foreground and the background is also shown.

Table 6.2: The Precision, Recall, and F-measure metrics for the *SOAC*, *KDE*-based, *GMM*-based, and *CSOM* models, applied to the images presented in the chapter.

| Fig. (row) | SOAC | | | KDE-based model | | | GMM-based model | | | CSOM model | | |
|---|---|---|---|---|---|---|---|---|---|---|---|---|
| | P(%) | R(%) | F-m.(%) | P(%) | R(%) | F-m.(%) | P(%) | R(%) | F-m.(%) | P(%) | R(%) | F-m.(%) |
| 6.1 (3)  | 87   | 100  | 93   | 94.7 | 99.4 | 96.9 | 85.1 | 99.8 | 90.6 | 54.6 | 99.6 | 70.5 |
| 6.5 (1)  | 100  | 88.3 | 93.7 | 79.5 | 94.5 | 86.3 | 76.9 | 99.3 | 81.3 | 99.6 | 92.1 | 95.7 |
| 6.5 (2)  | 98.6 | 98   | 98.2 | 46.1 | 100  | 63.1 | 80.2 | 100  | 70.6 | 100  | 92.1 | 95.9 |
| 6.5 (3)  | 63.4 | 90.3 | 74.5 | 6.9  | 99   | 12.9 | 5.2  | 98.6 | 10   | 39.2 | 95.4 | 55.6 |
| 6.5 (4)  | 81.9 | 98.8 | 89.5 | 88.5 | 98   | 93   | 87.4 | 98.9 | 92.8 | 47.8 | 97.9 | 72.7 |
| 6.5 (5)  | 87.9 | 96.2 | 91.9 | 86.3 | 91.6 | 88.9 | 89.7 | 93.2 | 91.4 | 80.3 | 95.9 | 87.4 |
| 6.8 (1)  | 95.6 | 95.7 | 95.6 | 62.7 | 100  | 77   | 70.2 | 99.9 | 73.4 | 96.8 | 86.4 | 91.3 |
| 6.8 (2)  | 96.9 | 97.2 | 97   | 92.8 | 23.4 | 37.3 | 97.1 | 69.8 | 53.9 | 95.2 | 85.9 | 90.3 |
| 6.10 (1) | 95   | 91.6 | 94.6 | 92.7 | 99.1 | 95.8 | 92.1 | 97.4 | 94.7 | 80.1 | 81.5 | 80.8 |
| 6.10 (2) | 86.9 | 83.8 | 85.3 | 5.8  | 100  | 11   | 9.6  | 99.9 | 17.6 | 79.4 | 94.4 | 86.2 |
| 6.10 (3) | 75   | 77.7 | 76.3 | 57.9 | 98.4 | 72.9 | 64.2 | 95.7 | 76.9 | 5.4  | 62.4 | 10   |

Fig. 6.5 illustrates the effectiveness of *SOAC* in handling different synthetic and real images. Fig. 6.5 shows images with blurred edges, noisy background, intensity inhomogeneity, and shadows. For the case of Fig. 6.5(row 1), among the considered models, only *SOAC* was able to find the whole object, while both the *KDE*-based model and the *GMM*-based model failed in the segmentation, due to the occurrence of a leaking problem. In Fig. 6.5(row 2), the *GMM*-based model had a similar performance to *SOAC*, while the *KDE*-based model failed to find accurately the object. Differently from the *SOAC* model, both the *KDE*-based model and the *GMM*-based model failed completely to find the object in Fig. 6.5(row 3), due to the increased amount of intensity inhomogeneity. However, they showed a similar performance to *SOAC* when handling the image shown in Fig. 6.5(row 4), while they failed again on the image in Fig. 6.5(row 5). For these images, the (unsupervised) *C-V* model generated good segmentations only in Fig. 6.5(row 2) and (row 4), while the *CSOM* model produced acceptable results only in Fig. 6.5(row 1) and (row 2). In any case, their performance was in general smaller than the one of the *SOAC* model (with the exception of Fig. 6.5(d), in which the *C-V* model had slightly better performance). Finally, we observe that, despite the rectangular contour initialization, some of the segmentations reported in Fig. 6.5 (and also in the next Fig. 6.10 for the case of *RGB* images) and obtained by the considered level set-based *ACM*s - especially the *KDE*-based and *GMM*-based models - contained holes. Their appearance depends on various reasons, such as the following: 1) since such models are used inside a variational level set-based framework, topological changes such as object merging and splitting - hence, also changes in the number of connected components - were allowed during the contour evolution; 2) the small number of training examples - compared to the complexity of such models - did not provide the *KDE*-based and *GMM*-based models the ability to drive the contour evolution in a proper way. As a consequence, during the contour evolution some of the foreground pixels were considered as background ones; 3) the sensitivity of such models to the contour initialization determined the extraction of undesirable objects during the contour evolution.

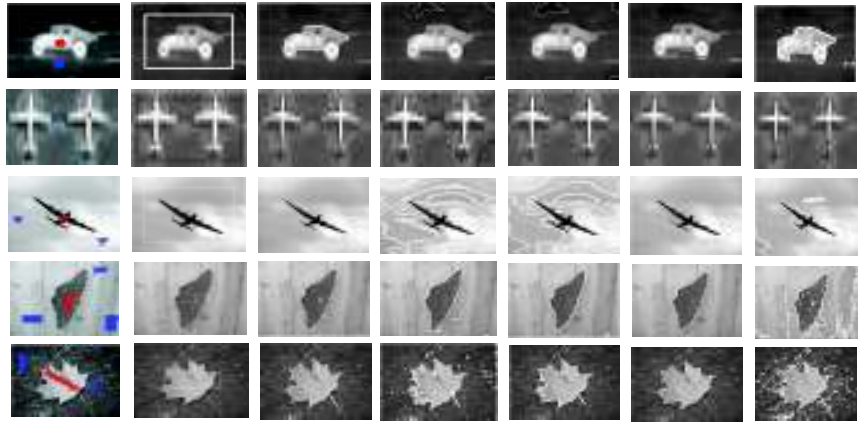

Figure 6.5: The segmentation results obtained on different real and synthetic images. Arranged in columns, from left to right: training examples (in red for the foreground, in blue for the background), the initial contour, and the segmentation results obtained, respectively by the *SOAC*, *KDE*-based, *GMM*-based, *C-V*, and *CSOM* models.

Fig. 6.6 shows the contour evolution over time for the *GMM*-based and *KDE*-based models, for some images considered in Fig. 6.5. For three cases in Fig. 6.6 (row 1 for the *GMM*-based model, and rows 3 and 4 for the *KDE*-based model) a leaking problem (i.e., in these cases, the appearance of holes in the final segmentation) occurred, while the final segmentation result was acceptable for the *GMM*-based model in row 2 (which refers to the same image in row 4, which was badly segmented, instead, by the *KDE*-based model).

To illustrate the robustness of the *SOAC* model with respect to the selection of the training examples, we trained the *SOAC* model with some foreground and background pixels belonging to the first image shown in Fig. 6.7. Then, in the on-line session, we applied the trained model to the remaining images shown in Fig. 6.7. As shown in the figure, the segmentations produced by the *SOAC* model in such a situation demonstrate its ability in finding

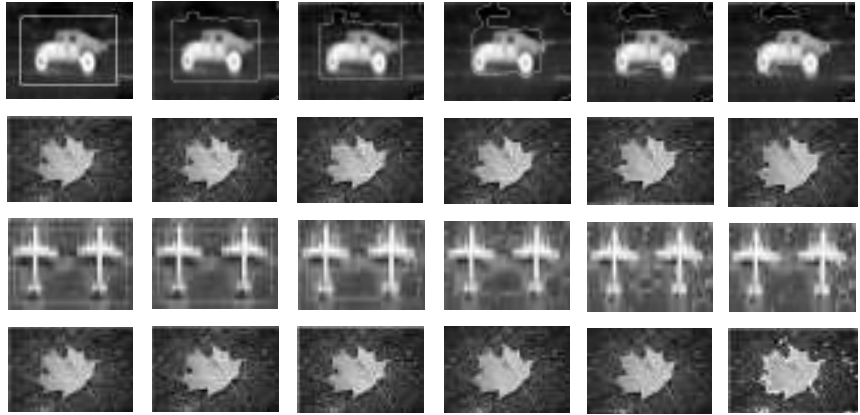

Figure 6.6: The contour evolution of the *GMM*-based and *KDE*-based models on some images in Fig. 6.5. Arranged in columns, from left to right: the initial contour, 4 intermediate contours, and the final contour. Arranged in rows: the contour evolutions of the *GMM*-based model for two images, and the ones of the *KDE*-based model for two images.

objects in test images different from the one used in the training phase.

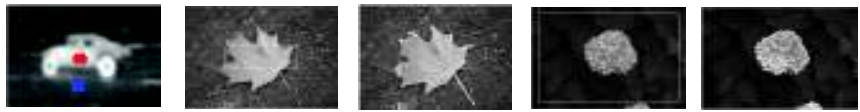

Figure 6.7: The sensitivity to the training pixels on some real images, taken from [13, 5]. From left to right: training examples (in red for the foreground, in blue for the background) of the first (training) image, the initial contour and the segmentation produced by *SOAC* for the second (test) image, and the initial contour and the segmentation produced by *SOAC* for the third (test) image.

To confirm the effectiveness of *SOAC* in handling images with intensity inhomogeneity, we tested *SOAC* on some real biomedical images. Fig. 6.8 illustrates the contour evolution for an increasing number of iterations, and confirms the ability of *SOAC* in handling images with intensity inhomogeneity, when a limited number of supervised examples is provided. On the other hand, the other supervised reference models considered in the comparison and trained on the same data failed to correctly segment the same images, as shown in Fig. 6.9.

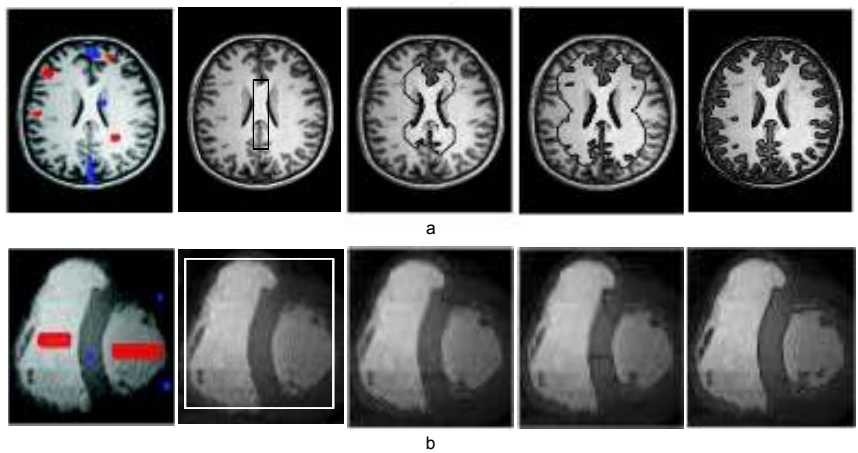

Figure 6.8: Segmentation results obtained by the *SOAC* model on two real images containing intensity inhomogeneity. Arranged in columns: the training examples used by the *SOAC* model (in red for the foreground, in blue for the background), respectively, for (a) a 174 × 238 brain image and (b) a 152 × 128 heart image; the initial contours used by the *SOAC* model for the two cases; the curve evolution at three successive stages of *SOAC*.

To illustrate the effectiveness of the extension of the *SOAC* model to the case of vector-valued images, we tested its performance in handling multi-spectral images. For the case considered in Fig. 6.10(first row) (an image with noisy background), the extended *SOAC* model and all the vectorial extensions of the supervised reference models were successfull in segmenting the im-age, apart from the vectorial extension of the *C-V* model, which was highly sensitive to the presence of noise. Fig. 6.10(second row) illustrates a comparison of the segmentation results obtained by the same models, when the multi-spectral image contained intensity inhomogeneity. In this case, only the vectorial extension of the *SOAC* model was able to accurately segment the image, whereas the *KDE*-based model showed unsatisfactory results, due to the small number of learning data, and the *C-V* model was not able to find the whole object, due to the overlap between the foreground/background intensity distributions. Fig. 6.10(third row) illustrates a comparison of segmentation results, when the multi-spectral image contained complicated overlaps in the foreground/background intensity distributions. Interestingly, the *CSOM* model failed completely to find the object of interest in Fig. 6.10(third row), due to the significant overlap in that image of the intensity distributions of the foreground and the background. However, the *SOAC* model (which, instead, embeds the concurrent *SOM*s in a variational level set framework) was able to find the same object.

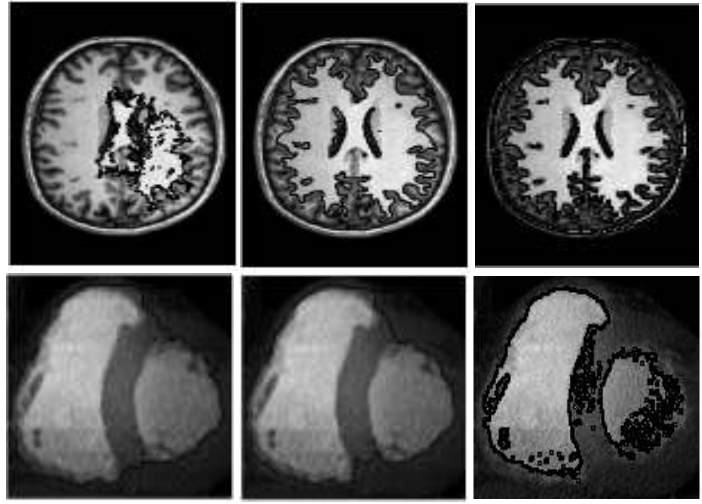

Figure 6.9: Segmentation results obtained for the real images shown in Fig. 6.8(a) and (b) by three supervised reference models, using the same training data as the *SOAC* model. Arranged in columns: the segmentation results obtained, respectively, by the *KDE*-based, *GMM*-based, and *CSOM* models.

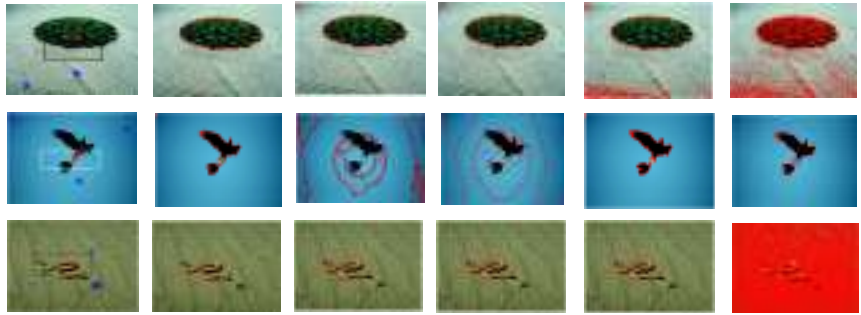

Figure 6.10: The segmentation results obtained on real multi-spectral images. Arranged in columns: three $481 \times 321$ real images with training examples (respectively, in red for the foreground, in blue for the background) and the initial contour (respectively, in black, white, and black); the segmentation results obtained on such images, respectively, by the vectorial versions of the *SOAC*, *KDE*-based, *GMM*-based, *C-V*, and *CSOM* models.

Table 6.2 shows the high accuracy of *SOAC* quantitatively in terms of Precision, Recall, and *F*-metrics, when compared with the proposed reference models, for the test images considered in this chapter. To keep the size of the table small, only the results of the supervised models are reported in the table. Additionally, to demonstrate the computational efficiency of *SOAC* when compared to the *KDE*-based model (which is a non-parametric model), Table 6.3 shows the *CPU* time in seconds and the final number of iterations for the two models, for the images considered in this experimental study. The three *RGB* images in Fig. 6.10 do not appear in the table since the convergence of the *KDE*-based model was too slow on them, because of their large size (due to the presence of three channels).

Fig. 6.11 provides a comparison of the pixel-by-pixel visual

Table 6.3: The *CPU* time and the number of iterations required by the *SOAC* model and the *KDE*-based model to segment the foreground for some images considered in this chapter.

| Fig. | SOAC model | | KDE-based model | |
|---|---|---|---|---|
| | CPU Time(s) | # Iterations | CPU Time(s) | # Iterations |
| 6.1 (row 3) | 15.6 | 32 | 18.65 | 57 |
| 6.5 (row 1) | 14.32 | 12 | 20.83 | 100 |
| 6.5 (row 2) | 52.83 | 20 | 86.04 | 18 |
| 6.5 (row 3) | 1.24 | 58 | 2.28 | 51 |
| 6.5 (row 4) | 4.54 | 100 | 8.18 | 240 |
| 6.5 (row 5) | 1.15 | 42 | 7.36 | 250 |
| 6.8 (row 1) | 54.3 | 24 | 269.03 | 31 |
| 6.8 (row 2) | 25.63 | 28 | 107.69 | 7 |

representations of the term

$$e_{SOAC}(x,C) := -\lambda^+ \left(I(x) - w_b^+(x,C)\right)^2 + \lambda^- \left(I(x) - w_b^-(x,C)\right)^2 \quad (6.14)$$

inside the level set formula of the *SOAC* model, and the ones of the terms

$$e_{KDE}(x), e_{GMM}(x) := \log p_{\text{in}}(I(x)) - \log p_{\text{out}}(I(x)) \quad (6.15)$$

inside the level set formula of the *KDE*-based and *GMM*-based models, for two selected images considered in the chapter. For simplicity of comparison, we have chosen two cases in which a small value of the parameter $\sigma$ was used, in order to have the two terms $w_b^+(x,C)$ and $w_b^-(x,C)$ not significantly influenced by the choice of the contour $C$, see formulas (6.3) and (6.4). As demonstrated in the figure, the term $e_{SOAC}$ in the *SOAC* model was able to identify accurately the actual pixels of the foreground and background, without being affected by the shadows in Fig. 6.11(first row) or by the intensity inhomogeneity and the overlap between the foreground/background intensity distributions in Fig. 6.11(second row). On the other hand, likely due to the small number of training examples (compared to the complexity of the models), both terms $e_{KDE}$ and $e_{GMM}$ in the *KDE*-based and *GMM*-based models failed to identify the pixels correctly.

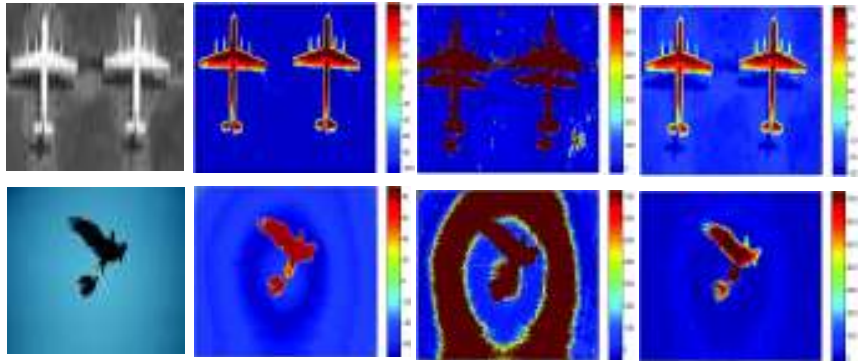

Figure 6.11: A comparison of the pixel-by-pixel visual representations of the term $e_{SOAC}$ in the *SOAC* model (formula (6.14)) and the terms $e_{KDE}$ and $e_{GMM}$ (formula (6.15)) in the *GMM*-based and *KDE*-based models, for two selected images considered in the chapter. Arranged in columns, from left to right: the original images, and the pixel-by-pixel visual representation of the terms $e_{SOAC}$, $e_{KDE}$ and $e_{GMM}$.

The above results clearly show that our proposed *Self Organizing Active Contour* (*SOAC*) model is an accurate, efficient and robust technique in image segmentation.

## 6.5  Summary

In this chapter, we have proposed a new supervised *ACM*, which we have termed *Self Organizing Active Contour* (*SOAC*) model. It is based on two sets of self organizing neurons to learn the dissimilarity between the intensity distributions of the foreground/background. In this way, the information about such distributions is integrated implicitly into the energy functional of the model by the learned prototypes of the two *SOM*s, helping in the guide of the contour evolution. *SOAC* is a Gaussian regularizing level set method, and it is robust to additive noise. The experimental results obtained on several synthetic and real images for both scalar-valued and vector-valued cases images demonstrate the high effectiveness and robustness of our model, when compared with state-of-the-art *ACM*s, in segmenting images with overlap between the foreground/background intensity distributions, intensity inhomogeneity, and/or containing objects characterized by many different intensities. In order to take advantage of *SOM* as an unsupervised visualization tool, in this thesis we describe also an unsupervised *SOM*-based *ACM*, termed *SOM-based Chan-Vese* (*SOMCV*) , which is presented in the following chapter.

# Chapter 7

# *SOM*-based Chan-Vese Model

## 7.1 Introduction

In this chapter, we present a novel active contour model, which we termed *SOM-based Chan-Vese* (*SOMCV*) [10]. It works by explicitly integrating the information coming from the prototypes of the neurons in a trained *SOM* to help choosing whether to shrink or expand the current contour during the optimization process, which is performed in an iterative way. Similarly to the *CSOM-CV* (chapter 5) and *SOAC* (chapter 6) models, *SOMCV* relies on a series of trained self-organizing neurons as a discriminative machine learning framework to approximate the image distribution. It also integrates their prototypes implicitly into the energy framework. However, *SOMCV* is presented as a global and unsupervised *SOM*-based *ACM*, which does not rely on training samples, differently from *CSOM-CV* and *SOAC*.

## 7.2 The *SOMCV* and $SOMCV_s$ models

In this section, we describe our *SOM*-based Chan-Vese (*SOMCV*) active contour model and its modification $SOMCV_s$. We first consider the case of scalar-valued images in Subsection 7.2.1.

Then, in Subsection 7.2.2, we briefly discuss the changes needed to deal with the case of vector-valued images. Finally, in Subsection 7.3, algorithmic details are provided.

### 7.2.1 The $SOMCV$ and $SOMCV_s$ models for scalar-valued images

Both the $SOMCV$ and $SOMCV_s$ segmentation frameworks for scalar-valued images are composed of two sessions: an unsupervised training session and a testing session, which are performed, respectively, off-line and on-line.

In the training session, after choosing a suitable topology of the $SOM$, the intensity $I^{(tr)}(x_t)$ of a randomly-extracted pixel $x_t$ of a training image is applied as input to the $SOM$ at time $t = 0, 1, \ldots, t_{\max}^{(tr)} - 1$, where $t_{\max}^{(tr)}$ is the number of iterations in the training of the $SOM$. Then, the neurons are trained in a self-organized way in order to be able to preserve the topological structure of the image intensity distribution at the end of training. Each neuron $n$ is connected to the input by a weight vector $w_n$ of the same dimension as the input (which - in this scalar case - is of dimension 1). After their random initialization, the weights $w_n$ are updated by the self-organization learning rule (as stated in formula (5.1)).

Since, after training, the inputs to the network are topologically arranged in the output map on the basis of the prototypes of the neurons that have the smallest distances from the inputs, we say that the learned prototypes have a global *Self-Organizing Topology Preservation* (*SOTP*) property, which allows one to represent the intensity distributions inside and outside the contour globally during the contour evolution.

Once the training of the $SOM$ has been accomplished, the trained network is applied on-line in the testing session, during the evolution of the contour $C$, to approximate and describe globally the foreground and background intensity distributions of a similar test image $I(x)$. Indeed, during the contour evolution, the two scalar intensities mean$(I(x)|x \in \text{in}(C))$ and mean$(I(x)|x \in \text{out}(C))$ are presented as inputs to the trained network. We now define, for each neuron $n$, the quantities

$$A_n^+(C) := |w_n - \text{mean}(I(x)|x \in \text{in}(C))|, \tag{7.1}$$
$$A_n^-(C) := |w_n - \text{mean}(I(x)|x \in \text{out}(C))|, \tag{7.2}$$

which are, respectively, the distances of the associated prototype $w_n$ from the mean intensities of the current approximations of the foreground and the background. Then, we define the two sets

$$\{w_j^+(C)\} := \{w_n : A_n^+(C) \leq A_n^-(C)\}, \tag{7.3}$$
$$\{w_j^-(C)\} := \{w_n : A_n^+(C) > A_n^-(C)\}, \tag{7.4}$$

of cardinalities $N^+(C) := |\{w_j^+(C)\}|$ and $N^-(C) := |\{w_j^-(C)\}|$, which are the sets of neurons whose prototypes are associated, respectively, with the current approximations of the foreground and the background. Such prototypes are chosen as representatives of the foreground and background intensity distributions according to their closeness to the two mean intensities. So, they are extracted as global regional intensity descriptors and included in the energy functional to be minimized in our proposed *SOMCV* model, which has the following expression:

$$\begin{aligned} E_{SOMCV}(C) &:= \lambda^+ \int_{\text{in}(C)} e^+(x,C)dx \\ &\quad + \lambda^- \int_{\text{out}(C)} e^-(x,C)dx, \end{aligned} \tag{7.5}$$

$$e^+(x,C) := \sum_{j=1,\ldots,N^+(C)} \left(I(x) - w_j^+(C)\right)^2, \tag{7.6}$$

$$e^-(x,C) := \sum_{j=1,\ldots,N^-(C)} \left(I(x) - w_j^-(C)\right)^2, \tag{7.7}$$

where the parameters $\lambda^+, \lambda^- \geq 0$ are, respectively, the weights of the two image energy terms $\int_{\text{in}(C)} e^+(x,C)dx$ and $\int_{\text{out}(C)} e^-(x,C)dx$, inside and outside the contour.

Now, we replace the contour curve $C$ with the level set function $\phi$, obtaining

$$E_{SOMCV}\left(\phi\right) = \lambda^+ \int_{\phi>0} e^+(x,\phi)dx + \lambda^- \int_{\phi<0} e^-(x,\phi)dx, \qquad (7.8)$$

where we have also made explicit the dependence of $e^+$ and $e^-$ on $\phi$. In terms of the Heaviside step function $H(\cdot)$, the *SOMCV* energy functional can be also written as follows:

$$\begin{aligned}E_{SOMCV}\left(\phi\right) &= \lambda^+ \int_{\Omega} e^+(x,\phi)H(\phi(x))dx \\ &+ \lambda^- \int_{\Omega} e^-(x,\phi)(1 - H(\phi(x)))dx. \end{aligned} \qquad (7.9)$$

Finally, proceeding likewise in Chapter 5 and 6, the evolution of the contour in the *SOMCV* model is described by the PDE

$$\frac{\partial \phi}{\partial t} = \delta\left(\phi\right)\left[-\lambda^+ e^+ + \lambda^- e^-\right], \qquad (7.10)$$

which shows how the trained neurons are used to determine the internal and external forces acting on the contour.

In the following, we describe also a simplification of the *SOMCV* model (which we term *SOMCV$_s$* model), which is based on an energy functional whose evaluation is easier from a computational point of view than the one of (7.5). This is obtained by replacing the sets $\{w_j^+(C)\}$ and $\{w_j^-(C)\}$ above by single prototypes $w_b^+$ and $w_b^-$, defined as follows:

$$w_b^+(C) := \operatorname{argmin}_n |w_n - \operatorname{mean}(I(x)|x \in \operatorname{in}(C))|, \qquad (7.11)$$
$$w_b^-(C) := \operatorname{argmin}_n |w_n - \operatorname{mean}(I(x)|x \in \operatorname{out}(C))|, \qquad (7.12)$$

where $w_b^+(C)$ is the prototype of the *BMU* neuron to the mean intensity inside the current contour, while $w_b^-(C)$ is the prototype of the *BMU* neuron to the mean intensity outside it. Then, we define the functional of the $SOMCV_s$ model as

$$E_{SOMCV_s}(C) := \lambda^+ \int_{in(C)} e_s^+(x,C)dx$$
$$+ \lambda^- \int_{out(C)} e_s^-(x,C)dx, \quad (7.13)$$
$$e_s^+(x,C) := \left(I(x) - w_b^+(C)\right)^2, \quad (7.14)$$
$$e_s^-(x,C) := \left(I(x) - w_b^-(C)\right)^2. \quad (7.15)$$

Then, proceeding as above, after replacing $C$ with the level set function $\phi$, the evolution of the contour is described by the PDE

$$\frac{\partial \phi}{\partial t} = \delta(\phi)\left[-\lambda^+ e_s^+ + \lambda^- e_s^-\right]. \quad (7.16)$$

Although the expressions of $e_s^+(x,C)$ and $e_s^-(x,C)$ are similar to those of the terms $(I(x)-c^+(C))^2$ and $(I(x)-c^-(C))^2$ used in the C-V model, the prototypes $w_b^+(C)$ and $w_b^-(C)$ may represent globally the two regional intensity distributions better than the mean intensities in the two regions. This can be shown in the following way: suppose that the current contour $C$ coincides with the actual object boundary, but that the image contains additive noise: then, the values of the mean regional intensities $c^+(C) := \text{mean}(I(x)|x \in in(C))$ and $c^-(C) := \text{mean}(I(x)|x \in out(C))$ depend on $C$ in a continuous way, likely making the contour evolve toward a worse approximation of the object boundary. Instead, the values of $w_b^+(C)$ and $w_b^-(C)$ may not change at all for small changes of $C$, providing more robustness of the model with respect to additive noise. In order to obtain such a behavior, one should keep the size of the network small. Otherwise, when using a network with a large number of neurons (then of propotypes), one may more likely obtain $w_b^+(C) \cong \text{mean}(I(x)|x \in in(C))$ and $w_b^-(C) \cong \text{mean}(I(x)|x \in out(C))$, losing the just-mentioned robustness.

Moreover, when the foreground/background intensity distributions are characterized by many different intensities, minimizing the functional of the C-V model - in which the dependence on the foreground/background intensity distributions is expressed only in terms of the mean regional intensities $c^+(C)$ and $c^-(C)$ - may result in under(over)-segmentation problems. Of course, such problems are still not solved by replacing $c^+(C)$ and $c^-(C)$ with the prototypes $w_b^+(C)$ and $w_b^-(C)$, since also $w_b^+(C)$ and $w_b^-(C)$ are only scalar quantities. So, in the case of skewness/multimodality of the two distributions, one expects better segmentation results when using the functional (7.5) of the *SOMCV* model, which represents the foreground/background intensity distributions by larger sets of weights for each of the two regions, as compared to the functional (7.13) of the $SOMCV_s$ model.

In Section 7.4, the robustness of the proposed model to additive noise and to intensity distributions characterized by many intensity values is investigated experimentally.

### 7.2.2 The *SOMCV* and $SOMCV_s$ models for vector-valued images

The *SOMCV* and $SOMCV_s$ models can be extended to the case of vector-valued images. Such an extension is particularly useful for the segmentation of multi-spectral images (see Section 7.4 for some related experiments). In the vectorial case, the image $\mathbf{I}(x)$ is made up of $D$ channels $I_i(x)(i = 1, ..., D)$, and also the *SOM* weights are vectors of dimension $D$. The only significant change with respect to the scalar case described in Subsection 7.2.1 is that, in the determination of the *BMU* neuron, the absolute values in formulas (7.1) and (7.2) are replaced by Euclidean norms in $\mathbb{R}^D$.

## 7.3 Implementation

Having discussed the formulations of the *SOMCV* and $SOMCV_s$ models, in the following, the procedural steps of their training and testing sessions are summarized in Algorithm 4 (to avoid redundancy, only the case of scalar-valued images is detailed here).

## Algorithm 3 $SOMCV$ and $SOMCV_s$ segmentation frameworks for scalar-valued images

1: **procedure**
- Input:
  - Training and test scalar-valued images.
  - Topology of the network (with 1-dimensional prototypes).
  - Number of iterations $t_{\max}^{(tr)}$ for training the neural map.
  - Maximum number of iterations $t_{\max}^{(evol)}$ for the contour evolution.
  - $\eta_0 > 0$: starting learning rate.
  - $r_0 > 0$: starting radius of the map.
  - $\tau_\eta, \tau_r > 0$: time constants in the learning rate and contour smoothing parameter.
  - $\lambda^+, \lambda^- \geq 0$: weights of the energy terms, respectively, inside and outside the contour.
  - $\sigma > 0$: Gaussian contour smoothing parameter.
  - $\rho > 0$: constant in the binary approximation of the level set function.
- Output:
  - Segmentation result.

*TRAINING SESSION:*

2:    Initialize randomly the prototypes of the neurons.
3:    **repeat**
4:       Choose randomly a pixel $x_t$ in the image domain $\Omega$ and determine the *BMU* neuron to the input intensity $I^{(tr)}(x_t)$.
5:       Update the prototypes $w_n$ using (5.1), (5.2), (5.3), and (5.4).
6:    **until** learning of the prototypes is accomplished (i.e., the number of iterations $t_{\max}^{(tr)}$ is reached).

*TESTING SESSION:*

7:    Choose a subset $\Omega_0$ (e.g., a rectangle) in the image domain $\Omega$ with boundary $\Omega_0'$, and initialize the level set function as:

$$\phi(x) := \begin{cases} \rho, & x \in \Omega_0 \setminus \Omega_0', \\ 0, & x \in \Omega_0', \\ -\rho, & x \in \Omega \setminus (\Omega_0 \cup \Omega_0'). \end{cases} \tag{7.17}$$

8:    Choose the functional to be minimized (the $E_{SOMCV}$ functional (7.5) or the $E_{SOMCV_s}$ functional (7.13)).
9:    **repeat**
10:       **if** $E_{SOMCV}$ functional (7.5) has been chosen **then**
11:          Determine, for each neuron, the quantities $A_n^+$ and $A_n^-$ from (7.1) and (7.2), then the sets $\{w_j^+\}$ and $\{w_j^-\}$ from (7.3) and (7.4).
12:          Evolve the level set function $\phi$ according to a finite difference approximation of (8.14).
13:       **else**
14:          Calculate $w_b^+$ and $w_b^-$ from (7.11) and (7.12).
15:          Evolve the level set function $\phi$ according to a finite difference approximation of (7.16).
16:       **end if**
17:       At each iteration of the finite-difference scheme, re-initialize the current level set function to be binary by performing the update

$$\phi \leftarrow \rho\left(H(\phi) - H(-\phi)\right), \tag{7.18}$$

then regularize by convolution the obtained level set function:

$$\phi \leftarrow g_\sigma \otimes \phi, \tag{7.19}$$

where $g_\sigma$ is a Gaussian kernel with $\int_{\mathbb{R}^2} g_\sigma(x)dx = 1$ and width $\sigma$.

18:    **until** the curve evolution converges (i.e., the curve does not change anymore) or the maximum number of iterations $t_{\max}^{(evol)}$ is reached.
19: **end procedure**

Algorithm 4 can be explained as follows. Once the topology of the neural map is defined (e.g., a 2×2 or a 3×3 square grid in the experiments described in Section 7.4, see Tables 7.3 and 7.4), the neurons of the map start to be trained using a learning algorithm composed of a competitive phase and a cooperative one (see formula (5.1)). As a result, through the prototypes of the neurons, the set of the trained neurons carries significant information about the intensity distribution of the given image, which reflects the topological structure of the intensity distribution. Once the training is accomplished, the prototypes of selected neurons in the case of the functional (7.5) of the *SOMCV* model - or of the best-matching neurons to the mean intensities in the two regions, in the case of the functional (7.13) of the *SOMCV$_s$* model - are used as global regional descriptors for the foreground and background intensity distributions. Then, in the testing phase, they are used as core components of the level set energy functional to guide the evolution of the contour. Moreover, in order to keep the contour and the level set function smooth at each iteration without losing information on the displacement of the current contour, the current level set function $\phi$ is first re-initialized to be binary, then convolved with a Gaussian kernel function. The smoothness degree of the updated level set function is controlled by the width parameter $\sigma$ of the Gaussian as described in Subsection 7.2.1.

Fig. 7.1 illustrates the off-line and on-line components of the *SOMCV* and *SOMCV$_s$* models in a vector-valued (more specifically, *RGB*) image segmentation framework (the scalar case is similar, but uses scalar prototypes and preferably a 1-*D* grid). Fig. 7.1(a) shows the input layer of the *SOM*, whose dimension is equal to the one of the voxel intensities of the image to be segmented. For example, in the case of *RGB* images, the input layer of the map has dimension 3, since it receives the *R*, *G*, and *B* channels of the vector-valued image. The red cube in Fig. 7.1(a) represents a voxel intensity presented as input to the *SOM*, in this case made up of $3 \times 3$ neurons (Fig. 7.1(b)). The small circles in Fig. 7.1(b) represent the neurons of the map, where each neuron is associated with a three-dimensional prototype, of the same dimension as the input. The prototypes of the neurons are modified during the training phase. This is ac-complished by finding the best-matching neuron (the blue circle in Fig. 7.1(b)) to each input voxel intensity, and updating its prototype and the ones of all its neighbors as described in formulas (3.1), (5.2), (5.3), and (5.4), extended to the three-dimensional case as described in Subsection 7.2.2. Once the learning is accomplished, the prototypes associated with selected neurons of the learned map (Fig. 7.1(b)) are ready to be integrated into the energy functional (7.5) during the on-line session (i.e., during the curve evolution process) as global regional intensity descriptors. Fig. 7.1(d) represents a test image to be segmented (the gray circle represents the foreground). Starting from an initial contour (the black curve in Fig. 7.1(d)), the mean intensities of inside and outside the contour are presented as inputs to the learned map in Fig. 7.1(c) to classify (see Fig. 7.1(e), top) the prototypes associated with the neurons into foreground (in red) and background (in black) global intensity descriptors. Then, the contour evolution is guided by the extracted prototypes associated with the two sets of foreground and background neurons. In the case of $SOMCV_s$ (see Fig. 7.1(e), down), only one prototype is used as a global intensity descriptor for each region.

## 7.4 Experimental study

In this section, we demonstrate the effectiveness and robustness of the $SOMCV$ and $SOMCV_s$ models, compared to the *C-V* model described in chapter 2, in handling real and synthetic images. For a fair comparison, the $SOMCV$, $SOMCV_s$ and the *C-V* model used in this experiment are all implemented in Matlab R2012a on a PC with the following configuration: 2.5 GHz Intel(R) Core(TM) 2 Duo, and 2.00 GB RAM. In each experiment, the $r_0$ and $\sigma$ parameters are expressed in pixels. Moreover, the $SOMCV$ and $SOMCV_s$ parameters are fixed[1] as follows: $\eta_0 = .9$, $\sigma = 1.5$, and the weight parameters (i.e., $\lambda^+$, $\lambda^-$ for the scalar-valued case, and $\lambda_i^+$, $\lambda_i^-$ in the vector-valued case) are fixed to

---

[1] In the experiments presented in Fig. 7.3 and 7.7, also the choice $\sigma = .5$ was considered, together with $\sigma = 1.5$.

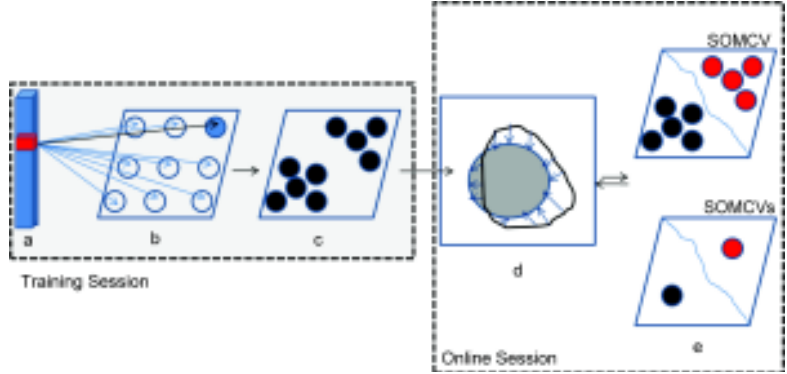

Figure 7.1: The architecture of *SOMCV* for *RGB* images: (a) the input intensities of a training voxel; (b) a $3 \times 3$ *SOM* neural map (with a three-dimensional prototype associated with each neuron); (c) the trained *SOM*; (d) the contour evolution process; and (e) the foreground (in red) and background (in black) representative neurons for the *SOMCV* (top) and the *SOMCV$_s$* (down) models. For a scalar-valued image, a similar model is used, but the prototypes have dimension 1, and a 1-*D* grid is used.

1. Also, $r_0 := \max(M, N)/2$, where $M$ and $N$ are the numbers of rows and columns of the installed neural map, $t_{\max}^{(tr)} = 10000$, $t_{\max}^{(evol)} = 1000$, $\tau_\eta := t_{\max}^{(tr)}$, $\tau_r := t_{\max}^{(tr)}/\ln(r_0)$, $\rho = 1$. For the experiments performed on the scalar-valued images considered in the chapter, the *SOM* network has been chosen as a 1-*D* neural map composed of 5 neurons (i.e., $M = 5$ and $N = 1$), whereas for the case of vector-valued images, it was a $3 \times 3$ grid of neurons in most experiments ($M = N = 3$) and a $2 \times 2$ grid ($M = N = 2$) for the other experiments (see Tables 7.3 and 7.4). In the *C-V* model, $\lambda^+$, $\lambda^-$ for the scalar-valued case and $\lambda_i^+$, $\lambda_i^-$ in the vector-valued case are also fixed to 1, $\mu$ is chosen such that the final contour is smooth enough and $\nu = 0$ (as made in [27, p. 268]). Moreover, in the comparison, the *SOMCV$_s$* model is considered with the same parameters of the *SOMCV* model. Unless stated otherwise, the training image used in the unsupervised training session coincides with the test image. Otherwise, it is an image similar to the test image (obtained, e.g., by adding Gaussian noise). In all the testing sessions, the initial contour has been chosen as rectangular. For the case of gray-level images, the range of the values assumed by the intensity is 0-255 as all the considered gray-level images are 8-bit images.

Fig. 7.2 illustrates the fast convergence of *SOMCV* (and its variation *SOMCV$_s$*) for scalar-valued images and the associated contour evolution process when compared to the *C-V* model. As Fig. 7.2 shows, the final contours obtained by the *SOMCV* and *SOMCV$_s$* models converge with similar numbers of performed iterations and similar performances because of the large intensity homogeneity of the image considered in the experiment.

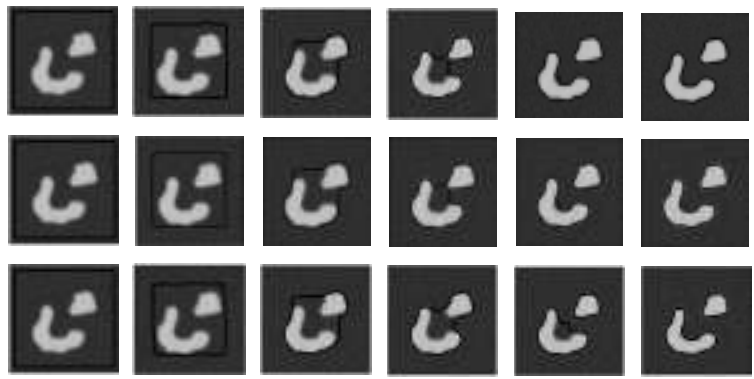

Figure 7.2: The rapid contour evolution of the *SOMCV* and *SOMCV$_s$* models when compared to the contour evolution of the *C-V* model, in the scalar case. The first and second rows show, respectively, the contour evolution of *SOMCV* and *SOMCV$_s$*. From left to right: initial contour (in black), contour after 3, 6, 9, 12 iterations, and final contour (15 iterations). The third row shows the contour evolution of the *C-V* model. From left to right: initial contour (in black), contour after 50, 100, 150, 200 iterations, and final contour (260 iterations).

Fig. 7.3 illustrates the effectiveness and robustness of *SOMCV* in handling images containing objects characterized by many different intensities and skewness/multimodality of the foreground intensity distribution, in the presence of noise. Compared to the *C-V* model, the *SOMCV* model shows better results, due to its automated ability to preserve the topological structure of the foreground intensity distribution (this is not needed, instead, for the background distribution, which is simpler). Moreover, the segmentation performance of the *SOMCV$_s$* model is quite similar to the one of the *SOMCV* model but more sensitive to the noise.

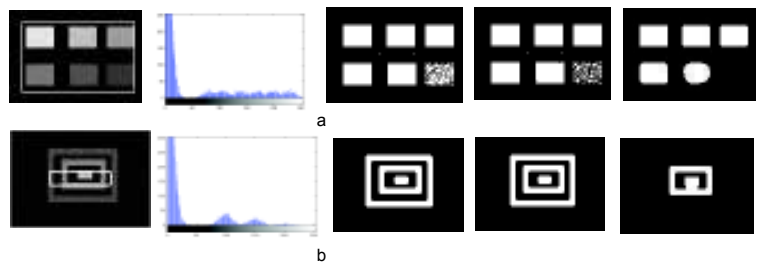

Figure 7.3: The effectiveness of the *SOMCV* model in dealing with objects characterized by many different intensities and skewness/multimodality of the foreground intensity distribution. Arranged in rows there are: (a) a noisy $140 \times 100$ image (with Gaussian noise added, standard deviation $SD = 10$) with six different intensities 80, 100, 140, 170, 200, and 230 in its foreground; (b) a noisy $90 \times 122$ image (with Gaussian noise added, standard deviation $SD = 10$) with three different intensities 100, 150, and 200 in its foreground. The columns from left to right are: the images with the additions of the initial contours, the histograms of the intensities of the images, and, respectively, the segmentation results of the *SOMCV*, *SOMCV$_s$* ($\sigma = .5, 1.5$, respectively, for (a) and (b)), and *C-V* models.

Then, in order to demonstrate the robustness of *SOMCV* and *SOMCV$_s$* to the additive noise, in the experiment described in Fig. 7.4 we have used the top left image of Fig. 7.3 in the training session of *SOMCV* and *SOMCV$_s$*, then the trained *SOM* (whose values of the weights are common to the two models) has been applied on-line to various test images obtained adding to such an image different levels of Gaussian noise. As shown in Fig. 7.4, for this case *SOMCV* is more robust and less sensitive to the additive noise than *SOMCV$_s$*, since the regions of the foreground are detected more accurately by *SOMCV*.

Similarly, the image of Fig. 7.3(b) has been used in the training session of *SOMCV* and *SOMCV$_s$*, then the trained *SOM* has been applied on-line to various test images obtained by adding to such an image different levels of Gaussian noise, as shown in Fig. 7.5. The results of these two experiments show the ability of *SOMCV* to find all the different regions of the object (which is characterized by many different intensities), and also its robustness to the additive noise and to the skewness/multimodality of the foreground intensity distribution. They also demonstrate that, in the case of images containing objects characterized by many different intensities or by skewed/multimodal intensity distributions, *SOMCV* usually produces better results than *SOMCV$_s$*.

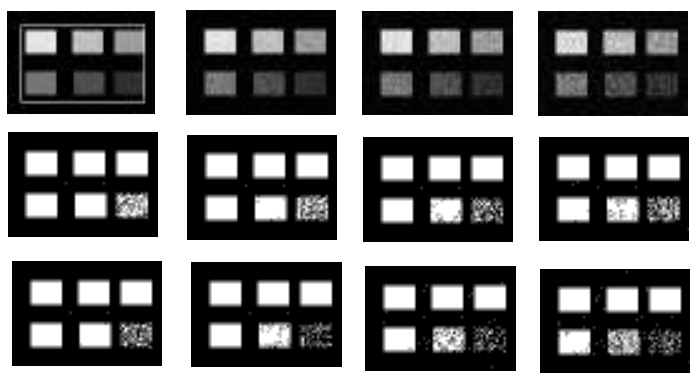

Figure 7.4: The robustness of the *SOMCV* and *SOMCV$_s$* models to the additive noise: the first row shows, from left to right, the image of Fig. 7.3(a) with the addition of different Gaussian noise levels (standard deviation $SD$ = 10, 15, 20, and 25, respectively); the second and third rows show, respectively, the corresponding segmentation results of *SOMCV* and *SOMCV$_s$*.

Fig. 7.6 illustrates the effectiveness of *SOMCV* in handling real and synthetic scalar-valued images. The segmentation results of the *SOMCV* model on the real images shown in the first and second columns show the ability of *SOMCV* to segment objects with blurred edges and background, while the *C-V* model provides a worse segmentation for the image in the first column, and incurs in an under-segmentation problem for the image in the second column. Similarly, *SOMCV* outperforms *C-V* also in handling synthetic images as shown in the third and fourth columns. Moreover, *SOMCV* and *SOMCV$_s$* behave exactly the same as *C-V* in handling binary gray images as in the case of the image shown in the right-most column. This is because in this case the mean intensities inside and outside the contour are accurate enough to approximate the foreground/background intensity distributions. For the images presented in Fig. 7.6, *SOMCV* outperforms also *SOMCV$_s$*.

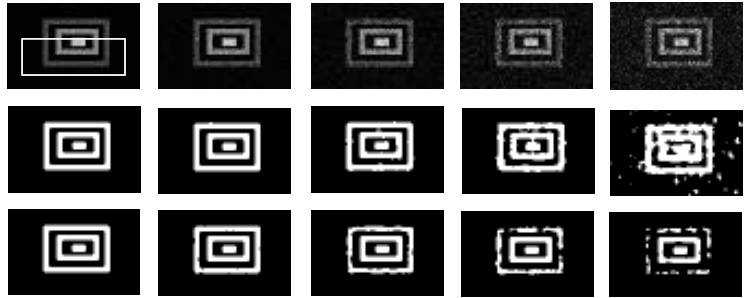

Figure 7.5: The robustness of the *SOMCV* and *SOMCV$_s$* models to the additive noise: the first row shows, from left to right, the image of Fig. 7.3(b) with the addition of different Gaussian noise levels (standard deviation *SD* = 10, 20, 30, 40, and 50, respectively); the second and third rows show, respectively, the corresponding segmentation results of *SOMCV* and *SOMCV$_s$*.

To illustrate the effectiveness of *SOMCV* and its variation *SOMCV$_s$* in handling real and synthetic vector-valued images, we have tested the extension of *SOMCV* and *SOMCV$_s$* to the vectorial framework on *RGB* real and synthetic images, which is shown in Fig. 7.7 in comparison with the vectorial *C-V* model from [27]. The segmentation results of *SOMCV* are similar to the ones of *C-V* in handling the image shown in the fourth column, while *SOMCV* outperforms *C-V* in all the other shown images. For these images, *SOMCV* outperforms also *SOMCV$_s$*, which, however, provides better results than *C-V*, apart from the cases of the images considered in the first two columns, for which the results are similar.

In the following, we provide also a quantitative study to confirm the effectiveness of *SOMCV* and *SOMCV$_s$*, when compared to *C-V*. To demonstrate quantitatively the accuracy of the

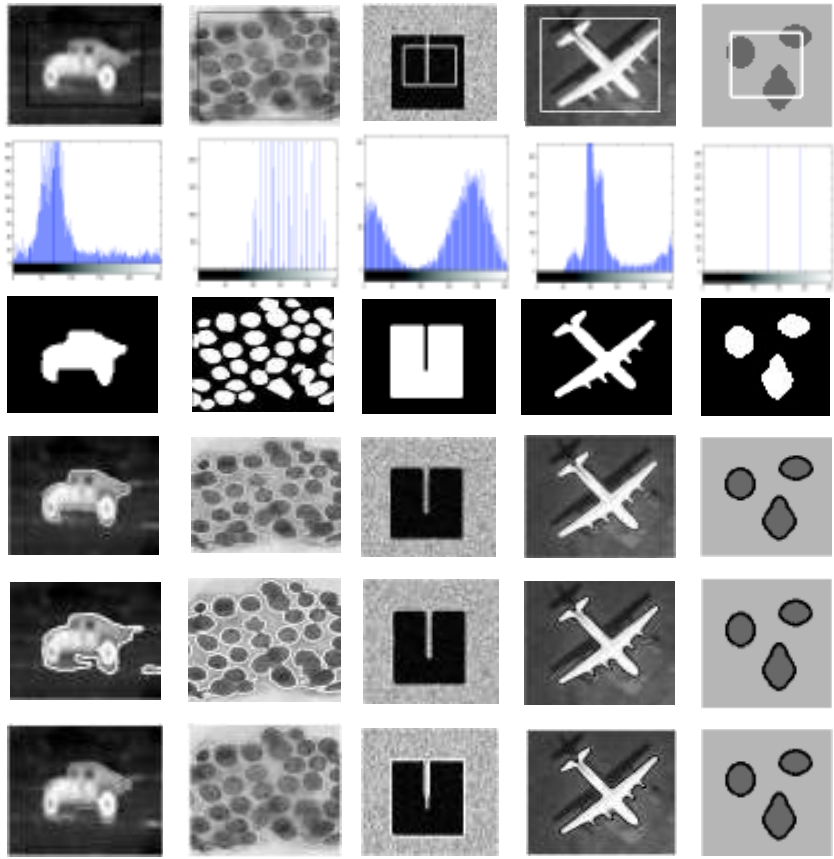

Figure 7.6: The segmentation results obtained on real and synthetic scalar-valued images. The first, second and third row show the original images with the initial contours, the histograms of the image intensities and their ground truth, respectively, while the fourth, fifth, and sixth rows show, respectively, the corresponding segmentation results of the *SOMCV*, *SOMCV$_s$* and *C-V* models.

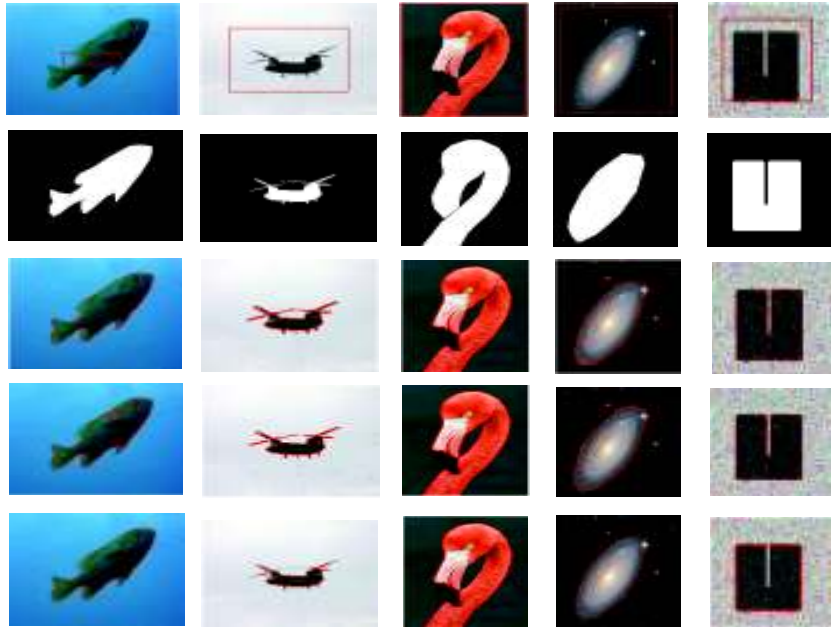

Figure 7.7: The segmentation results on real images from [5, 13], and synthetic vector-valued images. The first and second rows show the original images with the initial contours, respectively, while the third, fourth, and fifth rows show, respectively, the corresponding segmentation results of the vectorial versions of the *SOMCV*, *SOMCV$_s$* and *C-V* models. Note that $\sigma$ = .5 has been used by *SOMCV* and *SOMCV$_s$* for the image in the second column.

*SOMCV* and *SOMCV$_s$* models in segmenting the images shown in Fig. 7.6 and 7.7, we have also compared the obtained segmentation results with their corresponding ground-truth data by adopting the Precision (*P*), Recall (*R*), and *F*-measure metrics.

Tables 7.1 and 7.2 illustrate the high segmentation accuracy of the *SOMCV* model and its variation *SOMCV$_s$* when compared to the *C-V* model, in terms of the three metrics defined above. As the two tables illustrate, the *SOMCV* model has shown a better performance than the *C-V* model in both the scalar and vectorial cases and for all the tested images used, respectively, in Fig. 7.6 and 7.7. Moreover, the $SOMCV_s$ model has usually shown a similar performance as the $SOMCV$ model.

Table 7.1: The Precision, Recall, and $F$-measure metrics for the scalar $SOMCV$, $SOMCV_s$ and $C\text{-}V$ models in the segmentation of the scalar images shown in Fig. 7.6.

| Image in | $SOMCV$ | | | $SOMCV_s$ | | | $C\text{-}V$ | | |
|---|---|---|---|---|---|---|---|---|---|
| | $P$ (%) | $R$ (%) | $F$-m (%) | $P$ (%) | $R$ (%) | $F$-m (%) | $P$ (%) | $R$ (%) | $F$-m (%) |
| column 1 | 98.8 | 99.9 | 99.3 | 75.5 | 100 | 86 | 91.8 | 83.3 | 87.4 |
| column 2 | 60.6 | 98.5 | 75 | 60.6 | 98.5 | 75 | 42.7 | 98.5 | 59.6 |
| column 3 | 100 | 100 | 100 | 100 | 100 | 100 | 99.2 | 88 | 93.3 |
| column 4 | 96.3 | 99.3 | 97.8 | 98.8 | 98.4 | 98.6 | 96.5 | 96.4 | 96.4 |
| column 5 | 100 | 100 | 100 | 100 | 100 | 100 | 99 | 100 | 99.5 |

Table 7.2: The Precision, Recall, and $F$-measure metrics for the vectorial $SOMCV$, $SOMCV_s$ and $C\text{-}V$ models in the segmentation of the $RGB$ images shown in Fig. 7.7.

| Image in | $SOMCV$ | | | $SOMCV_s$ | | | $C\text{-}V$ | | |
|---|---|---|---|---|---|---|---|---|---|
| | $P$ (%) | $R$ (%) | $F$-m. (%) | $P$ (%) | $R$ (%) | $F$-m. (%) | $P$ (%) | $R$ (%) | $F$-m. (%) |
| column 1 | 89.6 | 96.8 | 93 | 91.3 | 91.8 | 91.5 | 94.7 | 83.1 | 88.5 |
| column 2 | 71.7 | 97.6 | 82.7 | 72.3 | 97.3 | 82.9 | 84.5 | 81.9 | 83.2 |
| column 3 | 94.4 | 90.1 | 92.2 | 95 | 89 | 91.9 | 89.5 | 88.9 | 89.2 |
| column 4 | 96.1 | 85.5 | 90.5 | 93.5 | 91.7 | 92.6 | 96.1 | 86.9 | 91.3 |
| column 5 | 99.6 | 100 | 99.8 | 100 | 100 | 100 | 96.8 | 89.6 | 93.1 |

To demonstrate the computational efficiency of the $SOMCV$ and $SOMCV_s$ models when compared to the $C\text{-}V$ model, Table 7.3 shows, for each of the three methods, the $CPU$ time (in seconds) that was required for the contour evolution (i.e., the time required in the testing session) and the number of iterations performed before convergence for the real and synthetic images used in Fig. 7.6. Moreover, the computational effectiveness of the vectorial versions of $SOMCV$ and $SOMCV_s$ with respect to the vectorial $C\text{-}V$ model is illustrated in Table 7.4 for the $RGB$ images in Fig. 7.7 by showing, for all methods, the $CPU$ times and the number of iterations required in the testing session (note that, in the common training session of $SOMCV$ and $SOMCV_s$, the $CPU$ time is fixed by the number of iterations $t_{\max}^{(tr)}$). The sizes of the training and test scalar-valued and vector-valued images are also listed in the two tables. From these tables, we can observe that the *SOMCV* and *SOMCV$_s$* models were much faster than the *C-V* model in all the listed cases, as the contour evolution for *SOMCV* and *SOMCV$_s$* required less iterations to converge than for the *C-V* model, and also the computational time per iteration for the *SOMCV* and *SOMCV$_s$* models was smaller than the one for the *C-V* model. This is due to the fact that *SOMCV* and *SOMCV$_s$* models are Gaussian Regularizing Level Set Models, whereas the original *C-V* model has not this feature.

Concluding, the results shown in Tables 7.1-7.4 highlight several advantages of the *SOMCV* and *SOMCV$_s$* models with respect to the *C-V* model.

Table 7.3: The contour evolution time and number of iterations required by the *SOMCV*, *SOMCV$_s$*, and *C-V* models to segment the foreground for the scalar-valued images shown in Fig. 7.6.

| Image in | Image size | SOM topology | SOMCV CPU Time (s) | # Iterations | SOMCV$_s$ CPU Time (s) | # Iterations | C-V CPU Time (s) | # Iterations |
|---|---|---|---|---|---|---|---|---|
| Column 1 | 118 × 93 | 5 | 0.03 | 10 | 0.01 | 9 | 6.22 | 137 |
| Column 2 | 256 × 256 | 5 | 1.0 | 30 | 0.73 | 30 | 104.2 | 406 |
| Column 3 | 114 × 101 | 5 | 0.14 | 16 | 0.1 | 16 | 5.6 | 100 |
| Column 4 | 135 × 125 | 5 | 0.15 | 16 | 0.15 | 16 | 13.1 | 266 |
| Column 5 | 64 × 61 | 5 | 0.03 | 7 | .01 | 7 | 4.38 | 97 |

Table 7.4: The contour evolution time and number of iterations required by the *SOMCV*, *SOMCV$_s$*, and *C-V* models to segment the foreground for the vector-valued images shown in Fig. 7.7.

| Image in | Image size | SOM topology | SOMCV CPU Time (s) | # Iterations | SOMCV$_s$ CPU Time (s) | # Iterations | C-V CPU Time (s) | # Iterations |
|---|---|---|---|---|---|---|---|---|
| Column 1 | 300 × 225 | 2 × 2 | 1.4 | 20 | 1.2 | 20 | 43 | 356 |
| Column 2 | 300 × 225 | 3 × 3 | 3.08 | 37 | 2.57 | 37 | 46 | 400 |
| Column 3 | 300 × 451 | 3 × 3 | 14.35 | 80 | 12.19 | 80 | 588.4 | 551 |
| Column 4 | 272 × 297 | 3 × 3 | 2.9 | 25 | 2.79 | 27 | 212.3 | 1612 |
| Column 5 | 114 × 101 | 2 × 2 | .9 | 16 | .4 | 16 | 4.6 | 78 |

In order to compare our *SOMCV* model with some representative global pixel-based segmentation techniques, we have applied the Otsu's method [78] and the multi-threshold Otsu's method [63] to some of the scalar-valued images considered in this chapter. Such methods belong to the class of thresholding image segmentation methods, as they segment a scalar-valued image by comparing the pixel intensity with one or multiple thresholds, respectively. The main reason for selecting the Otsu's method is that its threshold is chosen in such a way to optimize a trade-off between the maximization of the inter-class variance (i.e., between pairs of pixels beloging to the foreground and the background, respectively) and the minimization of the intra-class variance (i.e., between pairs of pixels belonging to the same region). The multi-threshold the Otsu's method is similar but uses more thresholds, segmenting the image in more than 2 regions. Fig. 7.8 shows the segmentation results obtained by the Otsu's method (second row) and the multi-threshold Otsu's method (third row) on some of the scalar-valued images considered in this chapter. For a fair comparison, in the case of multi-threshold Otsu's method we have also merged some of the objects found for different numbers of thresholds (as shown in the fourth row), then we have applied the classical Otsu's method to the resulting image (fifth row). As illustrated by Fig. 7.8, the Otsu's and multi-threshold Otsu's methods demonstrated to be more sensitive to noise than our proposed *SOMCV* model. As an additional drawback, post-processing operations were also required for the multi-threshold Otsu's method. The quantitative results corresponding to Fig. 7.8 are reported in Table 7.5.

Table 7.5: The Precision, Recall, and *F*-measure metrics for the Otsu's method and the multi-threshold Otsu's method (with post-processing) in the segmentation of the images shown in Fig. 7.8 (second and fifth rows, respectively) compared with the *SOMCV* model (sixth row).

| Image in | Otsu's method | | | multi-threshold Otsu's method | | | SOMCV | | |
|---|---|---|---|---|---|---|---|---|---|
| | *P* (%) | *R* (%) | *F*-m (%) | *P* (%) | *R* (%) | *F*-m (%) | *P* (%) | *R* (%) | *F*-m (%) |
| column 1 | 97.7 | 98 | 97.8 | 100 | 64.3 | 78.3 | 98.8 | 99.9 | 99.3 |
| column 2 | 100 | 78.8 | 88.1 | 100 | 55.2 | 71.1 | 100 | 90.5 | 95 |
| column 3 | 94.4 | 84.1 | 89 | 98.7 | 52.3 | 68.4 | 100 | 84.4 | 91.5 |

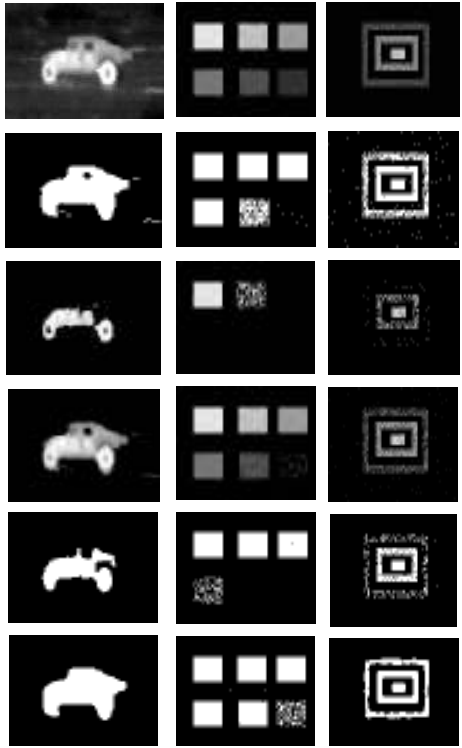

Figure 7.8: The segmentation results of the Otsu's and the multi-threshold Otsu's methods on some of the scalar-valued images considered in this chapter. The first row shows the original images. The second row shows the segmentation results, corresponding to the images of first row, obtained by the Otsu's method. The third row shows the object of interests obtained by the multi-threshold Otsu's method when the number of thresholds is five. The fourth row shows the merged objects obtained by first applying the multi-Otsu's method when the number of thresholds is 2, 3, 4, and 5, then merging some of the obtained objects. The fifth row shows the segmentation results of the Otsu's method applied on the images of the fourth row. Finally, the sixth row shows the segmentation results obtained by *SOMCV* on the images of the first row.

Finally, we have trained the neural map on a single frame of a real aircraft video [70] (the top left image in Fig. 7.9(a)) and applied the trained network on-line to segment individually - using *SOMCV* - some of its *RGB*-frames, which are shown in Fig. 7.9(a) (the initial contours for the video frames are similar to the initial contour - shown in red - which has been used for the first image). Fig. 7.9(b) shows the segmentation results of *SOMCV* in handling the selected frames in Fig. 7.9(a) and demonstrates its robustness to scene changes and object motions. Concluding, this experiment hightlights the robustness of *SOMCV* model to the contour initialization, scene changes and illumination variations when being used in an on-line framework.

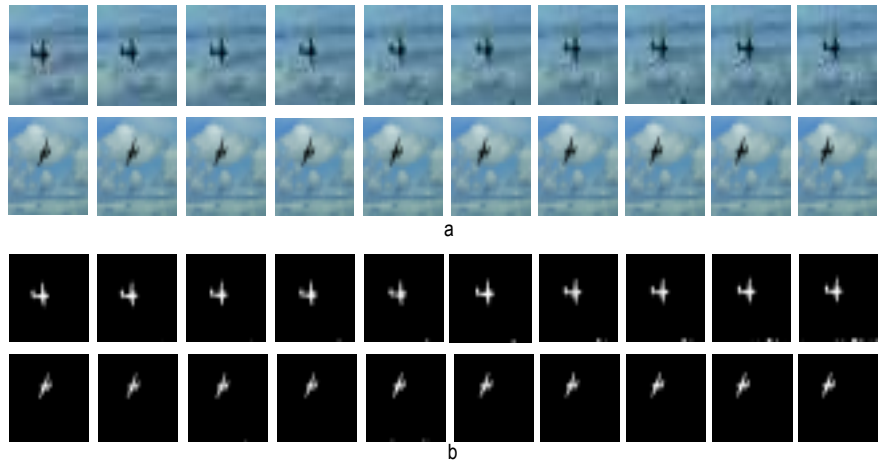

Figure 7.9: The robustness of the *SOMCV* model to scene changes and moving objects. (a) The first row shows the original early frames (frames 50-59, from left to right) of a real-aircraft video while later frames (frames 350-359, from left to right) are shown in the second row. (b) shows the segmentation results obtained by *SOMCV*, on the frames shown in part (a).

## 7.5 Summary

In this chapter, we have proposed a novel global *ACM*, termed *SOM-based Chan-Vese* (*SOMCV*). The *SOMCV* model is a global and an unsupervised *ACM* that integrates effectively the advantages of *ACM*s and self-organizing networks. *SOMCV* has a *Self-Organizing Topology Preservation* (*SOTP*) property, which allows to preserve the topological structures of the foreground/background intensity distributions during the active contour evolution. Indeed, *SOMCV* relies on a set of self-organized neurons by automatically extracting the prototypes of selected neurons as global regional descriptors and iteratively, in an unsupervised way, integrates them during the evolution of the contour.

In order to highlight the robustness of *SOMCV*, several synthetic and real images with different kinds of intensity distributions have been handled effectively in the experimental studies presented in Section 8.4. Also the variation of *SOMCV* - the *SOMCV$_s$* model - has provided good results in most cases. The capability of *SOMCV* and *SOMCV$_s$* to handle images globally without relying on a particular statistical assumption is the main contribution of this chapter. Moreover, the effectiveness and robustness of *SOMCV* and *SOMCV$_s$* may find applications in various other problems in computer vision. In a similar way to the *SOMCV*, in the following chapter we also describe another *SOM*-based *ACM*, that we have proposed which takes advantage of both local and global information in order to improve the robustness of the segmentation to the contour initialization, and to the presence of noise and intensity inhomogenity.

# Chapter 8

# *SOM*-based Regional *AC* Model

## 8.1 Introduction

Local Active Contour Models (local *ACM*s) constitute an efficient image segmentation framework, which is driven by local information about the intensity of an image. Most of the existing local *ACM*s can handle images with intensity inhomogeneity by integrating explicity local intensity information into the objective functional to be optimized. However, in this case, the success of local *ACM*s depends on how accurate the initial contour is. Then, a challenge in the use of such models consists in handling images with intensity inhomogeneity in such a way to obtain robustness with respect to the contour initialization. In this chapter we propose a model, termed *SOM-based Regional Active Contour* (*SOM-RAC*) model, with such a property. The *SOM-RAC* model relies on the global information coming from selected prototypes associated with a Self Organizing Map (*SOM*), which is trained off-line to model the intensity distribution of an image, and used on-line to segment an identical or similar image. In order to improve the robustness of the model, global and local information are combined in the on-line process, as the selection of the weights of the trained *SOM* is driven by local information on the intensity of the image.

Experimental results show the high accuracy of the segmentations obtained by the *SOM-RAC* model on several synthetic and real images, when compared with a state-of-the-art local *ACM*, the *Local Region-based Chan-Vese model*.

The main motivation for this model is to combine global and local intensity information in an *ACM* through a *SOM*-based approach. In the first phase of *SOM-RAC*, a set of neurons is trained to model globally the intensity distribution of the image by a self-organization learning procedure. Such weights are used to integrate the intensity distribution implicitly - as global Region of Interest (*ROI*) descriptors - into the objective functional of the proposed *SOM-RAC* model, to guide the evolution of the active contour. In a second phase, local image intensity information is used during the actual segmentation process in combination with the global information coming from the weights of the trained *SOM*, and is combined with the global information above inside the objective functional. In this way, the proposed *ACM* model is able to make use of both global and local information.

## 8.2 The *SOM-RAC* model

In this section, we describe our *SOM*-based Regional Active Contour (*SOM-RAC*) model. Although the model is presented here for the case of scalar-valued images (e.g., gray-level images), it can be extended straightforwardly to the case of vector-valued images (e.g., *RGB* images).

The *SOM-RAC* segmentation framework is composed of an unsupervised training session and a testing session. The two sessions are performed, respectively, off-line and on-line.

The training session of this model is the same as the training session of the *SOMCV* model, which was presented in chapter 7 (see Subsection 7.2.1).

Once its training has been accomplished, the *SOM* network is applied in the testing session, during the evolution of the active contour *C*, to approximate and describe globally the foreground and background intensity distributions of a similar test image $I(x)$ to be segmented. The use of such a global information helps in providing to the model robustness to the contour initialization and to the additive noise. Moreover, during the active contour evolution, a combination of local and global information is exploited in order to provide to the model robustness to the intensity inhomogeneity and to possible changes in the intensity distribution itself, when moving from the training image to the test one. More precisely, as a first step, one determines, for each pixel $x \in \Omega$ (the domain of the image), the best-matching-unit (*BMU*) neuron $w_b(x)$ to the local weighted mean intensity of the image

$$c(x) := \frac{\int_\Omega g_{\sigma^*}(x-y)I(y)\,dy}{\int_\Omega g_{\sigma^*}(x-y)dy}, \qquad (8.1)$$

i.e.,

$$w_b(x) := \operatorname{argmin}_{w_n} |w_n - c(x)|, \qquad (8.2)$$

where $g_{\sigma^*}$ is a Gaussian kernel function with $\int_{\mathbb{R}^2} g_{\sigma^*}(x)dx = 1$ and width $\sigma^* > 0$. Such a choice of $w_b(x)$ does not depend on the current contour. Then, the two local weighted mean intensities in the foreground and the background, respectively as defined in Equations 8.3 and 8.4, are compared to $w_b(x)$, to define suitable regional intensity descriptors.

$$c_+(x,C) := \frac{\int_{\text{in}(C)} g_\sigma(x-y)I(y)\,dy}{\int_{\text{in}(C)} g_\sigma(x-y)dy}, \qquad (8.3)$$

$$c_-(x,C) := \frac{\int_{\text{out}(C)} g_\sigma(x-y)I(y)\,dy}{\int_{\text{out}(C)} g_\sigma(x-y)dy}. \qquad (8.4)$$

where $0 < \sigma^* < \sigma$. More precisely, one sets

$$A_b^+(x, C) := \left| w_b(x) - \frac{\int_{\text{in}(C)} g_\sigma(x - y) I(y) \, dy}{\int_{\text{in}(C)} g_\sigma(x - y) dy} \right|, \quad (8.5)$$

$$A_b^-(x, C) := \left| w_b(x) - \frac{\int_{\text{out}(C)} g_\sigma(x - y) I(y) \, dy}{\int_{\text{out}(C)} g_\sigma(x - y) dy} \right|, \quad (8.6)$$

which are, respectively, the distances of the weight $w_b(x)$ from the two local weighted mean intensities in the two regions around the pixel $x$. Then, one defines the two weights

$$w_b^+(x, C) := \begin{cases} w_b(x), & \text{if } A_b^+(x,C) < A_b^-(x,C), \\ 0, & \text{otherwise}, \end{cases} \quad (8.7)$$

$$w_b^-(x, C) := \begin{cases} w_b(x), & \text{if } A_b^+(x,C) > A_b^-(x,C), \\ 0, & \text{otherwise}. \end{cases} \quad (8.8)$$

Such weights are extracted as regional intensity descriptors and included in the objective functional to be minimized in our proposed *SOM-RAC* model, which has the following expression[1]:

$$\begin{align}
E_{SOM-RAC}(C) &:= \lambda^+ \int_{\text{in}(C)} e^+(x, C) dx \\
&\quad + \lambda^- \int_{\text{out}(C)} e^-(x, C) dx, \quad (8.9) \\
e^+(x, C) &:= \left( I(x) - w_b^+(x, C) \right)^2, \quad (8.10) \\
e^-(x, C) &:= \left( I(x) - w_b^-(x, C) \right)^2. \quad (8.11)
\end{align}$$

where the parameters $\lambda^+, \lambda^- \geq 0$ are, respectively, the weights of the image energy terms $\int_{\text{in}(C)} e^+(x, C) dx$ and $\int_{\text{out}(C)} e^-(x, C) dx$, inside and outside the contour.

---

[1] When one of the index set is empty, one can use several strategies: e.g., replacing the associated summation with its value at its last previous evaluation (however, this was not needed for the case studies shown in Section 8.4).

Now, we replace the contour curve C with the level set function $\phi$, obtaining

$$E_{SOM-RAC}(\phi) = \lambda^+ \int_{\phi>0} e^+(x,\phi)dx + \lambda^- \int_{\phi<0} e^-(x,\phi)dx, \quad (8.12)$$

where we have also made explicit the dependence of the functions $e^+$ and $e^-$ on $\phi$. In terms of the Heaviside step function $H(\cdot)$, the *SOM-RAC* objective functional can be also written as follows:

$$\begin{aligned} E_{SOM-RAC}(\phi) &= \lambda^+ \int_\Omega e^+(x,\phi)H(\phi(x))dx \\ &+ \lambda^- \int_\Omega e^-(x,\phi)(1-H(\phi(x)))dx. \end{aligned} \quad (8.13)$$

Finally, the evolution of the contour in the *SOM-RAC* is described by the *PDE*

$$\frac{\partial \phi}{\partial t} = \delta(\phi)[-\lambda^+ e^+ + \lambda^- e^-], \quad (8.14)$$

which shows how the learned neurons are used to determine the internal and external forces acting on the contour during its evolution. Apart from this difference, Eq. (8.14) can be solved iteratively using the same smoothing and discretization techniques used in the *C-V* model. Moreover, at each iteration of a finite-difference approximation of (8.14), we also perform a regularization of the current level set function by replacing it with its convolution with a Gaussian filter of suitable width $\sigma' > 0$. Such a convolution can be preceded by a binarization[2] of the function $\phi$, without loss of information about the current contour.

## 8.3 Implementation

The procedural steps of the two sessions of the *SOM-RAC* model are summarized in the following Algorithm 4.

---

[2]See formula (8.16) in Algorithm 4.

## Algorithm 4 *SOM-RAC* segmentation framework

1: **procedure**
   - Input:
     - Training and test scalar-valued images.
     - Topology of the network.
     - Number of iterations $t_{\max}^{(tr)}$ for training the neural map.
     - Maximum number of iterations $t_{\max}^{(evol)}$ for the contour evolution.
     - $\eta_0 > 0$: starting learning rate.
     - $r_0 > 0$: starting radius of the map.
     - $\tau_\eta, \tau_r > 0$: time constants in the learning rate and contour smoothing parameter.
     - $\lambda^+, \lambda^- \geq 0$: weights of the energy terms, respectively, inside and outside the contour.
     - $\sigma^\star, \sigma, \sigma' > 0$: Gaussian intensity and contour smoothing parameters.
     - $\rho > 0$: constant in the binary approximation of the level set function.
   - Output:
     - Segmentation result.

   *TRAINING SESSION:*
2:   Initialize randomly the weights of the neurons.
3:   **repeat**
4:     Choose randomly a pixel $x_t$ in the image domain $\Omega$ and determine the *BMU* neuron to the input intensity $I^{(tr)}(x_t)$.
5:     Update the weights $w_n$ using (5.1), (5.2), (5.3), and (5.4).
6:   **until** learning of the weights is accomplished (i.e., the number of iterations $t_{\max}^{(tr)}$ is reached).

   *TESTING SESSION:*
7:   Choose a subset $\Omega_0$ (e.g., a rectangle) in the image domain $\Omega$ with boundary $\Omega_0'$, and initialize the level set function as:

$$\phi(x) := \begin{cases} \rho, & x \in \Omega_0 \setminus \Omega_0', \\ 0, & x \in \Omega_0', \\ -\rho, & x \in \Omega \setminus (\Omega_0 \cup \Omega_0'). \end{cases} \quad (8.15)$$

8:   **repeat**
9:     Determine, for each pixel $x \in \Omega$, the weight $w_b(x)$ from (8.2), then, for the current contour $C$ and each pixel $x \in \Omega$, the quantities $A_b^+(x,C)$ and $A_b^-(x,C)$ from (8.5) and (8.6), finally the weights $w_b^+(x,C)$ and $w_b^-(x,C)$ from (8.7) and (8.8).
10:    Evolve the level set function $\phi$ according to a finite difference approximation of (8.14).
11:    At each iteration of the finite-difference scheme, re-initialize the current level set function to be binary by performing the update

$$\phi \leftarrow \rho \left( H(\phi) - H(-\phi) \right), \quad (8.16)$$

then regularize by convolution the obtained level set function:

$$\phi \leftarrow g_{\sigma'} \otimes \phi, \quad (8.17)$$

where $g_{\sigma'}$ is a Gaussian kernel with $\int_{\mathbb{R}^2} g_{\sigma'}(x)dx = 1$ and width $\sigma'$.

12:  **until** the curve evolution converges (i.e., the curve does not change anymore) or the maximum number of iterations $t_{\max}^{(evol)}$ is reached.
13: **end procedure**

## 8.4 Experimental study

In this section, we demonstrate experimentally the efficiency and robustness of the *SOM-RAC* model in handling real and synthetic images, as compared with the *LRCV* model described in Chapter 2. For a fair comparison, in these experiments both the *SOM-RAC* and the *LRCV* models are implemented in Matlab R2012a on a PC with the following configuration: 1.8 GHz Intel(R) Core(TM) i3-3217U, and 4.00 GB RAM.

In each experiment, the $r_0$, $\sigma$ and $\sigma'$ parameters are expressed in pixels. Moreover, the *SOM-RAC* parameters are fixed as follows: $\eta_0 = .9$, $\sigma^* = 0.1$, $\sigma = 30$ (apart from the experiments described in Figure 8.6, in which several values for $\sigma$ have been considered), $\sigma' = 1.5$, and $\lambda^+ = \lambda^- = 1$. Also, $r_0 := \max(M,N)/2$, where $M = N = 4$ are the numbers of rows and columns of the *SOM* in the *SOM-RAC* model, $t_{\max}^{(tr)} = 10000$, $t_{\max}^{(evol)} = 1000$, $\tau_\eta := t_{\max}^{(tr)}$, $\tau_r := t_{\max}^{(tr)}/\ln(r_0)$, $\rho = 1$. In the *LRCV* model, $\lambda^+$, $\lambda^-$ are also fixed to 1, and the same values of $\sigma$ and $\sigma'$ as above are considered. Unless stated otherwise, the training image used in the unsupervised training session coincides with the test image. Otherwise, it is an image similar to the test image (obtained, e.g., by adding Gaussian noise). In all the testing sessions, the initial contour has been chosen as rectangular (which is a standard choice for *ACM*s). All the considered gray-level images are 8-bit images, so the range of the values assumed by the intensity is 0-255.

To illustrate the effectiveness of *SOM-RAC* in handling real and synthetic images with respect to the *LRCV* model, in Fig. 8.1 we compare the segmentation results obtained by the *SOM-RAC* and the *LRCV* models, the contour initialization and amount of local information (controlled by the parameter $\sigma$) being the same for the two models. As such figure illustrates, the *SOM-RAC* model is more effective in handling real and synthetic images. More precisely, the segmentation results obtained by the *SOM-RAC* model on the images considered in this experiment show its ability of segmenting images with intensity inhomogeneity and objects with blurred edges (first and second rows), images with intensity inhomogeneity only (third and fourth rows), and a synthetic image containing a shadow (fifth row). The final contours obtained by the *SOM-RAC* model demonstrate its high effectiveness, whereas unsatisfying results are obtained in these cases by *LRCV* model, due to its sensitivity to the contour initialization.

The experiments illustrated in the remaining of this section can be divided into four parts, as they test the robustness of the *SOM-RAC* model, respectively, to the contour initialization, additive noise, scene changes, and choice of the locality parameter $\sigma$.

In Fig. 8.2 and 8.3, we test the robustness of the *SOM-RAC* model in handling images with different initial rectangular contours. Fig. 8.2 illustrates the robustness of *SOM-RAC* to contour initialization in handling a synthetic image with intensity inhomogeneity when compared to the *LRCV* model. As such figure shows, in this experiment the *SOM-RAC* model is less sensitive to the location of the initial contour than the *LRCV* model, and the final contours obtained by the *SOM-RAC* model converge to the true object boundary with similar performances for all the considered contour initializations. Additionally, the robustness of *SOM-RAC* to contour initialization in handling a real image with intensity inhomogeneity and weak boundaries is illustrated in Fig. 8.3 in comparison with the *LRCV* model. In this case, the final contours obtained by *SOM-RAC* for all the initial contours show its ability to find the object with a high accuracy, while for the case of the *LRCV* model a leaking problem occurs for all the initial contours.

Fig. 8.4 illustrates the effectiveness and robustness of the *SOM-RAC* model in handling the synthetic image already shown in Fig. 8.2 with the addition of different Gaussian noise levels. In this experiment, the *SOM-RAC* model has been trained off-line on the image shown in the first row, then it has been used to segment its noisy versions in the on-line phase. As shown by the figure, the segmentations obtained in this experiment by the *SOM-RAC* model demonstrate its small sensitivity to the additive noise. On the other hand, in this experiment the *LRCV* model is more sensitive to the additive noise, which affects gradually its segmentation results.

In order to demonstrate the robustness of the *SOM-RAC* model with respect to scene changes, we have trained off-line the *SOM-RAC* model on the images shown in the first column of Fig. 8.5, then we have applied it on-line to segment different images, as shown in the second column of Fig. 8.5. Then, we have also trained it on the images shown in the third column to segment the images in the fourth column. In this experiment, the robustness of *SOM-RAC* has been confirmed in several situations: when intensity inhomogeneity occurs in real and synthetic images with weak boundaries (first row of Fig. 8.5), in the presence/absence of the intensity inhomogeneity (second row), in synthetic images (third row), and in handling images containing an overlap of the foreground/background intensity distributions.

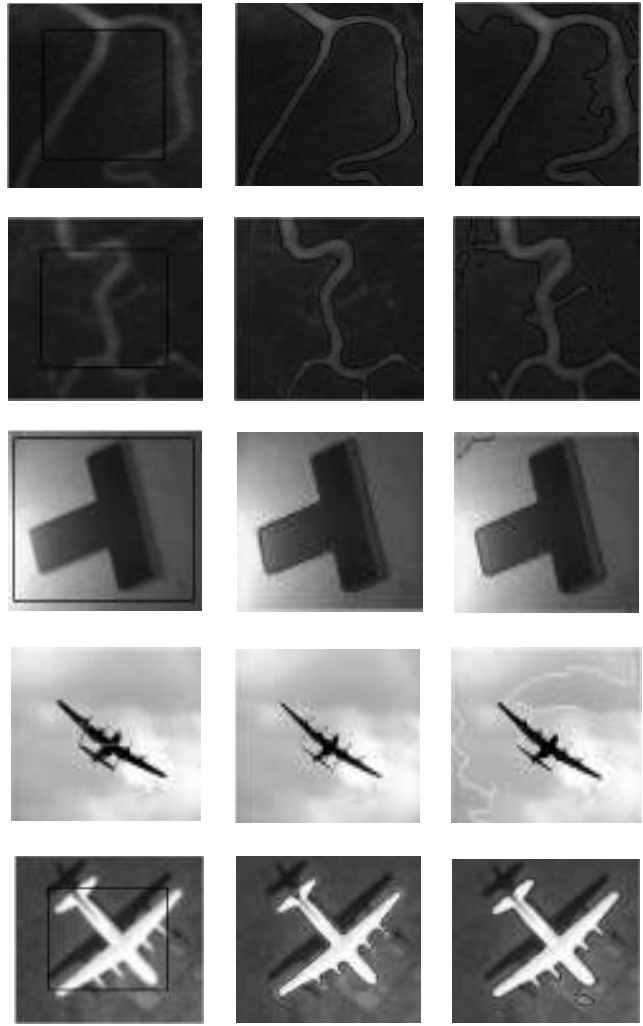

Figure 8.1: The segmentation results obtained on real and synthetic images by the *SOM-RAC* and the *LRCV* models. The first column shows the original images with the initial contours, while the second and third columns show, respectively, the corresponding segmentation results obtained by the two models.

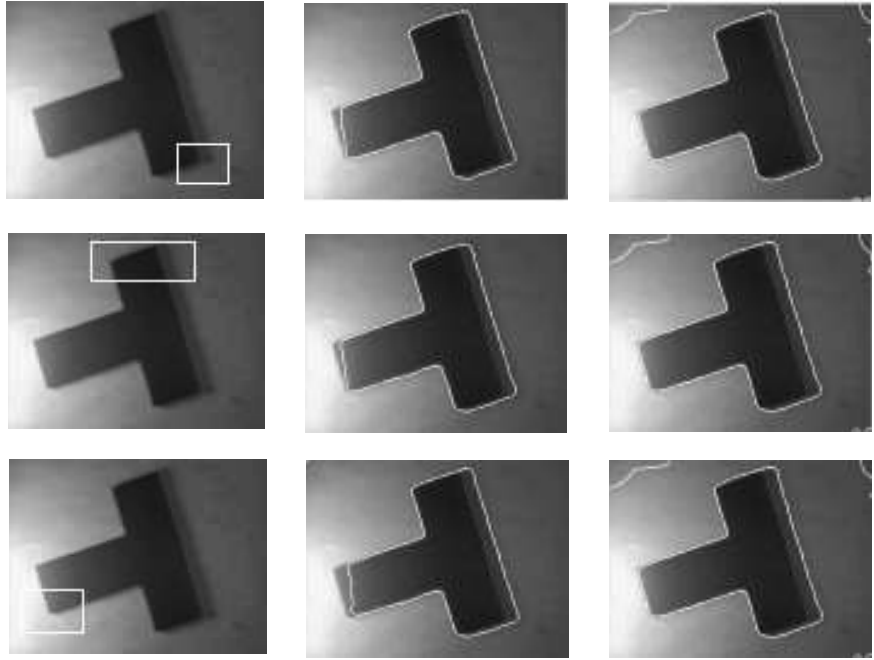

Figure 8.2: The robustness of the *SOM-RAC* model with respect to the contour initialization, as compared to the segmentation results obtained by the *LRCV* model, for a synthetic 127 × 96 image with intensity inhomogeneity. The first column shows the original image with three different rectangular initial contours (in white). The second and third columns show, respectively, the segmentation results obtained by the *SOM-RAC* and the *LRCV* models.

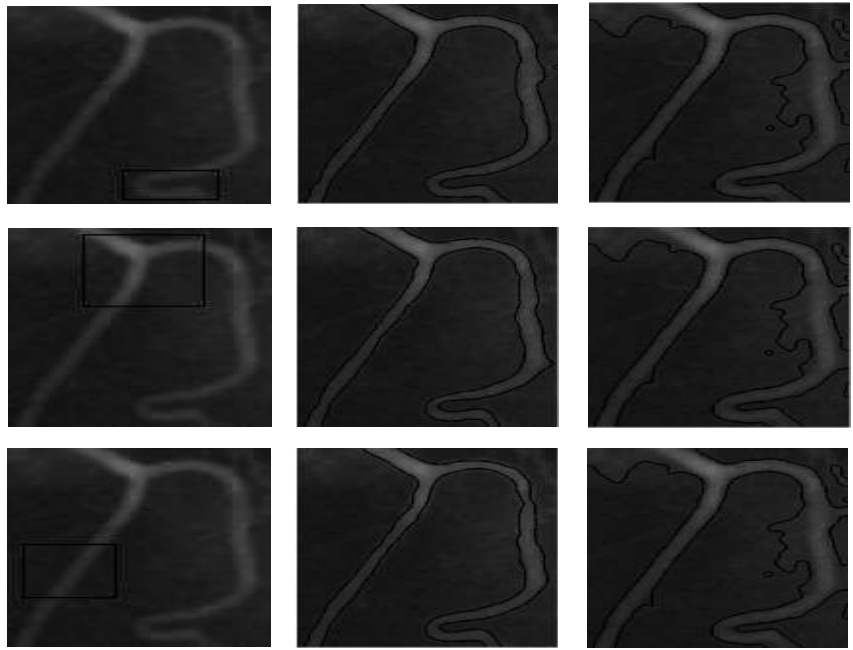

Figure 8.3: The robustness of the *SOM-RAC* model with respect to the contour initialization in handling a real 103 × 131 image in the presence of intensity inhomogeneity and weak edges, as compared to the segmentation results obtained by the *LRCV* model on the same image. The first column shows the original image with three different rectangular initial contours (in black). The second and third columns show, respectively, the segmentation results obtained by the *SOM-RAC* and the *LRCV* models.

Finally, in order to study the robustness of the *SOM-RAC* model with respect to the locality parameter $\sigma$, we have trained the *SOM-RAC* model on the same synthetic image already considered in Fig. 8.5, then we have used it on-line to segment the same image for several values of $\sigma$ (the initial contour being the same as the one shown in Fig. 8.5 for the same image). Then we have also applied the *LRCV* model to segment the same image, for the same choices of σ. Fig. 8.6 (a) demonstrates the robustness of *SOM-RAC* to changes in the parameter σ, since in all the considered cases the objects are segmented accurately when σ is more than or equal to 20. On the other hand, as shown in Fig. 8.6 (b), in this experiment the *LRCV* model is able to find the object correctly only when σ is equal to 15.

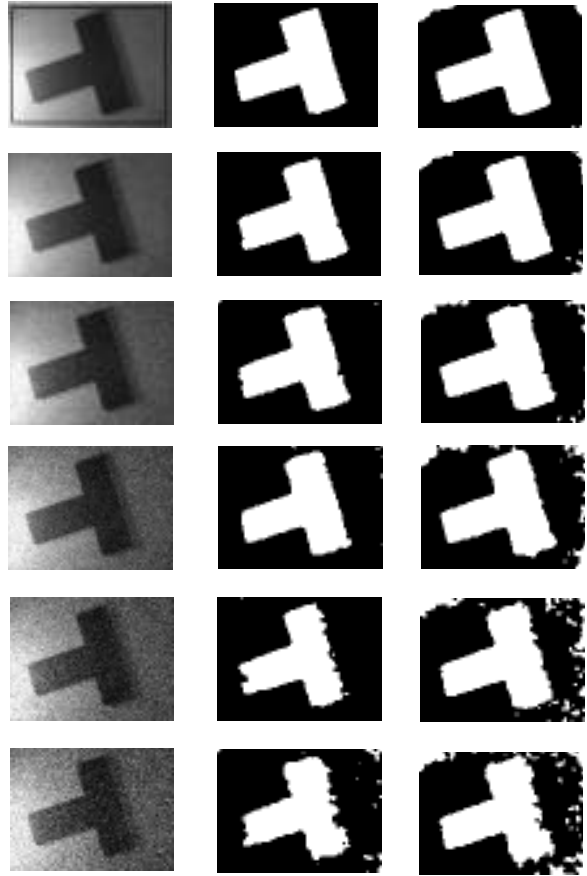

Figure 8.4: The robustness of the *SOM-RAC* model with respect to additive noise in handling a synthetic 127×96 image in the presence of intensity inhomogeneity, as compared to the segmentation results obtained by the *LRCV* model. The first column shows, from top to down, the image of Fig. 8.2 with the addition of different Gaussian noise levels (with standard deviations $SD$ = 0, 5, 10, 15, 20, and 25, respectively). The second and third columns show, respectively, the segmentation results obtained by the *SOM-RAC* and the *LRCV* models.

## 8.5 Summary

In this chapter, with the aim of developing an *ACM* that is at the same time effective and robust in handling complex images containing intensity inhomogeneity, we have proposed a novel *ACM*, termed *SOM-based Regional Active Contour* (*SOM-RAC*) model. The main motivation for the *SOM-RAC* model is to deal with some drawbacks of global *ACM*s and local *ACM*s through the combination of global and local information by a *SOM*-based approach. Indeed, global information plays an important role to improve the robustness of *ACM*s against the contour intialization and the additive noise but - if used alone - it is usually not sufficient to handle images containing intensity inhomogeneity. On the other hand, local information allows one to deal effectively with the intensity inhomogeneity but - if used alone - it produces usually *ACM*s very sensitive to the contour initialization. The *SOM-RAC* model combines both kinds of information relying on global regional descriptors (i.e., suitably selected weights of a trained *SOM*) on the basis of local regional descriptors (i.e., the local weighted mean intensities). In this way, the *SOM-RAC* model is able to integrate the advantages of local *ACM*s and Self Organizing Maps.

In order to highlight the robustness of the proposed *SOM-RAC* model, we have tested and compared it with a state-of-the-art local *ACM*. In the experimental studies presented in Section 8.4, several synthetic and real images with various intensity distributions and locations of the initial contours have been handled effectively by the proposed *SOM-RAC* model, which has also out-performed the other *ACM* model, showing more robustness with respect to several factors. In the following chapter, we conclude the thesis by highlighting the motivation and contributions of the presented models and, identifying some possible future research directions.

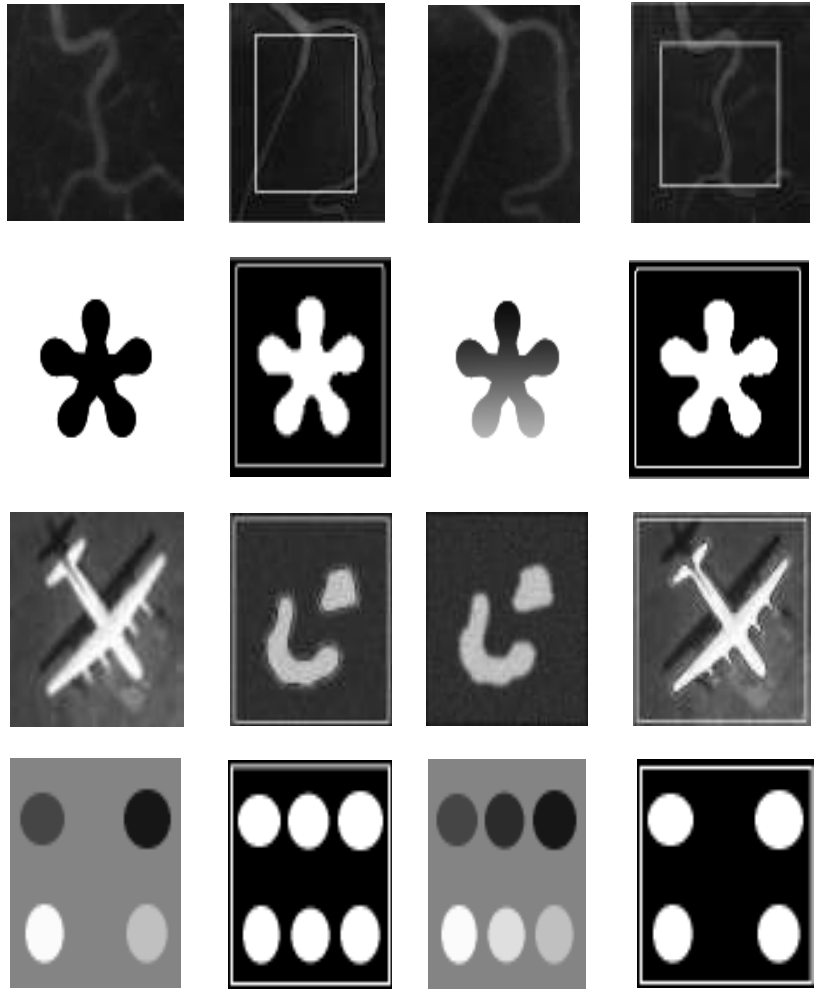

Figure 8.5: The robustness of the *SOM-RAC* model with respect to scene changes. The first and third columns show the training images while the second and fourth columns show, respectively, the segmentations results obtained by the *SOM-RAC* model on different test images. For each row, the *SOM-RAC* model was trained on the first (respectively third) image, then it was used to segment the second (respectively, fourth) image. For the second and fourth columns, the initial contours are shown in white, whereas the final segmentation results are shown in black.

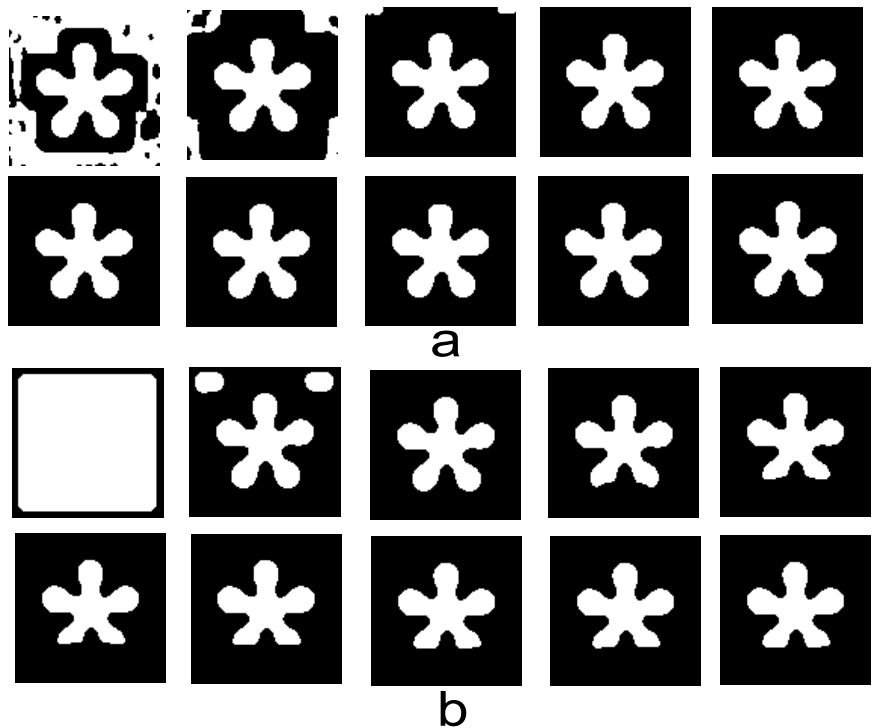

Figure 8.6: The robustness of the *SOM-RAC* model with respect to the locality parameter $\sigma$ in handling a synthetic $100 \times 100$ image (already shown in the second row of Fig. 8.5) when compared to *LRCV* model (the same initial contour has been used in the comparison). Parts (a) shows the segmentation results obtained by the *SOM-RAC* model with $\sigma = 5, 10, 15, 20, 25$ in the first row and $\sigma = 30, 35, 40, 45, 50$ in the second row. Part (b) shows the corresponding segmentation results obtained by the *LRCV* model.

# Chapter 9

# Conclusions and Future Work

In this thesis, we first provided a literature review to show various kinds of active contour models to deal with some challenging problems in computer vision. Moreover, a number of novel models have been presented to deal (in effective, efficient, and/or robust way) with the image segmentation problem, and have been compared with state-of-the-art active contour models.

In chapter 4, a novel energy based-active contour model based on a new Globally Signed Region Pressure Force (*GSRPF*) function has been proposed. The *GSRPF* considers the global information extracted from a ROI and accommodates also foreground intensity distributions that are not necessarily symmetric. It automatically and efficiently modulates the signs of the pressure forces inside and outside the contour. Compared with other *ACMs*, the resulting method is less sensitive to noise, contour initialization, and can handle images with complex intensity distributions in the foreground and/or background. *GSRPF* is a Gaussian regularizing level set model that relies only on a single parameter. It is designed to have a quadratic behaviour and converge in a few iterations without penalizing the segmentation accuracy. Results on synthetic and real images from a variety of scenarios have demonstrated the superior segmentation accuracy of *GSRPF*, when com-pared with well regarded global level set methods, such as the *SBGFRLS* and *C-V* models.

Chapter 5 describes a novel *SOM*-based *ACM* model, the Concurrent Self Organizing Map-based Chan-Vese model (*CSOM-CV*), which relies mainly on a set of prototypes coming from two trained *SOMs* to guide the evolution of the active contour. *CSOM-CV* is a supervised and global region-based *ACM*. It has been demonstrated to be efficient and robust to the noise. As compared to the *C-V* model, our proposed solution consists instead in modeling globally in a supervised way the intensity distributions of the foreground/background (relying on a few supervised pixels) without using parametric models, but relying on a set of prototypes resulting from the training of a *CSOM*. Moreover, as compared to *CSOM* and in general to previous *SOM*-like models used in image segmentation, our solution consists in modeling the active contour using a variational level set method and relying at the same time on a few prototypes coming from the learned *CSOM*. In this way, the *CSOM-CV* model is able to produce a final segmentation result characterized by a smooth contour while most *SOM*-like models usually produce segmentations characterized by disconnected boundaries. Moreover, the *CSOM-CV* has shown to be more robust to two different kinds of noise.

We have also proposed a new supervised *ACM*, which we have termed *Self Organizing Active Contour* (*SOAC*) model in chapter 6. It is based on two sets of self organizing neurons to learn the dissimilarity between the foreground/background intensity distributions. In this way, the information about such distributions is integrated implicitly into the energy functional of the model by the learned prototypes of the two *SOMs*, helping in the guide of the contour evolution. *SOAC* is a Gaussian regularizing level set method, and it is robust to additive noise. The experimental results obtained on several synthetic and real images for both scalar-valued and vector-valued cases images demonstrate the high effectiveness and robustness of our model, when compared with state-of-the-art *ACMs* (e.g., *KDE*-based, *GMM*-based, *C-V*, *LRCV* models), in segmenting images with overlap between the foreground/background intensity distributions, intensity inho-mogeneity, and/or containing objects characterized by many different intensities.

A novel global *ACM*, termed *SOM-based Chan-Vese* (*SOMCV*) was proposed in chapter 7. The *SOMCV* model is a global and an unsupervised *ACM* that integrates globally and effectively the advantages of *ACM*s and self-organizing networks. *SOMCV* has a *Self-Organizing Topology Preservation* (*SOTP*) property, which allows to preserve the topological structures of the foreground/background intensity distributions during the active contour evolution. Indeed, *SOMCV* relies on a set of self-organized neurons by automatically extracting the prototypes of selected neurons as global regional descriptors and iteratively, in an unsupervised way, integrates them during the evolution of the contour. The robustness, effectiveness and efficiency of the *SOMCV* model on several synthetic and real images with different kinds of intensity distributions has been demonstrated in that chapter and compared to a global *ACM* (e.g., the *C-V*) and other thresholding-based models (e.g., Otsu's and Multi-level Otsu's methods).

Eventually, with the aim of developing an *ACM* that is at the same time effective and robust in handling complex images containing intensity inhomogeneity, we have proposed, in chapter 8, another novel *ACM*, termed *SOM-based Regional Active Contour* (*SOM-RAC*) model. The main motivation for the *SOM-RAC* model is to deal with the sensitivity of Local *ACM*s to the contour initialization (when intensity inhomogeneity and additive noise occur in the images) through the combination of global and local information by a *SOM*-based approach. Indeed, global information plays an important role to improve the robustness of *ACM*s against the contour intialization and the additive noise but - if used alone - it is usually not sufficient to handle images containing intensity inhomogeneity. On the other hand, local information allows one to deal effectively with the intensity inhomogeneity but - if used alone - it produces usually *ACM*s very sensitive to the contour initialization. The *SOM-RAC* model combines both kinds of information relying on global regional descriptors (i.e., suitably selected weights of a trained *SOM*) on the basis of local regional descriptors (i.e., the local weighted mean intensities). In this way, the *SOM-RAC* model is able to integrate the advantages of local *ACM*s and Self Organizing Maps. In order to highlight the robustness of the proposed *SOM-RAC* model, we have tested and compared it with a state-of-the-art local *ACM* (e.g., *LRCV*) on several synthetic and real images with various intensity distributions and locations of the initial contours, showing more robustness with respect to several factors.

As discussed above, several *ACM*s have been proposed in the thesis, relying mainly on prior intensity information. As a possible future research direction, our models could benefit from other kind of prior information such as shape information. With the addition of such information, our models could behave nicely in handling complex images with some other kinds of challenging problems such as occlusion. Other future research directions include developing the machine learning components of our models from a streaming learning perspective, in order to better understand the content of videos, and handling them in real time. This could be possible by integrating streaming learning algorithms into the segmentation framework of our models.